\documentclass{article}

    \PassOptionsToPackage{numbers, compress}{natbib}
\usepackage[final]{neurips_2024}

\usepackage[utf8]{inputenc} %
\usepackage[T1]{fontenc}    %
\usepackage{url}            %
\usepackage{booktabs}       %
\usepackage{multirow}
\usepackage{siunitx}
\usepackage[table]{xcolor}
\usepackage{cancel}
\usepackage{tablefootnote}

\newcommand{\std}[1]{\textcolor{black}{\scriptsize{$\pm #1$}}}
\newcommand{\highlight}[1]{\cellcolor{blue!10}{#1}}
\newcommand{\highlightub}[1]{\cellcolor{green!10}{#1}}

\usepackage{amsmath,amsfonts,amssymb}       %
\usepackage{nicefrac}       %
\usepackage{microtype}      %
\usepackage{xcolor}         %
\usepackage{float}
\usepackage{graphicx}

\usepackage{booktabs} %
\usepackage{array}    %

\usepackage{algorithm,algorithmic}

\usepackage{wrapfig}

\usepackage{microtype}
\usepackage{graphicx}
\usepackage{subcaption}
\usepackage{booktabs} 
\usepackage{chngcntr}

\usepackage[colorlinks=true, allcolors=blue]{hyperref}

\usepackage{enumitem}

\usepackage{amsmath}
\usepackage{amssymb}
\usepackage{mathtools}
\usepackage{amsthm}
 \usepackage{textcomp}

\usepackage{amsthm}
\usepackage{xcolor}
\usepackage{amsmath,amsfonts,bm}
\makeatletter
\newtheorem*{rep@theorem}{\rep@title}
\newcommand{\newreptheorem}[2]{%
\newenvironment{rep#1}[1]{%
 \def\rep@title{#2 \ref{##1}}%
 \begin{rep@theorem}}%
 {\end{rep@theorem}}}
\makeatother

\newreptheorem{theorem}{Theorem}
\definecolor{myred}{RGB}{215,48,39}
\definecolor{mygreen}{RGB}{26,152,80}

\newcommand{\halfmark}{\textcolor{gray}{\checkmark\kern-1.1ex\raisebox{.7ex}{\rotatebox[origin=c]{125}{--}}}}
\usepackage[framemethod=TikZ]{mdframed}
\mdfdefinestyle{MyFrame}{%
    linecolor=black,
    outerlinewidth=.3pt,
    roundcorner=5pt,
    innertopmargin=1pt, %\baselineskip,
    innerbottommargin=1pt, %\baselineskip,
    innerrightmargin=1pt,
    innerleftmargin=1pt,
    backgroundcolor=black!0!white}

\mdfdefinestyle{MyFrame2}{%
    linecolor=white,
    outerlinewidth=1pt,
    roundcorner=1pt,
    innertopmargin=0,
    innerbottommargin=0,
    innerrightmargin=7pt,
    innerleftmargin=7pt,
    backgroundcolor=black!3!white}

\mdfdefinestyle{MyFrameEq}{%
    linecolor=white,
    outerlinewidth=0pt,
    roundcorner=0pt,
    innertopmargin=0pt,
    innerbottommargin=0pt,
    innerrightmargin=7pt,
    innerleftmargin=7pt,
    backgroundcolor=black!3!white}

\newcommand{\RNum}[1]{\uppercase\expandafter{\romannumeral #1\relax}}

\newcommand{\R}{\mathcal{R}}

\newcommand{\vertiii}[1]{{\left\vert\kern-0.25ex\left\vert\kern-0.25ex\left\vert #1 
    \right\vert\kern-0.25ex\right\vert\kern-0.25ex\right\vert}}
\newcommand{\vertiiii}[1]{{\vert\kern-0.25ex\vert\kern-0.25ex\vert #1 
    \vert\kern-0.25ex\vert\kern-0.25ex\vert}}

\usepackage{mathtools}

\DeclareMathOperator*{\argmax}{\arg\!\max}

\newcommand{\cut}[1]{}

\newcommand{\removelatexerror}{\let\@latex@error\@gobble}

\def\eqref#1{Eq.~\ref{#1}}
\def\1{\bm{1}}

\def\rva{{\mathbf{a}}}

\def\rvs{{\mathbf{s}}}

\def\rvw{{\mathbf{w}}}
\def\rvx{{\mathbf{x}}}
\def\rvy{{\mathbf{y}}}
\def\rvz{{\mathbf{z}}}

\DeclareMathAlphabet{\mathsfit}{\encodingdefault}{\sfdefault}{m}{sl}
\SetMathAlphabet{\mathsfit}{bold}{\encodingdefault}{\sfdefault}{bx}{n}

\def\gD{{\mathcal{D}}}
\def\gE{{\mathcal{E}}}

\def\gL{{\mathcal{L}}}

\def\gN{{\mathcal{N}}}

\def\sE{{\mathbb{E}}}

\def\R{{\mathbb{R}}}

\newcommand{\KL}{D_{\mathrm{KL}}}

\usepackage{newtxmath}

\usepackage[createShortEnv,conf={no link to proof, text link},commandRef=autoref]{proof-at-the-end}

\newcommand{\spacehack}[1]{\relax}

\makeatletter

\renewcommand{\appendixautorefname}{\S\@gobble}
\renewcommand{\sectionautorefname}{\S\@gobble}
\renewcommand{\subsectionautorefname}{\S\@gobble}
\renewcommand{\subsubsectionautorefname}{\S\@gobble}

\makeatother

\newcommand{\post}{^{\rm post}}
\newcommand{\LRTB}{\gL_{\rm RTB}}

\newcommand{\ie}{\textit{i.e.}}
\newcommand{\eg}{\textit{e.g.}}

\newcommand{\thegithuburl}{\href{https://github.com/GFNOrg/diffusion-finetuning}{\tt https://github.com/GFNOrg/diffusion-finetuning}}
\newcommand{\thegithuburlabstract}{\href{https://github.com/GFNOrg/diffusion-finetuning}{\tt this link}}

\makeatletter

\providecommand{\section}{}
\renewcommand{\section}{%
  \@startsection{section}{1}{\z@}%
                {-1.5ex \@plus -0.5ex \@minus -0.2ex}%
                { 1.0ex \@plus  0.3ex \@minus  0.2ex}%
                {\large\bf\raggedright}%
}
\providecommand{\subsection}{}
\renewcommand{\subsection}{%
  \@startsection{subsection}{2}{\z@}%
                {-1.3ex \@plus -0.5ex \@minus -0.2ex}%
                { 0.3ex \@plus  0.2ex}%
                {\normalsize\bf\raggedright}%
}
\providecommand{\subsubsection}{}
\renewcommand{\subsubsection}{%
  \@startsection{subsubsection}{3}{\z@}%
                {-1.0ex \@plus -0.5ex \@minus -0.2ex}%
                { 0.3ex \@plus  0.2ex}%
                {\normalsize\bf\raggedright}%
}
\providecommand{\paragraph}{}
\renewcommand{\paragraph}{%
  \@startsection{paragraph}{4}{\z@}%
                {0.2ex \@plus 0.2ex \@minus 0.2ex}%
                {-1em}%
                {\normalsize\bf}%
}
\providecommand{\subparagraph}{}
\renewcommand{\subparagraph}{%
  \@startsection{subparagraph}{5}{\z@}%
                {1.5ex \@plus 0.5ex \@minus 0.2ex}%
                {-1em}%
                {\normalsize\bf}%
}

\makeatother

\title{Amortizing intractable inference in diffusion models\\for vision, language, and control}

\author{%
  Siddarth Venkatraman\textsuperscript{*}\\Mila, Universit\'e de Montr\'eal
    \And
    Moksh Jain\textsuperscript{*}\\Mila, Universit\'e de Montr\'eal
    \And
    Luca Scimeca\textsuperscript{*}\\Mila, Universit\'e de Montr\'eal
    \And
    Minsu Kim\textsuperscript{*}\\Mila, Universit\'e de Montr\'eal\\KAIST
    \And
    Marcin Sendera\textsuperscript{*}\\Mila, Universit\'e de Montr\'eal\\Jagiellonian University
    \And
    Mohsin Hasan\\Mila, Universit\'e de Montr\'eal
    \And
    Luke Rowe\\Mila, Universit\'e de Montr\'eal
    \And
    Sarthak Mittal\\Mila, Universit\'e de Montr\'eal
    \And
    Pablo Lemos\\Mila, Universit\'e de Montr\'eal\\Ciela Institute\\Dreamfold
    \And
    Emmanuel Bengio\\Recursion
    \And
    Alexandre Adam\\Mila, Universit\'e de Montr\'eal\\Ciela Institute
    \And
    Jarrid Rector-Brooks\\Mila, Universit\'e de Montr\'eal\\Dreamfold
    \And
    Yoshua Bengio\\Mila, Universit\'e de Montr\'eal\\CIFAR
    \And
    Glen Berseth\\Mila, Universit\'e de Montr\'eal\\CIFAR
    \And
    Esmeralda S. Whitammer\\Mila, Universit\'e de Montr\'eal\\University of Edinburgh
    \AND
    \tt
$\left\{\text{\begin{minipage}{3.5in}\centering siddarth.venkatraman,moksh.jain,luca.scimeca\\minsu.kim,marcin.sendera,\dots,esmeralda.whitammer\end{minipage}}\right\}$@mila.quebec
}

\begin{document}

\maketitle

\begin{abstract}
\looseness=-1
Diffusion models have emerged as effective distribution estimators in vision, language, and reinforcement learning, but their use as priors in downstream tasks poses an intractable posterior inference problem. This paper studies \emph{amortized} sampling of the posterior over data, $\rvx\sim p\post(\rvx)\propto p(\rvx)r(\rvx)$, in a model that consists of a diffusion generative model prior $p(\rvx)$ and a black-box constraint or likelihood function $r(\rvx)$. We state and prove the asymptotic correctness of a data-free
learning objective, \emph{relative trajectory balance}, for training a diffusion model that samples from this posterior, a problem that existing methods solve only approximately or in restricted cases. Relative trajectory balance arises from the generative flow network perspective on diffusion models, which allows the use of deep reinforcement learning techniques to improve mode coverage. 
We illustrate the broad potential of unbiased inference of arbitrary posteriors under diffusion priors across a collection of experiments: in vision (classifier guidance), language (infilling under a discrete diffusion LLM), and multimodal data (text-to-image generation). Beyond generative modeling, we apply relative trajectory balance to the problem of continuous control with a score-based behavior prior, achieving state-of-the-art results on benchmarks in offline reinforcement learning. Code is available at \thegithuburlabstract.
\end{abstract}

\section{Introduction}
\label{sec:intro}

\looseness=-1
Diffusion models \cite{sohl2015diffusion,ho2020ddpm,song2021score} are a powerful class of hierarchical generative models, used to model complex distributions over images \cite{nichol2021improvedddpm,dhariwal2021diffusion,rombach2021high}, text \citep{austin2021structured,dieleman2022continuous,li2022diffusion,han-etal-2023-ssd,gulrajani2024likelihood,lou2023discrete}, and actions in reinforcement learning \cite{janner2022diffuser,wang2023diffusion,kang2024efficient} to name a few. In each of these domains, downstream problems require sampling product distributions, where a pretrained diffusion model serves as a prior $p(\rvx)$ that is multiplied by an auxiliary constraint $r(\rvx)$. For example, if $p(\rvx)$ is a prior over images defined by a diffusion model, and $r(\rvx)=p(c\mid\rvx)$ is the likelihood that an image $\rvx$ belongs to class $c$, then class-conditional image generation requires sampling from the Bayesian posterior $p(\rvx\mid c)\propto p(\rvx)p(c\mid\rvx)$. In offline reinforcement learning, if $\mu(a\mid s)$ is a conditional diffusion model over actions serving as a behavior policy, KL-constrained policy improvement~\citep{peng2019advantageweighted, lu2023contrastive} requires sampling from the normalized product of $\mu(a\mid s)$ with a Boltzmann distribution defined by a $Q$-function, $\pi^*(a\mid s)\propto \mu(a\mid s)\exp(\beta Q(s,a))$. In language modeling, various conditional generation problems~\citep{lou2023discrete,gulrajani2024likelihood,hu2023amortizing} amount to posterior sampling under a discrete diffusion model prior. \autoref{tab:examples} summarizes four such problems that the proposed method improves upon prior work.

\begin{figure}[t!]
    \vspace*{-1em}
    \centering
    \begin{minipage}{0.16\textwidth}
        \centering
        \includegraphics[width=\textwidth]{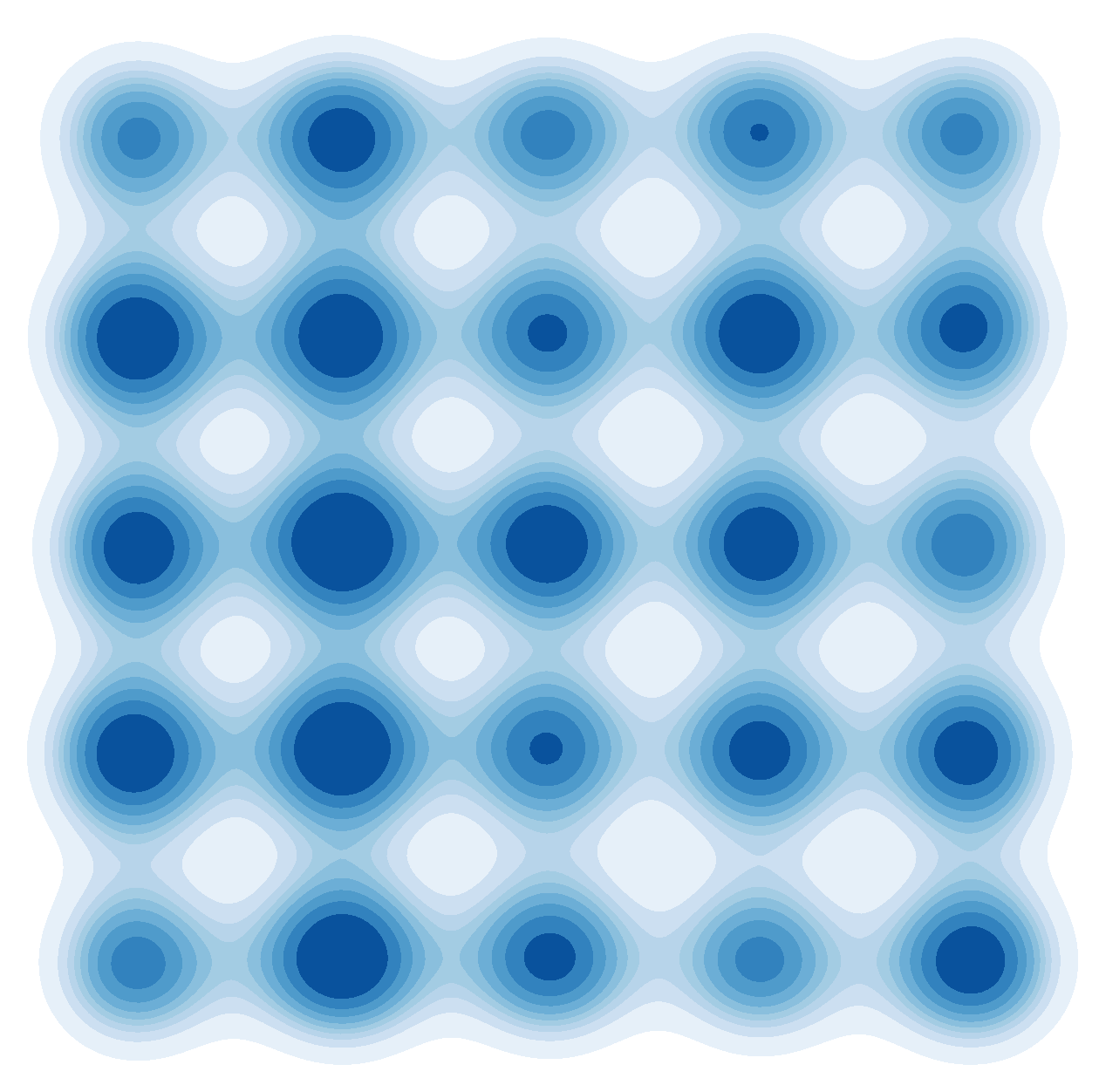} %
        \subcaption{Prior}
        \label{fig:2d_prior}
    \end{minipage}\hfill
    \begin{minipage}{0.16\textwidth}
        \centering
        \includegraphics[width=\textwidth]{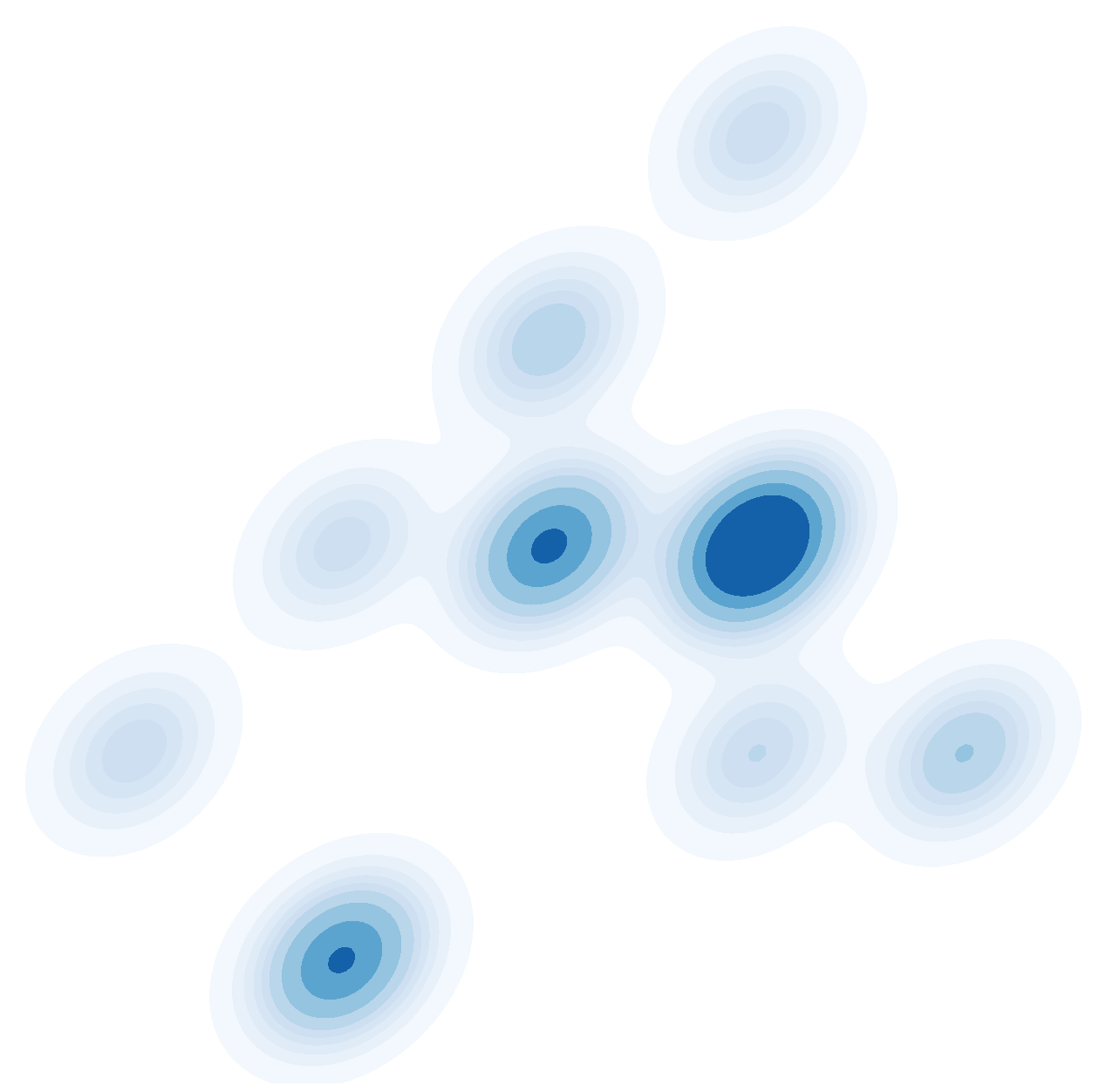} %
        \subcaption{Posterior}
        \label{fig:2d_posterior}
    \end{minipage}
    \hfill
    \begin{minipage}{0.16\textwidth}
        \centering
        \includegraphics[width=\textwidth]{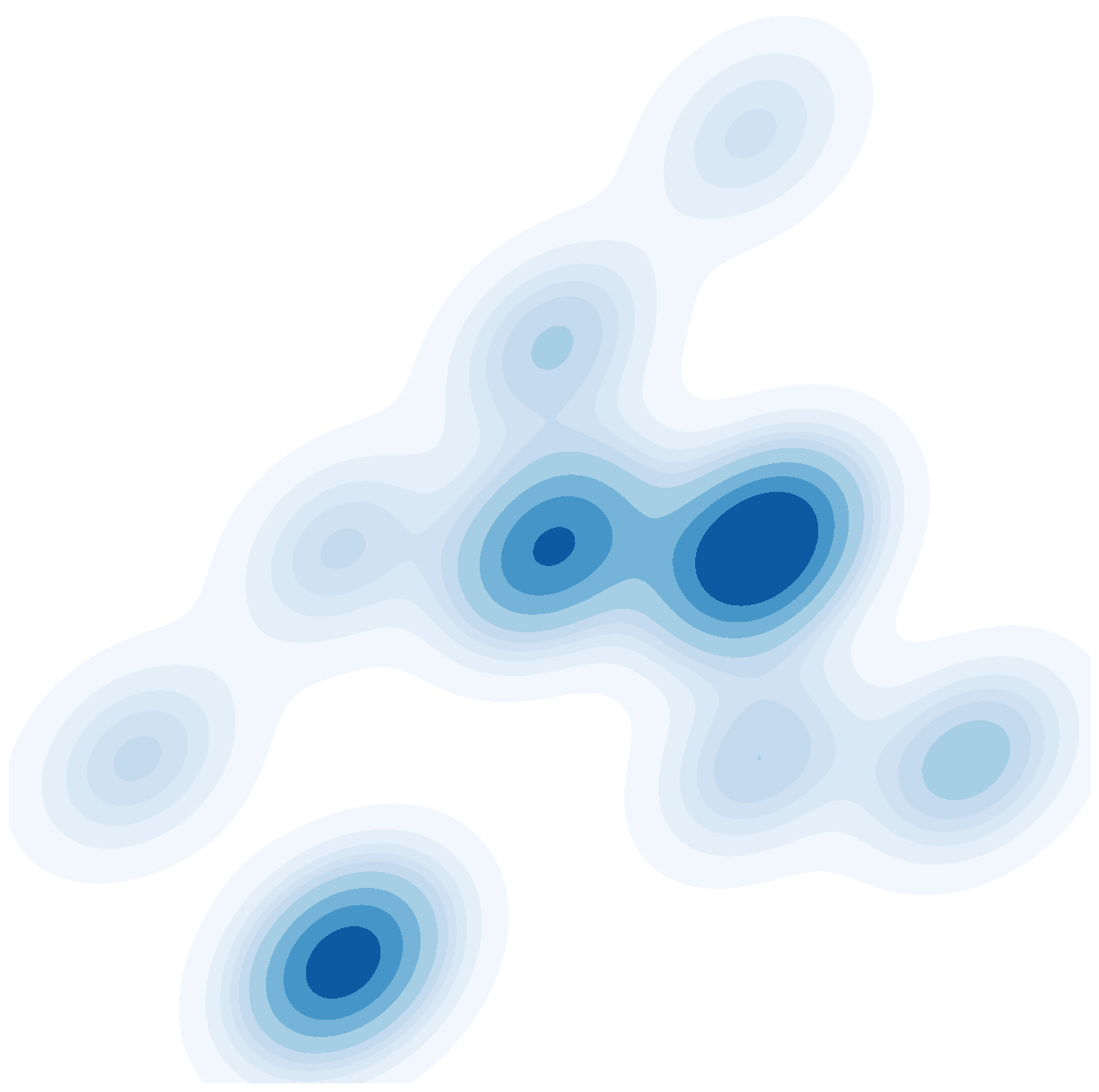} %
        \subcaption{RTB (ours)}
        \label{fig:2d_rtb}
    \end{minipage}\hfill
    \begin{minipage}{0.16\textwidth}
        \centering
        \includegraphics[width=\textwidth]{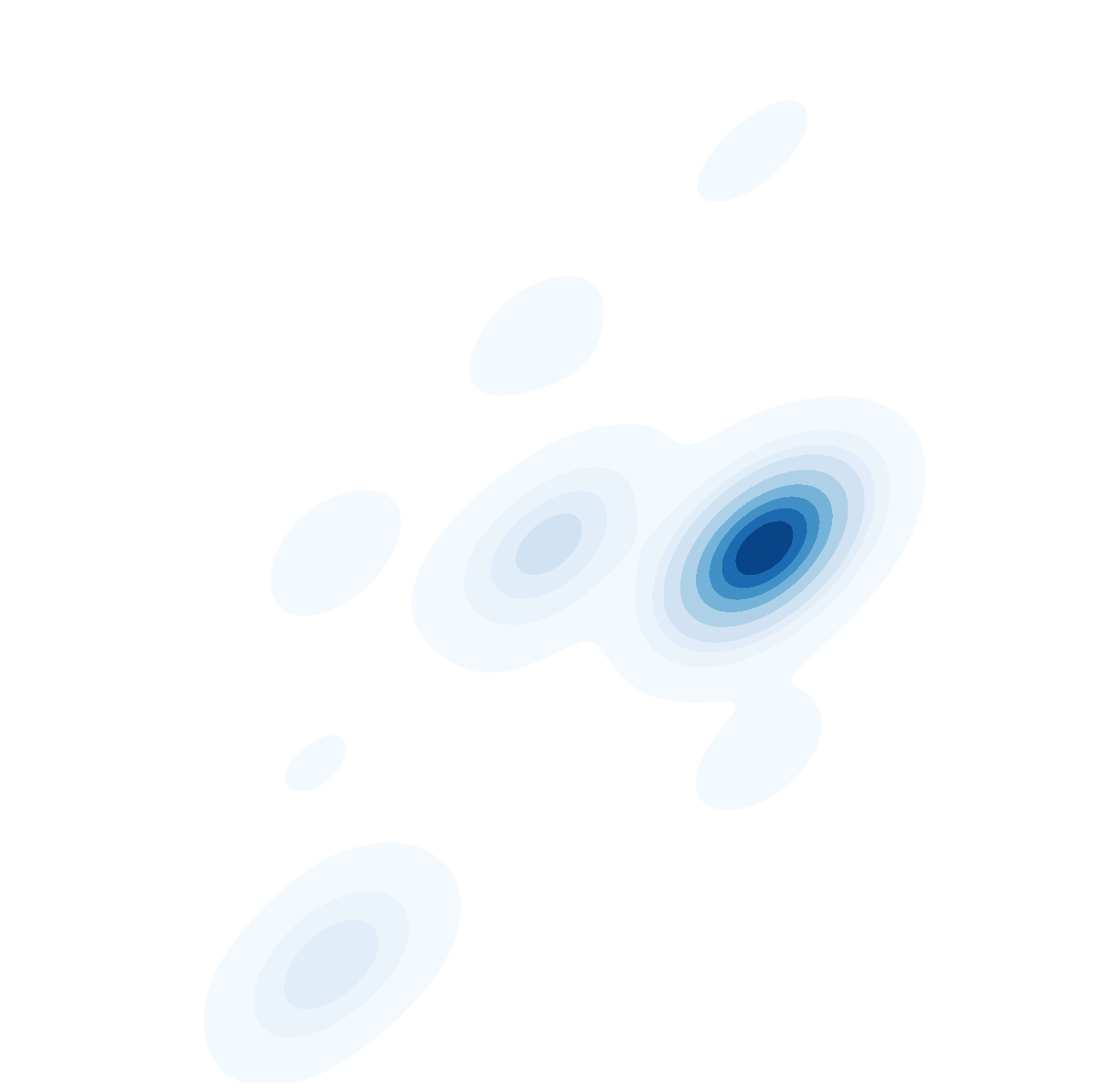} 
        \subcaption{RL (KL reg.)}
        \label{fig:2d_rl_kl}
    \end{minipage}\hfill
    \begin{minipage}{0.16\textwidth}
        \centering
        \includegraphics[width=\textwidth]{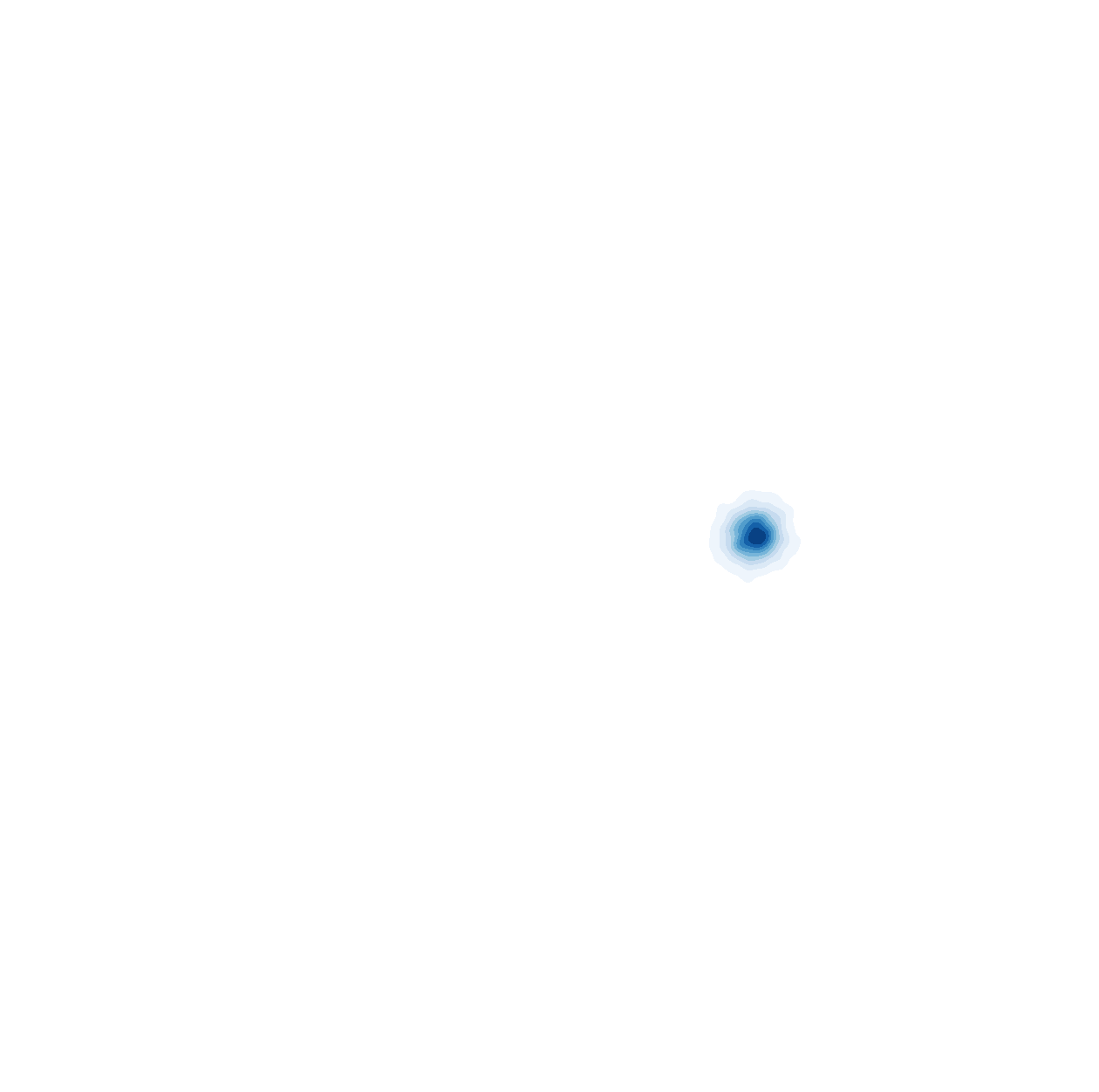} 
        \subcaption{RL (no reg.)}
        \label{fig:2d_no_kl}
    \end{minipage}\hfill
    \begin{minipage}{0.16\textwidth}
        \centering
        \includegraphics[width=\textwidth]{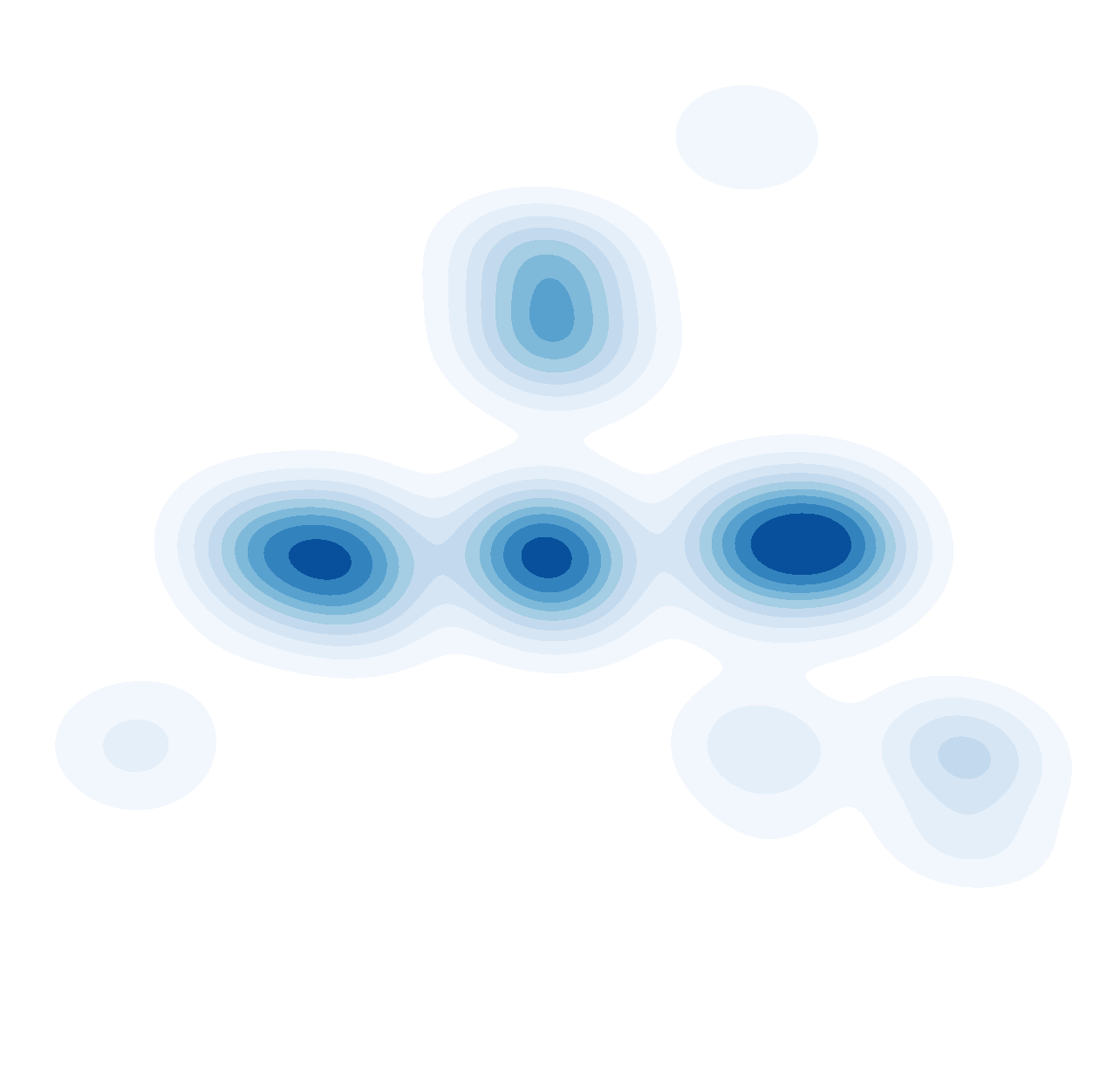} 
        \subcaption{CG}
        \label{fig:2d_classifier}
    \end{minipage}
    \caption{Sampling densities learned by various posterior inference methods. The prior is a diffusion model sampling a mixture of 25 Gaussians (a) and the posterior is the product of the prior with a constraint that masks all but 9 of the modes (b). Our method (RTB) samples close to the true posterior (c). RL methods with tuned KL regularization yield inaccurate inference (d), while without KL regularization, they mode-collapse (e). A classifier guidance (CG) approximation (f) results in biased outcomes. For details, see~\autoref{app:2d_gmm}.
}
    \label{fig:2d_gmm}
\end{figure}

The hierarchical nature of the generative process in diffusion models, which generate samples from $p(\rvx)$ by a deep chain of stochastic transformations, makes exact sampling from posteriors $p(\rvx)r(\rvx)$ under a black-box function $r(\rvx)$ intractable. Common solutions to this problem involve inference techniques based on linear approximations \cite{song2022solving,kawar2021snips,kadkhodaie2021solving,chung2023diffusion} or stochastic optimization \cite{graikos2022diffusion,mardani2024variational}. Others estimate the `guidance' term -- the difference in drift functions between the diffusion models sampling the prior and posterior -- by training a classifier on noised data \cite{dhariwal2021diffusion}, but when such data is not available, one must resort to approximations or Monte Carlo estimates \citep{song2023loss,dou2024diffusion,cardoso2024montecarlo}, which are challenging to scale to high-dimensional problems. Reinforcement learning methods that have recently been proposed for this problem \cite{black2024training, fan2023reinforcement} are biased and prone to mode collapse (\autoref{fig:2d_gmm}).

\paragraph{Contributions.} \looseness=-1 Inspired by recent techniques in training diffusion models to sample distributions defined by unnormalized densities \cite{zhang2021path,richter2023improved,vargas2023denoising,sendera2024diffusion}, we propose an asymptotically unbiased training objective, called relative trajectory balance (RTB), for training diffusion models that sample from posterior distributions under a diffusion model prior (\autoref{sec:intractable_posterior}). RTB is derived from the perspective of diffusion models as continuous generative flow networks \cite{lahlou2023theory}. This perspective also allows us to freely leverage off-policy training, when data with high density under the posterior is available (\autoref{sec:training}). RTB can be applied to iterative generative processes beyond standard diffusion models: our methods generalize to discrete diffusion models and extend existing methods for autoregressive language models (\autoref{sec:gflownet}).

\begin{table}[t]
    \vspace*{-1em}
    \caption{Sources of diffusion priors and constraints.}\label{tab:examples}%
    \resizebox{1\linewidth}{!}{
        \begin{tabular}{@{}llll}\toprule
            Domain & Prior $p(\rvx)$ & Constraint $r(\rvx)$ & Posterior \\\midrule
            Conditional image generation (\autoref{sec:experiments:cls_guidance})
            & Image diffusion model $p(\rvx)$
            & Classifier likelihood $p(c\mid\rvx)$
            & Class-conditional distribution $p(\rvx\mid c)$\\
            Text-to-image generation (\autoref{sec:experiments:text2image})
            & Text-to-image foundation model
            & RLHF reward model
            & Aligned text-to-image model\\
            Language infilling (\autoref{sec:experiments:language_infilling})
            & Discrete diffusion model
            & Autoregressive completion likelihood
            & Infilling distribution\\
            Offline RL policy extraction (\autoref{sec:experiments:rl_offline})
            & Diffusion model as behavior policy
            & Boltzmann dist.\ of $Q$-function
            & Optimal KL-constrained policy\\
            \bottomrule
        \end{tabular}
    }
\end{table}

Our experiments demonstrate the versatility of our approach in a variety of domains:
\begin{itemize}[nosep,left=0pt]
\item In \textbf{vision}, we show that RTB achieves competitive classifier-guided image generation for unconditional diffusion vision priors (\autoref{sec:experiments:cls_guidance}) and can be used to improve caption-conditioned generation under text-to-image foundation model priors (\autoref{sec:experiments:text2image}). 
\item In \textbf{language modeling}, we report strong results for infilling tasks with discrete diffusion language models (\autoref{sec:experiments:language_infilling}). 
\item Finally, we show that RTB achieves state-of-the-art results on \textbf{continuous control} benchmarks that leverage score-based behavior priors (\autoref{sec:experiments:rl_offline}). 
\end{itemize}

\section{Learning posterior samplers with diffusion priors}
\label{sec:methods}

We consider the problem of posterior inference under a prior given by a hierarchical generative model. In this section, we present the mathematical setting (\autoref{sec:diffusion_prior}), our proposed RTB objective (\autoref{sec:intractable_posterior}), and training methods for RTB (\autoref{sec:training}). We will first discuss the case of a diffusion prior over $\R^d$, and later discuss how the methods generalize to arbitrary hierarchical priors (\autoref{sec:gflownet}).

\subsection{Background and setting: Diffusion models as hierarchical generative models}
\label{sec:diffusion_prior}

A denoising diffusion model generates data $\rvx_1$ by a Markovian generative process:
\begin{equation}\label{eq:diffusion_model}
    \text{\it (noise)}\quad \rvx_0\rightarrow\rvx_{\Delta t}\rightarrow\rvx_{2\Delta t}\rightarrow\ldots\rightarrow\rvx_1=\rvx\quad\text{\it (data)},
\end{equation}
where $\Delta t=\frac1T$ and $T$ is the number of discretization steps.\footnote{\looseness=-1 The time indexing suggestive of an SDE discretization is used for consistency with the diffusion samplers literature \cite{zhang2021path,sendera2024diffusion}. The indexing $\rvx_T\rightarrow\rvx_{T-1}\rightarrow\dots\rightarrow\rvx_0$ is often used for diffusion models trained from data.} The initial distribution $p(\rvx_0)$ is fixed (typically to $\gN(\boldsymbol{0},\boldsymbol{I})$) and the transition from $\rvx_{t-1}$ to $\rvx_t$ is modeled as a Gaussian perturbation with time-dependent variance:
\begin{equation}\label{eq:diffusion_transition}
    p(\rvx_{t+\Delta t}\mid\rvx_t)=\gN(\rvx_{t+\Delta t}\mid\rvx_t+u_t(\rvx_t)\Delta t,\sigma_t^2\Delta t\boldsymbol{I}).
\end{equation}
The scaling of the mean and variance by $\Delta t$ is insubstantial for fixed $T$, but ensures that the diffusion process is well-defined in the limit $T\to\infty$ assuming regularity conditions on $u_t$ \cite{oksendal,applied-sde}. The process given by (\ref{eq:diffusion_model}, \ref{eq:diffusion_transition}) is then identical to Euler-Maruyama integration of the stochastic differential equation (SDE) $d\rvx_t=u_t(\rvx_t)\,dt+\sigma_t\,d\rvw_t$.

The likelihood of a denoising trajectory $\rvx_0\rightarrow\rvx_{\Delta t}\rightarrow\dots\rightarrow\rvx_1$ factors as
\begin{equation}\label{eq:diffusion_likelihood}
    p(\rvx_0,\rvx_{\Delta t},\dots,\rvx_1)=p(\rvx_0)\prod_{i=1}^T p(\rvx_{i\Delta t}\mid\rvx_{(i-1)\Delta t})
\end{equation}
and defines a marginal density over the data space:
\begin{equation}\label{eq:diffusion_marginal}
    p(\rvx_1)=\int p(\rvx_0,\rvx_{\Delta t},\dots,\rvx_1)\,d\rvx_0\,d\rvx_{\Delta t}\ldots d\rvx_{1-\Delta t}.
\end{equation}
A reverse-time process, $\rvx_1\rightarrow\rvx_{1-\Delta t}\rightarrow\dots\rightarrow\rvx_0$, with densities $q$, can be defined analogously, and similarly defines a conditional density over trajectories:
\begin{equation}\label{eq:diffusion_likelihood_reverse}
    q(\rvx_0,\rvx_{\Delta t},\dots,\rvx_{1-\Delta t}\mid \rvx_1)=\prod_{i=1}^T q(\rvx_{(i-1)\Delta t}\mid\rvx_{i\Delta t}).
\end{equation}
In the training of diffusion models, as discussed below, the process $q$ is typically fixed to a simple distribution (usually a discretized Ornstein-Uhlenbeck process), and the result of training is that $p$ and $q$ are close as distributions over trajectories.

\paragraph{Diffusion model training as divergence minimization.}
Diffusion models parametrize the drift $u_t(\rvx_t)$ in (\autoref{eq:diffusion_transition}) as a neural network $u(\rvx_t,t;\theta)$ with parameters $\theta$ and taking $\rvx_t$ and $t$ as input. We denote the distributions over trajectories induced by (\autoref{eq:diffusion_likelihood}, \autoref{eq:diffusion_marginal}) by $p_\theta$ to show their dependence on the parameter. 

In the most common setting, diffusion models are trained to maximize the likelihood of a dataset. In the notation above, this corresponds to assuming $q(\rvx_1)$ is fixed to an empirical measure (with the points of a training dataset $\gD$ assumed to be i.i.d.\ samples from $q(\rvx_1)$). Training minimizes with respect to $\theta$ the divergence between the processes $q$ and $p_\theta$:
\begin{align}\label{eq:diffusion_kl_objective}
    &\KL(q(\rvx_0,\rvx_{\Delta t},\dots,\rvx_1)\,\|\,p_\theta(\rvx_0,\rvx_{\Delta t},\dots,\rvx_1))\\
    &=\KL(q(\rvx_1)\,\|\,p_\theta(\rvx_1))+\sE_{\rvx_1\sim q(\rvx_1)}\KL(q(\rvx_0,\rvx_{\Delta t},\dots,\rvx_{1-\Delta t}\mid \rvx_1)\,\|\,p_\theta(\rvx_0,\rvx_{\Delta t},\dots,\rvx_{1-\Delta t}\mid \rvx_1))
    \nonumber\\&\geq \KL(q(\rvx_1)\,\|\,p_\theta(\rvx_1))
    =\sE_{\rvx_1\sim q(\rvx_1)}[-\log p_\theta(\rvx_1)]+\text{\rm const}.\nonumber
\end{align}
where the inequality -- an instance of the data processing inequality for the KL divergence -- shows that minimizing the divergence between distributions over trajectories is equivalent to maximizing a lower bound on the data log-likelihood under the model $p_\theta$. 

As shown in \cite{song2021maximum}, minimization of the KL in (\autoref{eq:diffusion_kl_objective}) is essentially equivalent to the traditional approach to training diffusion models via denoising score matching \cite{vincent2011connection,sohl2015diffusion,ho2020ddpm}. Such training exploits that for typical choices of the noising process $q$, the optimal $u_t(\rvx_t)$ can be expressed in terms of the Stein score of $q(\rvx_1)$ convolved with a Gaussian, allowing an efficient stochastic regression objective for $u_t$. For full generality of our exposition for arbitrary iterative generative processes, we prefer to think of (\autoref{eq:diffusion_kl_objective}) as the primal objective and denoising score matching as an efficient means of minimizing it.

\paragraph{Trajectory balance and distribution-matching training.} From (\autoref{eq:diffusion_kl_objective}) we also see that the bound is tight if the conditionals of $p_\theta$ and $q$ on $\rvx_1$ coincide, \ie, $q$ is equal to the posterior distribution of $p$ conditioned on $\rvx_1$. Indeed, the model $p_\theta$ minimizes (\autoref{eq:diffusion_kl_objective}) for a distribution with continuous density $q(\rvx_1)$ if and only if, for all denoising trajectories,
\begin{equation}\label{eq:tb}
    p_\theta(\rvx_0,\rvx_{\Delta t},\dots,\rvx_1)=q(\rvx_1)q(\rvx_0,\rvx_{\Delta t},\dots,\rvx_{1-\Delta t}\mid\rvx_1).
\end{equation}
This was named the \emph{trajectory balance (TB) constraint} by \cite{lahlou2023theory} -- by analogy with a constraint for discrete-space iterative sampling \cite{malkin2022trajectory} -- and is a time-discretized version of a constraint used for enforcing equality of continuous-\emph{time} path space measures in \cite{nusken2021solving} (see \cite{berner2025discrete} for asymptotic analysis).

In \cite{richter2020vargrad,lahlou2023theory}, the constraint (\ref{eq:tb}) was used for the training of diffusion models in a \emph{data-free} setting, where instead of i.i.d.\ samples from $q(\rvx_1)$ one has access to a (possibly unnormalized) density $q(\rvx_1)=e^{-\gE(\rvx_1)}/Z$ from which one wishes to sample. These objectives minimize the squared log-ratio between the two sides of (\ref{eq:tb}), which allows the trajectories $\rvx_0\rightarrow \rvx_{\Delta t}\rightarrow\dots\rightarrow\rvx_1$ used for training to be sampled from any training distribution, such as `exploratory' modifications of $p_\theta$ or trajectories found by local search (MCMC) in the target space. The flexibility of off-policy exploration that this allows was studied by \cite{sendera2024diffusion}. Such objectives contrast with on-policy, simulation-based approaches that require differentiating through the sampling process \cite[\eg,][]{zhang2021path,vargas2023denoising,berner2022optimal,vargas2024transport}.

\subsection{Intractable inference under diffusion priors}
\label{sec:intractable_posterior}

Consider a diffusion model $p_\theta$, defining a marginal density $p_\theta(\rvx_1)$, and a positive constraint function $r:\R^d\to\R_{>0}$. We are interested in training a diffusion model $p_\phi\post$, with drift function $u_\phi\post$, that would sample the product distribution $p\post(\rvx_1)\propto p_\theta(\rvx_1)r(\rvx_1)$. If $r(\rvx_1)=p(\rvy\mid\rvx_1)$ is a conditional distribution over another variable $\rvy$, then $p\post$ is the Bayesian posterior $p_\theta(\rvx_1\mid\rvy)$.%

\looseness=-1
Because samples from $p\post(\rvx_1)$ are not assumed to be available, one cannot directly train $p$ using the objective (\ref{eq:diffusion_kl_objective}). Nor can one directly apply objectives for distribution-matching training, such as those that enforce (\ref{eq:tb}), since the marginal $p_\theta(\rvx_1)$ is not available. However, we make the following observation (proof in \autoref{sec:proofs}).
\begin{propositionE}[Relative TB constraint][end,restate]\label{prop:rtb}
    If $p_\theta$, $p_\phi\post$, and the scalar $Z_\phi$ jointly satisfy the \emph{relative trajectory balance (RTB) constraint}
    \begin{equation}\label{eq:rtb}
        Z_\phi\cdot p_\phi\post(\rvx_0,\rvx_{\Delta t},\dots,\rvx_1)=r(\rvx_1)p_\theta(\rvx_0,\rvx_{\Delta t},\dots,\rvx_1)
    \end{equation}
    for every denoising trajectory $\rvx_0\rightarrow\rvx_{\Delta t}\rightarrow\dots\rightarrow \rvx_1$, then $p_\phi\post(\rvx_1)\propto p_\theta(\rvx_1)r(\rvx_1)$, \ie, the diffusion model $p_\phi\post$ samples the posterior distribution.
    Furthermore, if $p_\theta$ also satisfies the TB constraint (\ref{eq:tb}) with respect to the noising process $q$ and some target density $q(\rvx_1)$, then $p_\phi\post$ satisfies the TB constraint with respect to the target density $q\post(\rvx_1)\propto q(\rvx_1)r(\rvx_1)$, and $Z=\int q(\rvx_1)r(\rvx_1)\,d\rvx_1$.
\end{propositionE}
\begin{proofE}
    Suppose that $p_\theta$, $p_\phi\post$, and $Z$ jointly satisfy (\ref{eq:rtb}). Then necessarily $Z\neq0$, since the quantities on the right side are positive. We then have, using (\ref{eq:diffusion_marginal}),
    \begin{align*}
        p_\phi\post(\rvx_1)
        &=
        \int p_\phi\post(\rvx_0,\rvx_{\Delta t},\dots,\rvx_1)\,d\rvx_0\,d\rvx_{\Delta t}\dots d\rvx_{1-\Delta t}\\
        &=\frac1Zr(\rvx_1)\int p_\theta(\rvx_0,\rvx_{\Delta t},\dots,\rvx_1)\,d\rvx_0\,d\rvx_{\Delta t}\dots d\rvx_{1-\Delta t}\\
        &=\frac1Zr(\rvx_1)p_\theta(\rvx_1)
        &\propto p_\theta(\rvx_1)r(\rvx_1),
    \end{align*}
    as desired.

    Now suppose that $p_\theta$ also satisfies the TB constraint (\ref{eq:tb}) with respect to $q(\rvx_1)$. Then, for any denoising trajectory,
    \begin{equation}
        q(\rvx_0,\rvx_{\Delta t},\dots,\rvx_{1-\Delta t}\mid\rvx_1)
        =\frac{p_\theta(\rvx_0,\rvx_{\Delta t},\dots,\rvx_1)}{q(\rvx_1)}\\
        =\frac{p_\phi\post(\rvx_0,\rvx_{\Delta t},\dots,\rvx_1)}{q(\rvx_1)r(\rvx_1)/Z}.\label{eq:rtb_implies_tb}
    \end{equation}
    showing that $p_\phi\post$ satisfies the TB constraint with respect to the noising process $q$ and the (not yet shown to be normalized) density $\frac1Zq(\rvx_1)r(\rvx_1)$. We integrate out the variables $\rvx_0,\rvx_{\Delta t},\dots,\rvx_{1-\Delta t}$ in (\ref{eq:rtb_implies_tb}), giving
    \begin{align*}
        1&=\frac{p_\phi\post(\rvx_1)}{q(\rvx_1)r(\rvx_1)/Z}\\
        q(\rvx_1)r(\rvx_1)&=Zp_\phi\post(\rvx_1).
    \end{align*}
    Integrating over $\rvx_1$ shows $\int q(\rvx_1)r(\rvx_1)\,d\rvx_1=Z$.
\end{proofE}
Note that the two joints appearing in (\ref{eq:rtb}) are defined as products over transitions, via (\ref{eq:diffusion_likelihood}).

\paragraph{Relative trajectory balance as a loss.} Analogously to the conversion of the TB constraint (\ref{eq:tb}) into a trajectory-dependent training objective in \cite{malkin2022trajectory,lahlou2023theory}, we define the \emph{relative trajectory balance loss} as the discrepancy between the two sides of (\ref{eq:rtb}), seen as a function of the vector $\phi$ that parametrizes the posterior diffusion model and the scalar $Z_\phi$ (parametrized via $\log Z_\phi$ for numerical stability):
\begin{equation}\label{eq:rtb_objective}
    \LRTB(\rvx_0\rightarrow\rvx_{\Delta t}\rightarrow\dots\rightarrow\rvx_1;\phi)
    :=
    \left(\log\frac
        {Z_\phi\cdot p_\phi\post(\rvx_0,\rvx_{\Delta t},\dots,\rvx_1)}{r(\rvx_1)p_\theta(\rvx_0,\rvx_{\Delta t},\dots,\rvx_1)}
    \right)^2.
\end{equation}
Optimizing this objective to 0 for all trajectories ensures that (\ref{eq:rtb}) is satisfied. While the RTB constraint (\ref{eq:rtb}) has a similar form to TB (\ref{eq:tb}), RTB involves the ratio of two denoising processes, while TB involves the ratio of a forward and a backward process. However, the name `relative TB' is justified by interpreting the densities in a TB constraint relative to a measure defined by the prior model; see \autoref{sec:gflownet}.

If we assume $p_\theta(\rvx_0)=p_\phi\post(\rvx_0)$ are fixed (\eg, to a standard normal), then (\ref{eq:rtb_objective}) reduces to
\begin{equation}\label{eq:rtb_termwise}
    \left(\log\frac{Z_\phi}{r(\rvx_1)}+
        \sum_{i=1}^T\log\frac{p_\phi\post(\rvx_{i\Delta t}\mid\rvx_{(i-1)\Delta t})}{p_\theta(\rvx_{i\Delta t}\mid\rvx_{(i-1)\Delta t})}
    \right)^2.
\end{equation}

Notably, the gradient of this objective with respect to $\phi$ does not require differentiation (backpropagation) into the sampling process that produced a trajectory $\rvx_0\rightarrow\dots\rightarrow\rvx_1$. This offers two advantages over on-policy simulation-based methods: (1) the ability to optimize $\LRTB$ as an off-policy objective, \ie, sampling trajectories for training from a distribution different from $p_\phi\post$ itself, as discussed further in \autoref{sec:training}; (2) backpropagating only to a subset of the summands in (\ref{eq:rtb_termwise}), when computing and storing gradients for all steps in the trajectory is prohibitive for large diffusion models (see \autoref{app:efficient}).

\paragraph{Comparison with classifier guidance.} It is interesting to contrast the RTB training objective with the technique of \emph{classifier guidance} \cite{dhariwal2021diffusion} used for some problems of the same form. If ${r(\rvx_1)=p(\rvy\mid\rvx_1)}$ is a conditional likelihood, classifier guidance relies upon writing $u_t(\rvx_t)-u_t\post(\rvx_t)$ explicitly in terms of $\nabla_{\rvx_t}\log p(\rvy\mid\rvx_t)$, by combining the expression of the optimal drift $u_t$ in terms of the score of the target distribution convolved with a Gaussian (cf.\ \autoref{sec:diffusion_prior}), with the `Bayes' rule' for the Stein score: $\nabla_{\rvx_t}\log p(\rvx_t\mid\rvy)=\nabla_{\rvx_t}\log p(\rvx_t)+\nabla_{\rvx_t}\log p(\rvy\mid\rvx_t)$.

\looseness=-1
Classifier guidance gives the \emph{exact} solution for the posterior drift when a differentiable classifier on noisy data, $p(\rvy\mid\rvx_t)=\int p(\rvy\mid\rvx_1)p(\rvx_1\mid\rvx_t)\,d\rvx_1$, is available. Unfortunately, such a classifier is not, in general, tractable to derive from the classifier on noiseless data, $p(\rvy\mid\rvx_1)$, and cannot be learned without access to unbiased data samples. RTB is an asymptotically unbiased objective that recovers the difference in drifts (and thus the gradient of the log-convolved likelihood) in a data-free manner.

\subsection{Training, parametrization, and conditioning} 
\label{sec:training}

\paragraph{Training and exploration.}
The choice of which trajectories we use to take gradient steps with the RTB loss can have a large impact on sample efficiency. In \emph{on-policy} training, we use the current policy $p_\phi\post$ to generate trajectories $\tau =(\rvx_0\rightarrow\ldots\rightarrow\rvx_1)$, evaluate the reward $\log r(\rvx_1)$ and the likelihood of $\tau$ under $p_\theta$, and a gradient updates on $\phi$  to minimize $\LRTB(\tau;\phi)$.

However, on-policy training may be insufficient to discover the modes of the posterior distribution.
In this case, we can perform \emph{off-policy} exploration to ensure mode coverage. For instance, given samples $\rvx_1$ that have high density under the target distribution, we can sample \emph{noising} trajectories $\rvx_1 \leftarrow \rvx_{1-\Delta t} \leftarrow \ldots \leftarrow \rvx_0$ starting from these samples and use such trajectories for training. Another effective off-policy training technique uses replay buffers. We expect the flexibility of mixing on-policy training with off-policy exploration to be a strength of RTB over on-policy RL methods, as was shown for distribution-matching training of diffusion models in \cite{sendera2024diffusion}.

\paragraph{Conditional constraints and amortization.}
Above we derived and proved the correctness of the RTB objective for an arbitrary positive constraint $r(\rvx_1)$. If the constraints depend on other variables $\rvy$ -- for example, $r(\rvx_1;\rvy)=p(\rvy\mid\rvx_1)$ -- then the posterior drift $u_\phi\post$ can be conditioned on $\rvy$ and the learned scalar $\log Z_\phi$ replaced by a model taking $\rvy$ as input. Such conditioning achieves amortized inference and allows generalization to new $\rvy$ not seen in training. Similarly, all of the preceding discussion easily generalizes to \emph{priors} that are conditioned on some context variable.%

\paragraph{Efficient parametrization and Langevin inductive bias.} Because the deep features learned by the prior model $u_\theta$ are expected to be useful in expressing the posterior drift $u_\phi\post$, we can choose to initialize $u_\phi\post$ as a copy of $u_\theta$ and to fine-tune it, possibly in a parameter-efficient way (as described in each section of \autoref{sec:experiments}). This choice is inspired by the method of amortizing inference in large language models by fine-tuning a prior model to sample an intractable posterior \cite{hu2023amortizing}.

Furthermore, if the constraint $r(\rvx_1)$ is differentiable, we can impose an inductive bias on the posterior drift similar to the one introduced for diffusion samplers of unnormalized target densities in \cite{zhang2021path} and shown to be useful for off-policy methods in \cite{sendera2024diffusion}. namely, we write
\begin{equation}\label{eq:langevin}
    u_{\phi}\post(\rvx_t, t) = \text{NN}_1(\rvx_t, t; \phi) + \text{NN}_2(\rvx_t, t, \phi) \nabla_{\rvx_t}\log r(\rvx_t),
\end{equation}
where $\text{NN}_1$ and $\text{NN}_2$ are neural networks outputting a vector and a scalar, respectively. This parametrization allows the constraint to provide a signal to guide the sampler at intermediate steps. %

\paragraph{Stabilizing the loss.} We propose two simple design choices for stabilizing RTB training. First, the loss in (\ref{eq:rtb_objective}) can be replaced by the empirical \emph{variance} over a minibatch of the quantity inside the square, which removes dependence on $\log Z_\phi$ and is especially useful in conditional settings, consistent with the findings of \cite{sendera2024diffusion}. This amounts to a relative variant of the VarGrad objective~\citep{richter2020vargrad} (see (\ref{eq:vargrad_rtb}) in \autoref{app:offline_rl}). Second, we employ loss clipping: to reduce sensitivity to an imperfectly fit prior model, we do not perform updates on trajectories where the loss is close to 0 (see \autoref{app:cls_rtb},\autoref{app:infilling}).

\subsection{Generative flow networks and extension to other hierarchical processes}
\label{sec:gflownet}

\paragraph{RTB as TB under the prior measure.}
The theoretical foundations for continuous generative flow networks \cite{lahlou2023theory} establish the correctness of enforcing constraints such as trajectory balance (\ref{eq:tb}) for training sequential samplers, such as diffusion models, to match unnormalized target densities. While we have considered Gaussian transitions and identified transition kernels with their densities with respect to the Lebesgue measure over $\R^d$, these foundations generalize to more general \emph{reference measures}. In \autoref{app:relative}, we show how the RTB constraint can be recovered as a special case of the TB constraint for a certain choice of reference measure derived from the prior.

\paragraph{Extension to arbitrary sequential generation.}

While our discussion was focused on diffusion models for continuous spaces, the RTB objective can be applied to any Markovian sequential generative process, in particular, one that can be formulated as a generative flow network in the sense of \cite{bengio2023gflownet,lahlou2023theory}. This includes, in particular, generative models that generate objects by a sequence of discrete steps, including autoregressive models and discrete diffusion models. In the case of discrete diffusion, where the intermediate latent variables $\rvx_t$ lie not in $\R^d$ but in the space of sequences, one simply replaces the Gaussian transition densities by transition probability \emph{masses} in the RTB constraint (\ref{eq:rtb}) and objective (\ref{eq:rtb_objective}). In the case of autoregressive models, where only one sequence of steps can generate any given object, the backward process $q$ becomes trivial, and the RTB constraint for a model $p_\phi\post$ to sample a sequence $\rvx$ from a distribution with density $r(\rvx)p_\theta(\rvx)$ is simply $Z_\phi p_\phi\post(\rvx)=r(\rvx)p_\theta(\rvx)$ for all sequences $\rvx$. We note that a sub-trajectory generalization of this objective was used in \cite{hu2023amortizing} to amortize intractable inference in autoregressive language models.

\section{Experiments} \label{sec:experiments}

In this section, we present empirical results to validate the efficacy of relative trajectory balance. Our experiments are designed to demonstrate the wide applicability of RTB to sample from posteriors for diffusion priors with arbitrary rewards on vision, language, and continuous control tasks.%

\begin{table}[t]
\vspace*{-1em}
\caption{\looseness=-1 Classifier-guided posterior sampling with pretrained unconditional diffusion priors. We report the mean\std{\rm std} of each metric computed across all relevant classes for each experiment set, and highlight $\pm$5\% from highest/lower experimental value. The FID is computed between learned posterior samples and the true samples from the class in question.  DP and LGD-MC fail to appropriately model the posterior distribution (high average $\log r(\rvx)$) while DDPO mode-collapses. RTB achieves comparable or superior performance to all other baselines, optimally balancing high reward and diversity as measured by FID. See \autoref{tab:conditional_mnist} for conditional variants.} 
\centering
\resizebox{1\linewidth}{!}{
\begin{tabular}{@{}lccccccccc}\toprule
Dataset $\rightarrow$ & \multicolumn{3}{c}{MNIST} & \multicolumn{3}{c}{MNIST even/odd} & \multicolumn{3}{c}{CIFAR-10}\\\cmidrule(lr){2-4}\cmidrule(lr){5-7}\cmidrule(lr){8-10} Algorithm $\downarrow$ Metric $\rightarrow$ 
        & $\sE[\log r(\rvx)]$ ($\uparrow$)       & FID ($\downarrow$)       & Diversity ($\uparrow$)             & $\sE[\log  r(\rvx)]$ ($\uparrow$)     & FID ($\downarrow$)   & Diversity ($\uparrow$)     & $\sE[\log  r(\rvx)]$ ($\uparrow$) & FID ($\downarrow$) & Diversity ($\uparrow$)\\\midrule

DPS    & $-2.1597$\std{$0.423$}      & $1.2913$\std{$0.410$}       & \highlight{$0.1609$\std{$0.000$}} & $-1.2270$\std{$0.202$} & \highlight{$1.1498$\std{$0.182$}}   & \highlight{$0.1713$\std{$0.000$}}& $-3.6025$\std{$0.503$} & $0.7371$\std{$0.216$}       & \highlight{$0.2738$\std{$0.000$}}       \\
LGD$-$MC   & $-2.1389$\std{$0.480$}      & $1.2873$\std{$0.412$}      & \highlight{$0.1600$\std{$0.000$}} & $-1.1720$\std{$0.199$}   & \highlight{$1.1445$\std{$0.184$}}  & $0.1600$\std{$0.000$}        & $-3.0988$\std{$0.359$} & $0.7402$\std{$0.214$}      & \highlight{$0.2743$\std{$0.000$}}       \\
DDPO    & \highlight{$-1.5$\std{$4.7$}$\times10^{-3}$} & $1.5822$\std{$0.583$}& $0.1350$\std{$0.005$}       & \highlight{$-8.6$\std{$12.3$}$\times10^{-11}$} & $1.8024$\std{$0.423$} & $0.1314$\std{$0.002$}   & \highlight{$-2.7$\std{$8.5$}$\times10^{-4}$} & $1.7686$\std{$0.589$} & $0.1575$\std{$0.015$}      \\
DPOK    & $-0.1379$\std{$0.225$}      & \highlight{$1.2063$\std{$0.316$}}       & $0.1442$\std{$0.004$}       & $-0.0783$\std{$0.082$} & $1.2536$\std{$0.206$}   & \highlight{$0.1631$\std{$0.007$}} & $-2.4414$\std{$3.266$} & $0.5316$\std{$0.157$}       & $0.2415$\std{$0.024$}      \\
\textbf{RTB (ours)} & $-0.1734$\std{$0.194$}      & \highlight{$1.1823$\std{$0.288$}} & $0.1474$\std{$0.003$}       & $-0.1816$\std{$0.175$} & \highlight{$1.1794$\std{$0.171$}}  & \highlight{$0.1679$\std{$0.004$}} & $-2.1625$\std{$0.879$} & \highlight{$0.4717$\std{$0.138$}} & $0.2440$\std{$0.011$}    \\\bottomrule
\end{tabular}
\label{tab:cls_finetuning_results}
}
\end{table}
\begin{figure}
\vspace*{-1em}
\centering
\includegraphics[width=1.\linewidth]{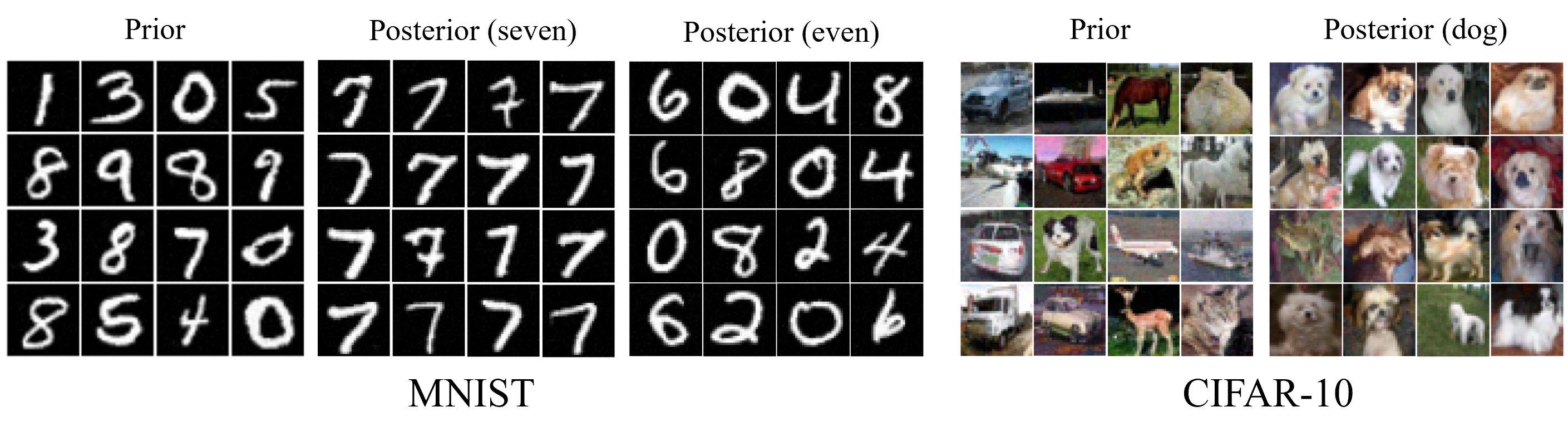}\vspace*{-1em}
\caption{Samples from RTB fine-tuned diffusion posteriors.} 
\label{fig:cls_finetuning_samples}
\vspace{-12pt}
\end{figure}
\subsection{Class-conditional posterior sampling from unconditional diffusion priors} \label{sec:experiments:cls_guidance}

We evaluate RTB in a classifier-guided visual task where we wish to learn a diffusion posterior ${p_\phi\post(\rvx\mid c) \propto p_\theta(\rvx)p(c\mid \rvx)}$ given a pretrained diffusion prior $p_\theta(\rvx)$ and a classifier $r(\rvx)= p(c\mid \rvx)$. 

\paragraph{Setup.} 
We consider two 10-class image datasets, MNIST and CIFAR-10, using off-the-shelf unconditional diffusion priors from~\citep{ho2020ddpm} and standard classifiers $p(c\mid\rvx)$ for both datasets. We perform parameter-efficient fine-tuning of $p_\phi\post$,
initialized as a copy of the prior $p_\theta$, using the RTB objective (see \autoref{app:cls_rtb:experimental_details} for details). The RTB objective is optimized on trajectories sampled on-policy from the current posterior model. We compare RTB with two RL-based fine-tuning techniques derived from DPOK~\citep{fan2023reinforcement} and DDPO~\citep{black2024training}
and with two classifier guidance baselines, namely DPS~\citep{chung2023diffusion}, and LGD-MC~\cite{song2023loss}. We consider three experimental settings: MNIST single-digit posterior (learning to sample images of each digit class $c$), CIFAR-10 single-class posterior (analogous to the previous), and MNIST multi-digit posterior. The latter is a multimodal posterior, for which we set ${r(\rvx)= \max_{i \in \{0,2,4,6,8\}} p(c=i\mid \rvx)}$ to generate even digits, and similarly for odd digits.

\paragraph{Results.} 

Samples from the RTB-fine-tuned posterior models are shown in  \autoref{fig:cls_finetuning_samples}. In \autoref{tab:cls_finetuning_results} we report mean\std{\rm std} of various metrics across all trained posteriors. We observe that models fine-tuned with RTB generate class samples with both the highest diversity (highest mean pairwise cosine distance in Inceptionv3 feature space) and closeness to true samples of the target classes (FID), while achieving high expected $\log r(\rvx)$. Pure RL fine-tuning (no KL regularization) displays mode collapse characteristics, achieving high rewards in exchange for significantly poorer diversity and FID scores (see also \autoref{app:fig:cls_guidance_collapse}). Classifier-guidance-based methods, like DP and LGD-MC, exhibit high diversity, but fail to appropriately model the posterior distribution (lowest $\log r(\rvx)$). Additional results can be found in \autoref{app:cls_rtb:additional_findings}.

\subsection{Fine-tuning a text-to-image diffusion model}\label{sec:experiments:text2image}

\looseness=-1
Diffusion models for text-conditional image generation~\citep[\eg][]{rombach2021high} can struggle to consistently generate images $\rvx$ that adhere to complex prompts $\rvz$, for example, those that involve composing multiple objects (\eg, ``A cat and a dog'') or specify ``unnatural'' appearances (\eg, ``A green-colored rabbit''). 
Fine-tuning pretrained text-to-image diffusion models $p_\theta(\rvx_1\mid\rvz)$ as RL policies to maximize some reward $r(\rvx_1, \rvz)$ based on human preferences has become the standard approach to tackle this issue~\citep{black2024training,fan2023reinforcement,uehara2024fine}. %
Simply maximizing the reward function can result in mode collapse as well as over-optimization of the reward. This is typically handled by constraining the fine-tuned model $\tilde p$ to be close to the prior $p$: %
\begin{equation}\label{eq:t2irlkl}
        \argmax_{\tilde p} \sE_{\tilde p(\rvx_1 \mid  \rvz)}[r(\rvx_1,\rvz)],
        \quad \KL[\tilde p(\rvx_1\mid \rvz) \parallel p(\rvx_1\mid \rvz)] \leq \epsilon.
\end{equation}
The optimal $\tilde p$ for (\ref{eq:t2irlkl}) is $\tilde p(\rvx_1 \mid \rvz) \propto p(\rvx_1 \mid \rvz) \exp(\beta r(\rvx_1,\rvz))$ for some inverse temperature $\beta$. The marginal KL is intractable for diffusion models, so methods like DPOK~\citep{fan2023reinforcement} optimize an upper bound on the marginal KL in the form of a per-step KL penalty $-\gamma\sum_{i=1}^T\KL[\tilde p(\rvx_{i\Delta t} \mid \rvx_{(i-1)\Delta t},\rvz)||p(\rvx_{i\Delta t}\mid \rvx_{(i-1)\Delta t},\rvz)]$ added to the reward. By contrast, RTB can avoid the bias in such an approximation and directly learn to generate unbiased samples from the posterior $\tilde p(\rvx_1\mid\rvz)$. 
\paragraph{Setup.} We demonstrate how RTB can be used to fine-tune pretrained text-to-image diffusion models. We use the latent diffusion model Stable Diffusion v1-5 \cite{rombach2021high} as a prior over $512\times512$ images. Following DPOK~\cite{fan2023reinforcement}, we use ImageReward \citep{xu2023imagereward}, which has been trained to match human preferences as well as prompt accuracy to attributes such as the number of objects, color, and compositionality, as the reward $\log r(\rvx_1, \rvz)$. As reference, we present comparisons against DPOK with the default KL regularization $\gamma = 0.01$ and DPOK with $\gamma=0.0$, which is equivalent to DDPO~\citep{black2024training}. %
We measure the final average reward and the diversity of the generated image, as measured by the average pairwise cosine distance between CLIP embeddings~\citep{radford2021learning}
of a batch of generated images. Further details about the experimental setup and ablations are discussed in \autoref{app:text2image}. %

\begin{wrapfigure}{r}{0.50\textwidth}
    \vspace*{-1em}
    \includegraphics[width=\linewidth]{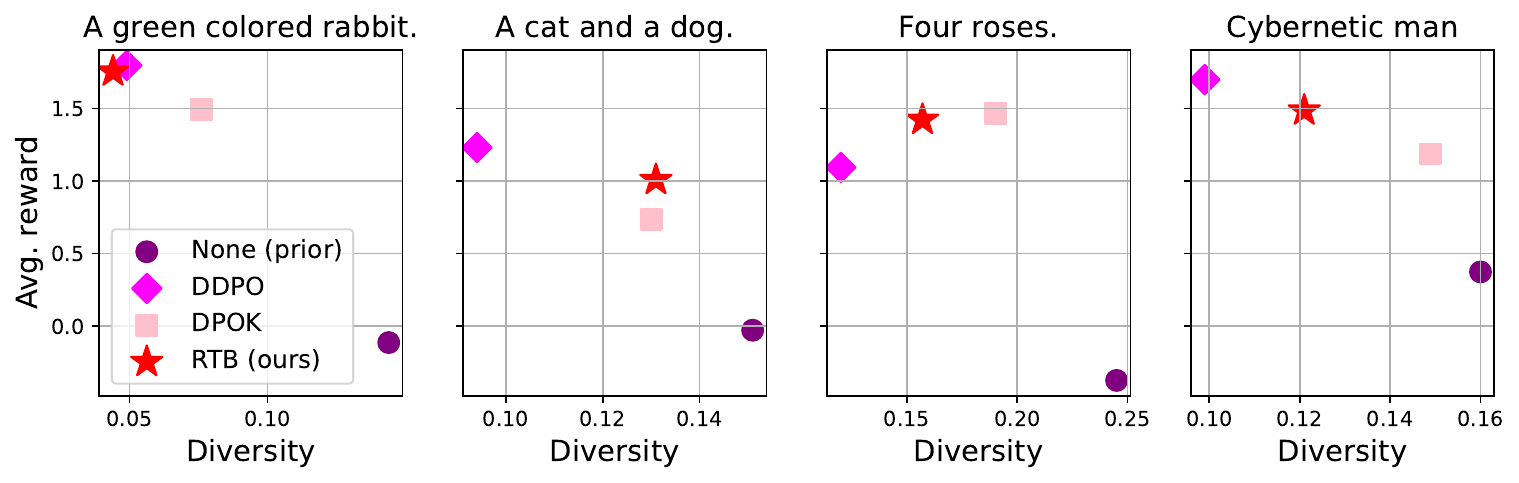}
    \caption{Fine-tuning Stable Diffusion with ImageReward. We report mean $\log r(\rvx_1, \rvz)$ and diversity, measured as the mean cosine distance between CLIP embeddings for a batch of 100 generated images.\protect\footnotemark 
    }
    \label{fig:text2image}
\end{wrapfigure}
\begin{figure}[t]
\vspace*{-1em}
    \centering
    \begin{subfigure}[b]{0.23\textwidth}
        \centering
        \begin{tabular}{@{}c@{}c@{}c@{}}
            \includegraphics[width=0.3\linewidth]{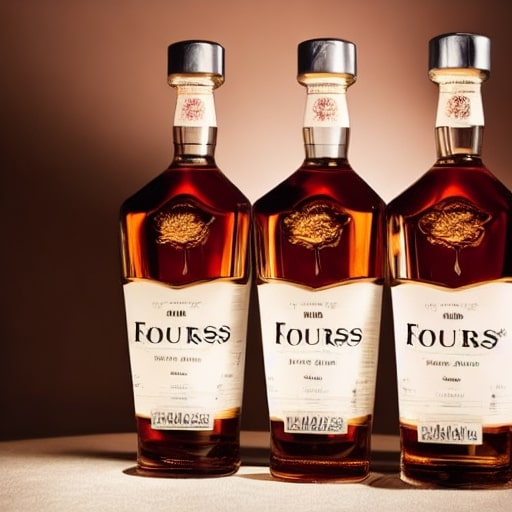} & 
            \includegraphics[width=0.3\linewidth]{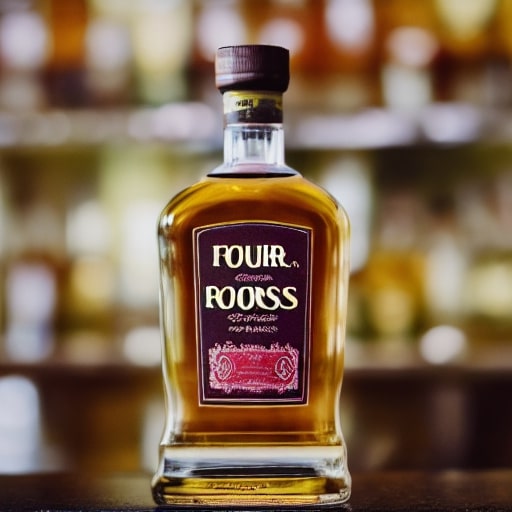} & 
            \includegraphics[width=0.3\linewidth]{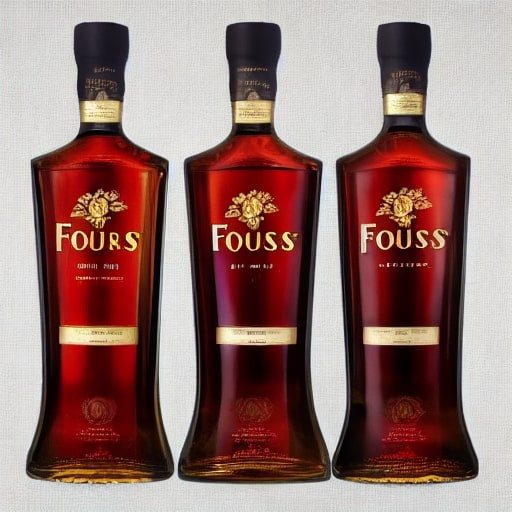} \\
            \includegraphics[width=0.3\linewidth]{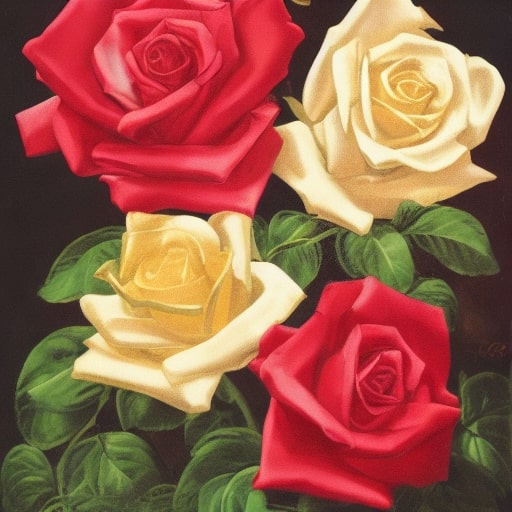} & 
            \includegraphics[width=0.3\linewidth]{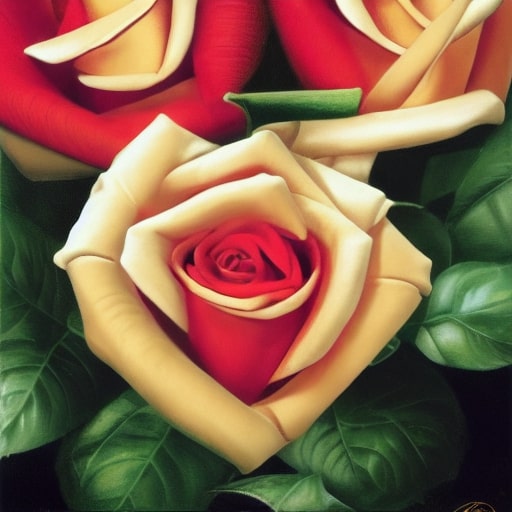} & 
            \includegraphics[width=0.3\linewidth]{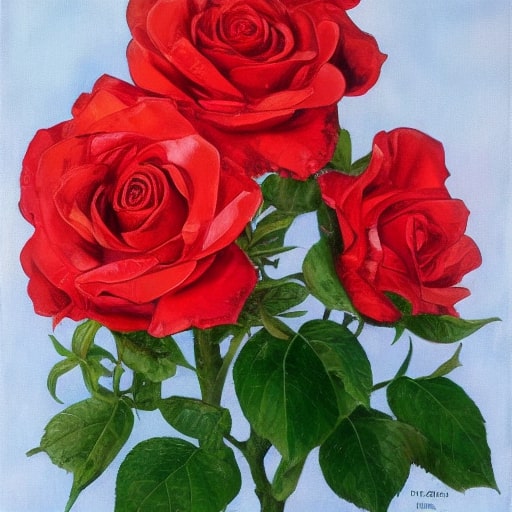} \\
            \includegraphics[width=0.3\linewidth]{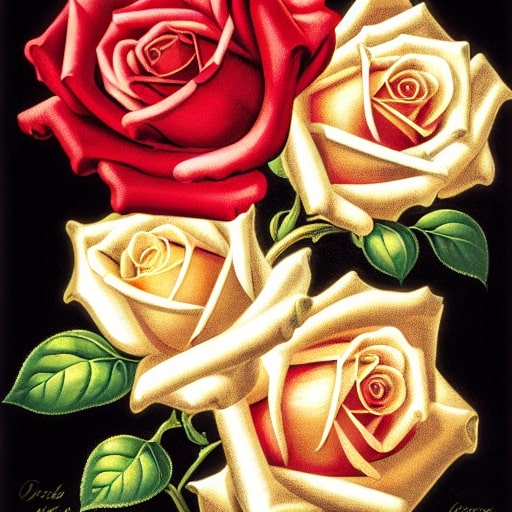} & 
            \includegraphics[width=0.3\linewidth]{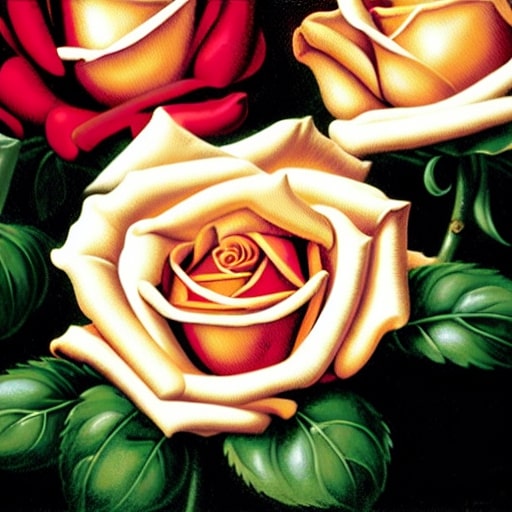} & 
            \includegraphics[width=0.3\linewidth]{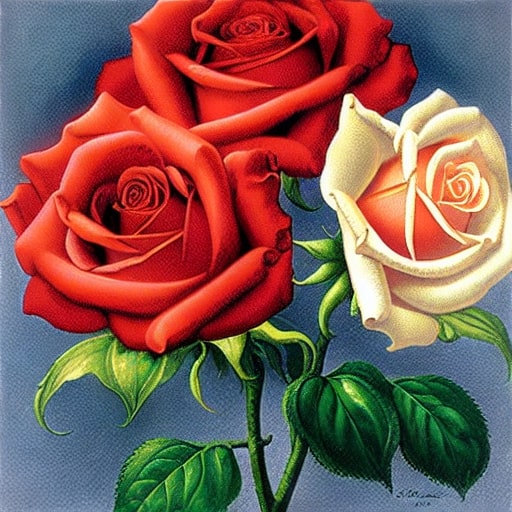} \\
        \end{tabular}
        \caption{Four roses.}
    \end{subfigure}
    \begin{subfigure}[b]{0.23\textwidth}
        \centering
        \begin{tabular}{@{}c@{}c@{}c@{}}
            \includegraphics[width=0.3\linewidth]{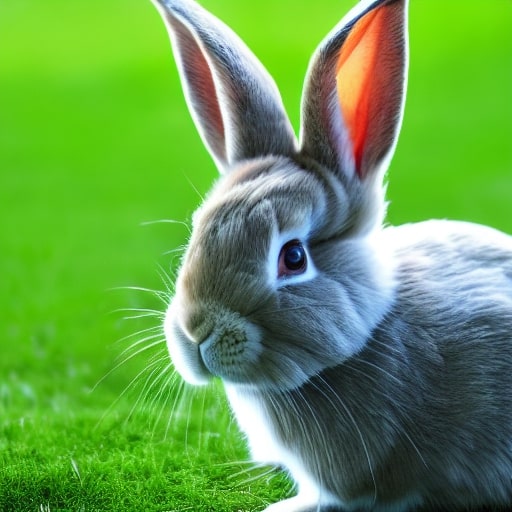} & 
            \includegraphics[width=0.3\linewidth]{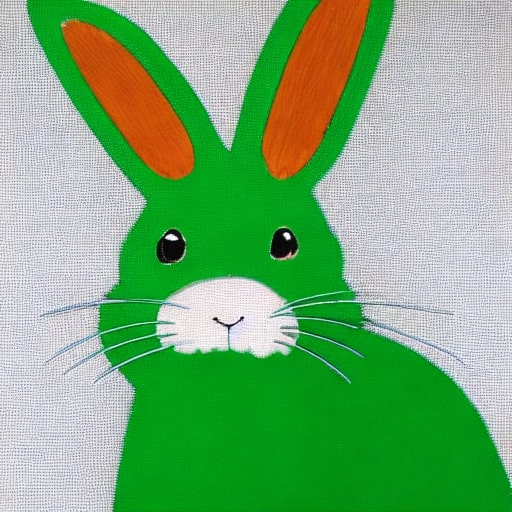} & 
            \includegraphics[width=0.3\linewidth]{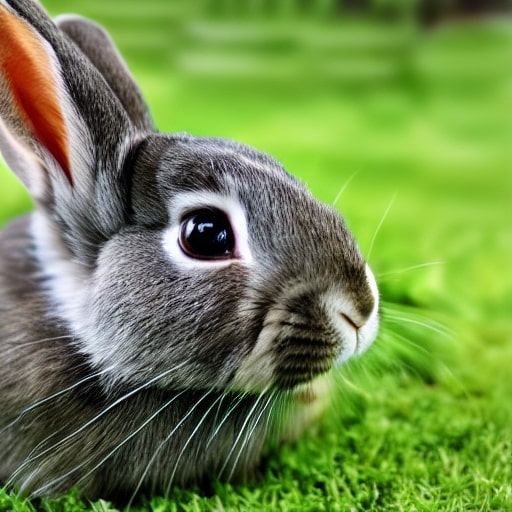} \\
            \includegraphics[width=0.3\linewidth]{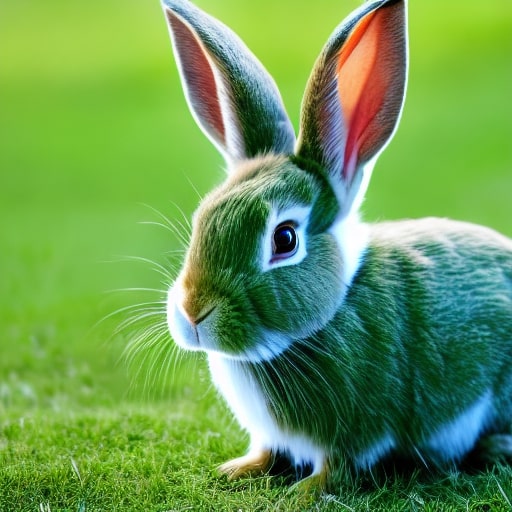} & 
            \includegraphics[width=0.3\linewidth]{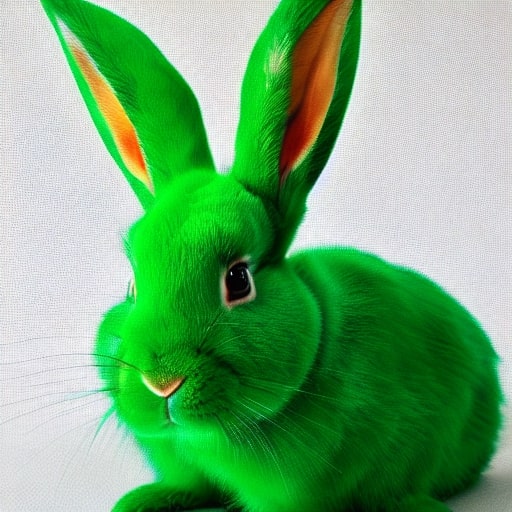} & 
            \includegraphics[width=0.3\linewidth]{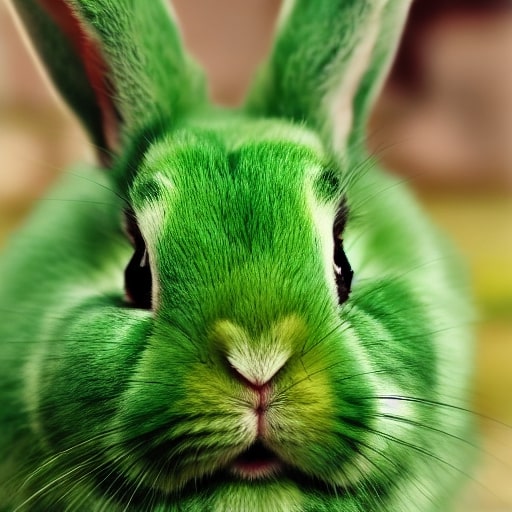} \\
            \includegraphics[width=0.3\linewidth]{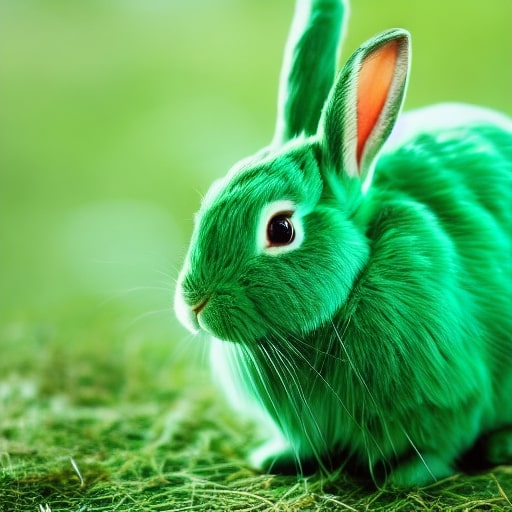} & 
            \includegraphics[width=0.3\linewidth]{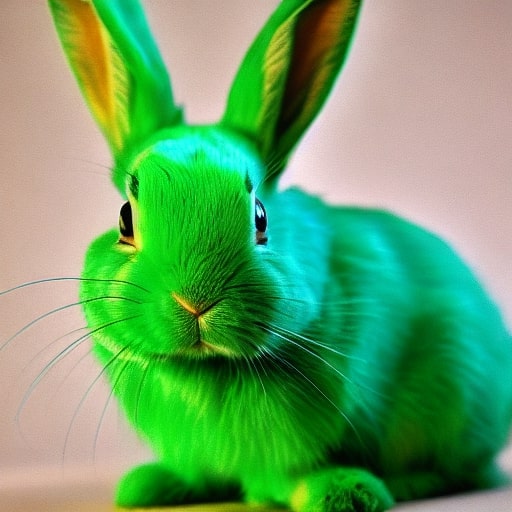} & 
            \includegraphics[width=0.3\linewidth]{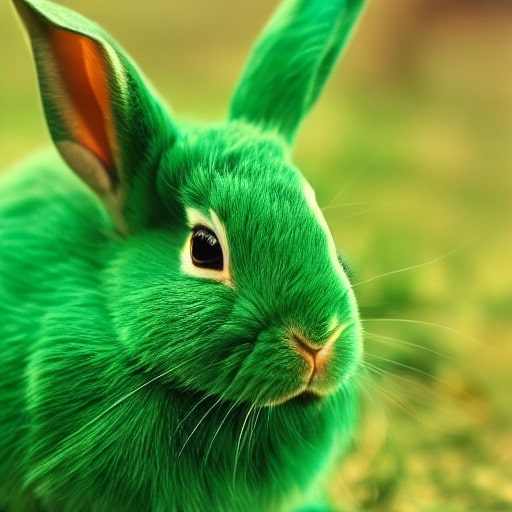} \\
        \end{tabular}
        \caption{A green rabbit.}
    \end{subfigure}
    \begin{subfigure}[b]{0.23\textwidth}
        \centering
        \begin{tabular}{@{}c@{}c@{}c@{}}
            \includegraphics[width=0.3\linewidth]{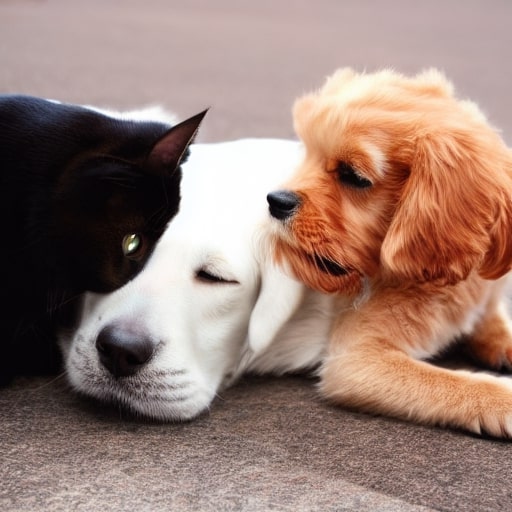} & 
            \includegraphics[width=0.3\linewidth]{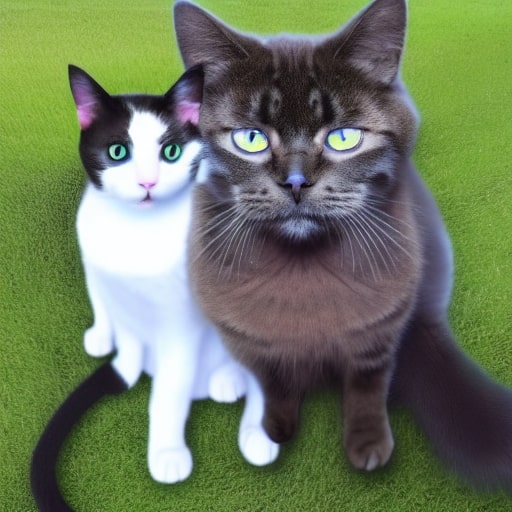} & 
            \includegraphics[width=0.3\linewidth]{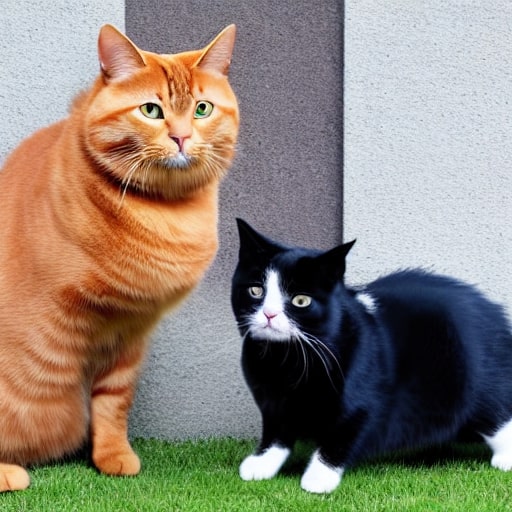} \\
            \includegraphics[width=0.3\linewidth]{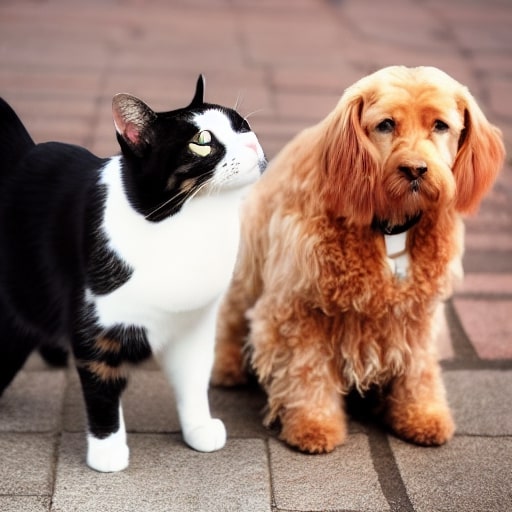} & 
            \includegraphics[width=0.3\linewidth]{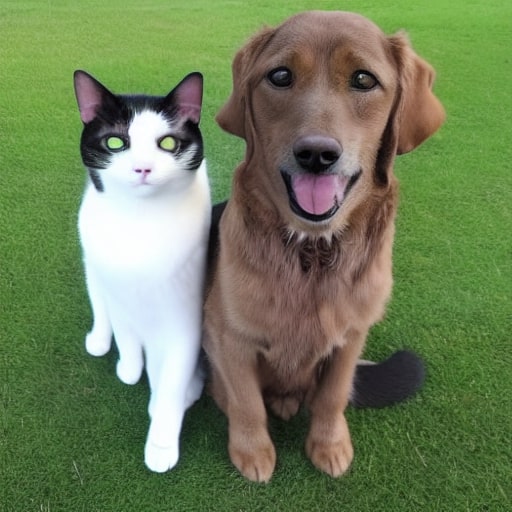} & 
            \includegraphics[width=0.3\linewidth]{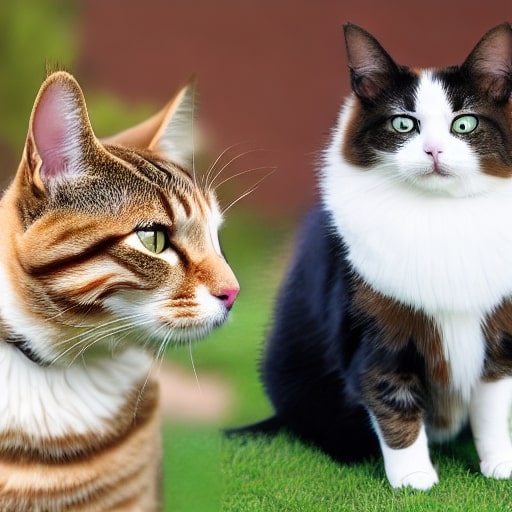} \\
            \includegraphics[width=0.3\linewidth]{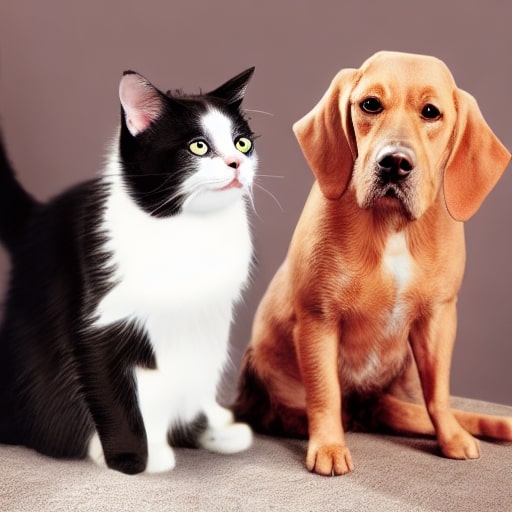} & 
            \includegraphics[width=0.3\linewidth]{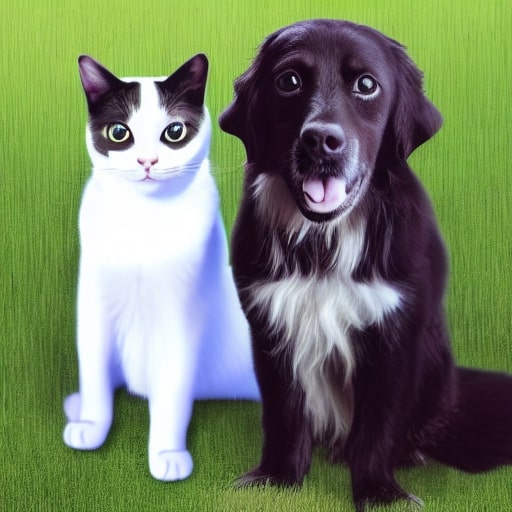} & 
            \includegraphics[width=0.3\linewidth]{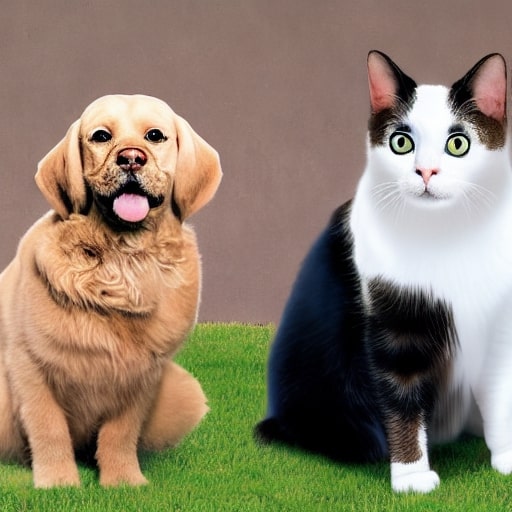} \\
        \end{tabular}
        \caption{A cat and a dog.}
    \end{subfigure}
    \begin{subfigure}[b]{0.23\textwidth}
        \centering
        \begin{tabular}{@{}c@{}c@{}c@{}}
            \includegraphics[width=0.3\linewidth]{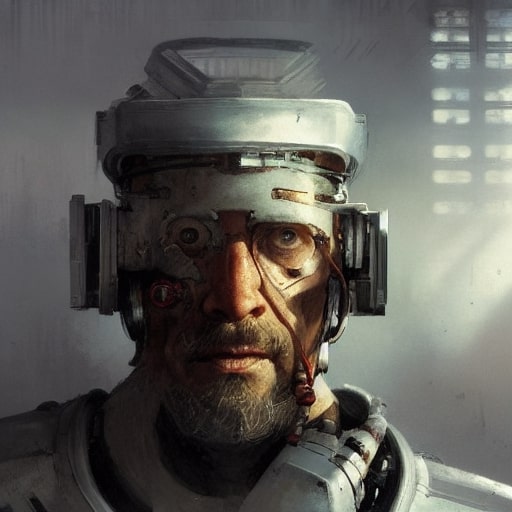} & 
            \includegraphics[width=0.3\linewidth]{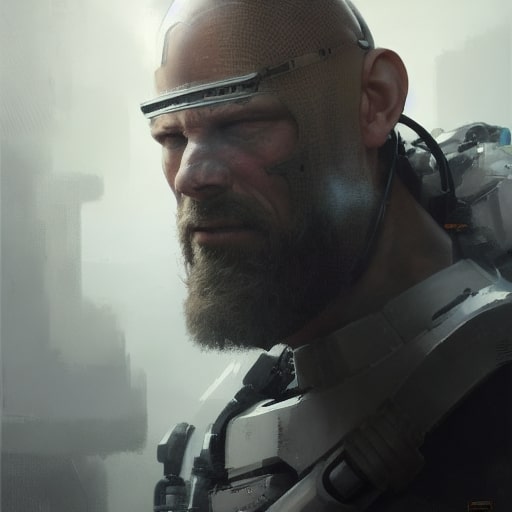} & 
            \includegraphics[width=0.3\linewidth]{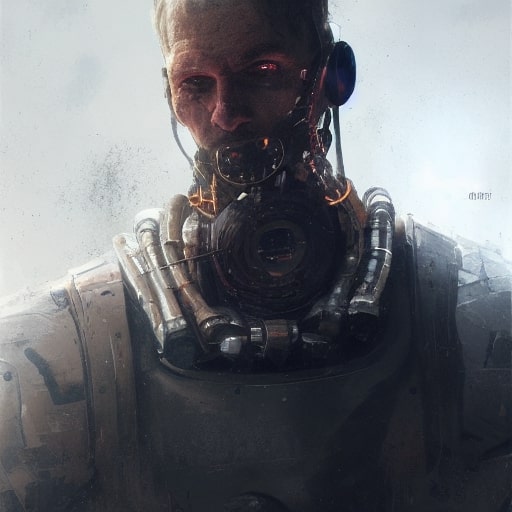} \\
            \includegraphics[width=0.3\linewidth]{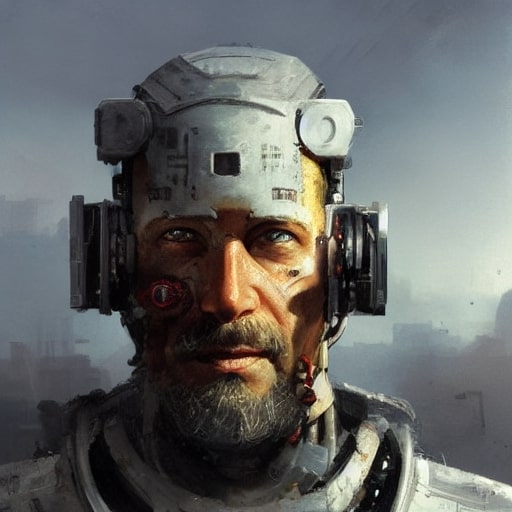} & 
            \includegraphics[width=0.3\linewidth]{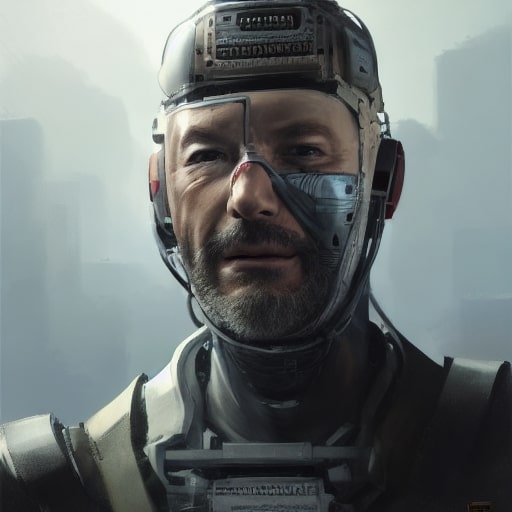} & 
            \includegraphics[width=0.3\linewidth]{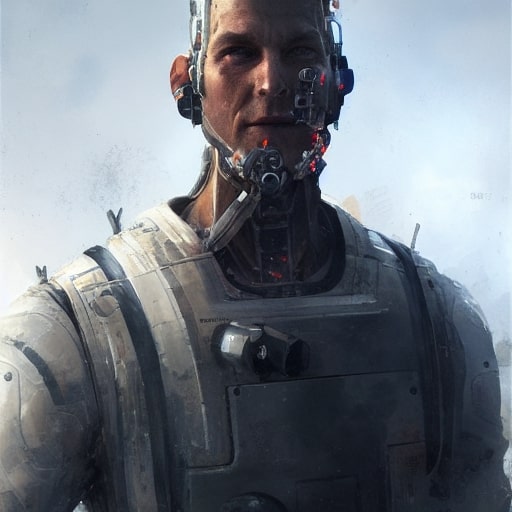} \\
            \includegraphics[width=0.3\linewidth]{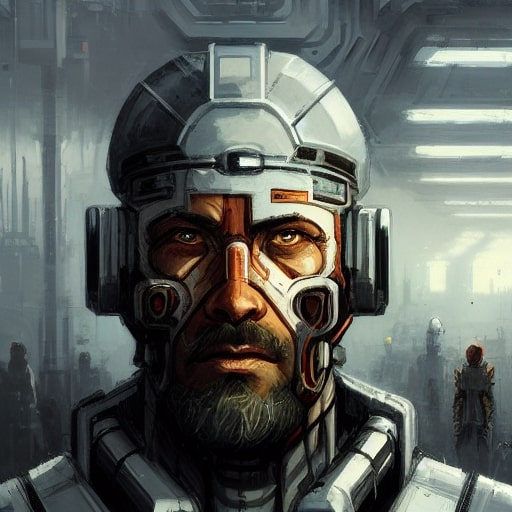} & 
            \includegraphics[width=0.3\linewidth]{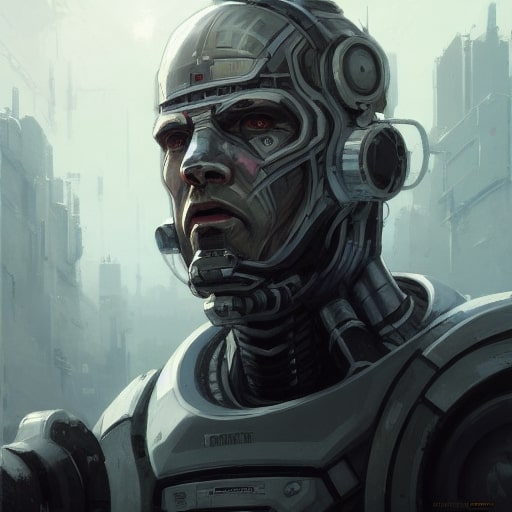} & 
            \includegraphics[width=0.3\linewidth]{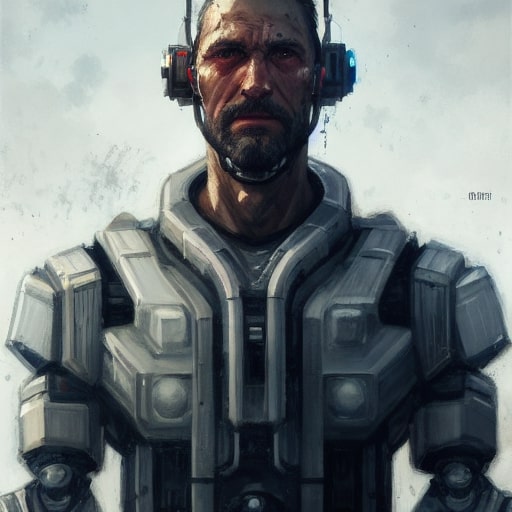} \\
        \end{tabular}
        \caption{Cybernetic man.}
    \end{subfigure}
    \caption{Images generated from prior (top row), DPOK (middle row) and RTB (bottom row) for 4 different prompts. Images in the same column share the random DDIM seed. More images in \autoref{app:promptimages}.}
    \label{fig:promptimage}
\end{figure}

\footnotetext{Full prompt for ``Cybernetic man'': ``A half - masked rugged laboratory engineer man with cybernetic enhancements as seen from a distance, scifi character portrait by greg rutkowski, esuthio, craig mullins.''}
\paragraph{Results.} \autoref{fig:text2image} plots the diversity versus log reward on a set of prompts from \cite{fan2023reinforcement,xu2023imagereward}. In terms of average $\log r(\rvx_1, \rvz)$, RTB either matches or outperforms DPOK, while generally achieving lower reward than DDPO. The CLIP diversity score for RTB and DPOK are on average higher than DDPO, which is expected since it does not use KL regularization. For qualitative image assessments, refer to \autoref{fig:promptimage} and \autoref{app:promptimages}. %
Through this experiment, we show that RTB scales well to high dimensional, multimodal data, matching state-of-the-art methods for fine-tuning text-to-image diffusion models.

\subsection{Text infilling with discrete diffusion language models}\label{sec:experiments:language_infilling}

\looseness=-1
To evaluate our approach on discrete diffusion models, we consider the problem of text infilling~\citep{zhu2019text}, which involves filling in missing tokens given some context tokens. While discrete diffusion models -- unlike their continuous counterparts -- can be challenging to train~\citep{austin2021structured,campbell2022continuous,meng2022concrete,sun2023scorebased}, \emph{score entropy discrete diffusion}~\citep[SEDD;][]{lou2023discrete} matches the language modeling performance of autoregressive language models of similar scale. Non-autoregressive generation in diffusion language models can provide useful inductive biases for infilling, such as the ability to attend to context on both sides of a target token.

\paragraph{Setup.}  \looseness=-1 We use the ROCStories corpus~\citep{mostafazadeh2016corpus}, a dataset of short stories containing 5
sentences each.  We adopt the task setup from~\cite{hu2023amortizing}, where the first 3 sentences of a story $\rvx$ and the last sentence $\rvy$ are given, and the goal is to generate the fourth
sentence $\rvz$ such that the overall story is coherent and consistent. The fourth sentence can involve a turning point in the story and is thus challenging to fill in. 
We aim to model the posterior $p\post(\rvz\mid\rvx,\rvy) \propto p(\rvz\mid\rvx) p_\text{reward}(\rvy\mid\rvx,\rvz)$ where $p$ is a SEDD language model prior (a conditional model over $\rvz$ given $\rvx$) and $p_\text{reward}$ is an autoregressive language model fine-tuned with a maximum likelihood objective on a held-out subset of the dataset. As baselines, we consider simply prompting the diffusion language model with $\rvx$ (Prompt $(\rvx)$) and $\rvx, \rvy$ (Prompt $(\rvx,\rvy)$). Additionally, to contextualize the performance, we also consider autoregressive language model baselines from~\citep{hu2023amortizing}, which studied this problem under an autoregressive prior $p(\rvz\mid\rvx)$. SFT is trained on $50,000$ examples compared to $1000$ for RTB, and serves as an upper bound on the performance in this task. See \autoref{app:infilling} for further details about the experimental setup.

\begin{wraptable}[13]{r}{0.5\textwidth}
\vspace*{-1.25em}
    \centering
    \caption{\looseness=-1 Results on the story infilling task with autoregressive and discrete diffusion language models. Metrics are computed with respect to reference infills from the dataset. All metrics are mean\std{\rm std} over 5 samples for each of the 100 test examples. RTB with discrete diffusion prior performs better than best baseline with autoregressive prior.
    }\vspace*{-0.5em}
    \resizebox{\linewidth}{!}{
    \begin{tabular}{@{}p{0.4in}lccc}
        \toprule
         Model & Algorithm $\downarrow$ Metric $\rightarrow$ & BLEU-4 & GLEU-4 & BERTScore \\
        \midrule
\multirow{3}{*}{Autoreg.}& Prompting & $0.010$\std{ 0.002} & $0.022$\std{ 0.001} & $0.005$\std{ 0.001} \\
        & Supervised fine-tuning & $0.012$\std{0.001} & $0.023$\std{ 0.001} & $0.013$\std{0.002} \\
        & GFN fine-tuning \cite{hu2023amortizing} & \highlight{$0.019$\std{0.001}} & \highlight{$0.031$\std{ 0.002}} & \highlight{$0.102$\std{ 0.005}} \\
         \midrule
         \multirow{4}{*}{\begin{minipage}{\linewidth}Discrete\\diffusion\end{minipage}} & Prompt $(\rvx)$& $0.011$\std{ 0.002} & $0.023$\std{ 0.002} & $0.014$\std{ 0.003} \\
         & Prompt $(\rvx, \rvy)$ & $0.014$\std{ 0.003} & $0.027$\std{ 0.003} & $0.092$\std{ 0.004}\\
         & \textbf{RTB (ours)} & \highlight{$0.025$\std{ 0.002}}& \highlight{$0.045$\std{0.002}} & \highlight{$0.156$\std{ 0.003}} \\
        & SFT (upper bound) & \highlightub{$0.031$\std{ 0.002}}& \highlightub{$0.057$\std{0.004}} & \highlightub{$0.182$\std{ 0.005}} \\
         \bottomrule
    \end{tabular}
    }
    \label{tab:infilling_results}
\end{wraptable}

\paragraph{Results.} \looseness=-1 Following~\citep{hu2023amortizing}, we use three standard metrics to measure the similarity of the generated infills with the reference infills from the dataset: BERTScore~\citep{Zhang*2020BERTScore:} (with DeBERTa~\citep{he2021deberta}), BLEU-4~\citep{papineni2002bleu}, and GLEU-4~\citep{wu2016google}. \autoref{tab:infilling_results} summarizes the results. We observe that the diffusion language model performs significantly better than the autoregressive language model without any fine-tuning. RTB further improves the performance over prompting, and even outperforms the strongest autoregressive baseline of GFlowNet fine-tuning. We provide some examples of generated text in \autoref{app:infilling}. %

\subsection{KL-constrained policy search in offline reinforcement learning}\label{sec:experiments:rl_offline}
The goal of RL algorithms is to learn a policy $\pi(a\mid s)$, \ie, a mapping from states $s$ to actions $a$ in an environment, that maximizes the expected cumulative discounted reward \cite{sutton2018reinforcement}. In the offline RL setting \citep{Levine2020OfflineRL}, 
the agent has access to a dataset $\mathcal{D}=\{(s_t^i,a_t^i,s_{t+1}^i,r_t^i)\}_{i=1}^N$ of transitions (where each sample $(s_t,a_t,s_{t+1},r_t)$ indicates that an agent taking action $a_t$ at state $s_t$ transitioned to the next state $s_{t+1}$ and received reward $r_t$). This dataset is assumed to be generated by a \emph{behavior policy} $\mu(a\mid s)$, which may be a diffusion model trained on $\gD$. Offline RL algorithms must learn a new policy $\pi$ which achieves high return using only this dataset without interacting with the environment. 

An important problem in offline RL is policy extraction from trained $Q$-functions \citep{peng2019advantageweighted,hansenestruch2023idql,lu2023contrastive}. For reliable extrapolation, one wants the policy to predict actions that have high $Q$-values, but also have high density under the behavior policy $\mu$, as naive maximization can result in choosing actions with low probability under $\mu$ and thus unreliable predictions from the $Q$-function. This is formulated as a KL-constrained policy search problem:
\begin{equation}\label{eq:kl_constrained_policy}
    \argmax_{\pi} \sE_{s\sim d_{\mu}, a\sim\pi(a\mid s)}[Q(s,a)],\quad \sE_{s\sim d_{\mu}}[\KL(\pi(a \mid s)\parallel \mu(a \mid s))]\leq \epsilon,
\end{equation}
where $d_\mu$ is the distribution over states induced by following the policy $\mu$. The optimal policy $\pi$ in (\ref{eq:kl_constrained_policy}) is the product distribution $\pi^*(a \mid s) \propto \mu(a \mid s)\exp(\beta Q(s,a))$ for some inverse temperature $\beta$. If $\mu(a\mid s)$ is a conditional diffusion model over continuous actions $a$ conditioned on state $s$, we use RTB to fine-tune a diffusion behavior policy to sample from $\pi^*$, using $\mu$ as the prior and $\exp(\beta Q(s,a))$ as the target constraint. We use a $Q$-function trained using IQL~\citep{kostrikov2022offline}.

\paragraph{Setup.}
We test on continuous control tasks in the D4RL suite \citep{fu2021d4rl}, which consists of offline datasets collected using a mixture of SAC policies of varying performance. We evaluate on the halfcheetah, hopper and walker2d MuJoCo~\citep{6386109} locomotion tasks, each of which contains three datasets of transitions: ``medium'' (collected from an early-stopped policy), ``medium-expert'' (collected from both an expert and an early-stopped policy) and ``medium-replay'' (transitions stored in the replay buffer prior to early stopping).
We compare against standard offline RL baselines (Behavior Cloning (BC), CQL \citep{NEURIPS2020_0d2b2061}, and IQL \citep{kostrikov2022offline}) and diffusion-based offline RL methods which are currently state-of-the-art: Diffuser \citep[D;][]{janner2022diffuser}, Decision Diffuser \citep[DD;][]{ajay2023is}, D-QL \citep{wang2023diffusion}, IDQL \citep{hansenestruch2023idql}, and QGPO \citep{lu2023contrastive}. For algorithm implementation details, hyperparameters, and a report of baselines, see \autoref{app:offline_rl}.

\paragraph{Results.}
\autoref{tab:offlinerl_results} shows that RTB matches state-of-the-art results across the D4RL tasks. In particular, RTB performs strongly in the medium-replay tasks, which contain the most suboptimal data and consequently the poorest behavior prior. We highlight that our performance is similar to QGPO~\citep{lu2023contrastive}, which learns intermediate energy densities for diffusion posterior sampling.%
\begin{table}[t]
    \vspace*{-1em}
    \centering
    \caption{Average rewards of trained policies on D4RL locomotion tasks (mean\std{\text{std}} over 5 random seeds). Following past work, numbers within 5\% of maximum in every row are highlighted.}
    \resizebox{\linewidth}{!}{
    \begin{tabular}{@{}lrrrrrrrrr}
        \toprule
          Task $\downarrow$ Algorithm $\rightarrow$ & BC & CQL & IQL & D & DD & D-QL & IDQL & QGPO & \textbf{RTB (ours)}\\
         \midrule
         halfcheetah-medium-expert & 55.2& \highlight{91.6}& 86.7& 79.8& 90.6\std{1.3}& \highlight{96.1\std{0.3}}& \highlight{95.9}& \highlight{93.5\std{0.3}}& 74.93\std{1.72}\\
         hopper-medium-expert & 52.5& 105.4& 91.5& \highlight{107.2}& \highlight{111.8\std{1.8}}& \highlight{110.7\std{1.3}}& \highlight{108.6}& \highlight{108.0\std{2.5}}& 96.71\std{3.53}\\
         walker2d-medium-expert & \highlight{107.5}& \highlight{108.8}& \highlight{109.6}& \highlight{108.4}& \highlight{108.8\std{1.7}}& \highlight{109.7\std{0.3}}& \highlight{112.7}& \highlight{110.7\std{0.6}}& \highlight{109.52\std{0.11}}\\
         \midrule
         halfcheetah-medium & 42.6& 44.0& 47.4& 44.2& 49.1\std{1.0}& 50.6\std{0.5}& 51.0& \highlight{54.1\std{0.4}}& \highlight{53.70\std{0.33}}\\
         hopper-medium & 52.9& 58.5& 66.3& 58.5& 79.3\std{3.6}& 82.4\std{4.6}& 65.4& \highlight{98.0\std{2.6}}& 82.76\std{7.07}\\
         walker2d-medium & 75.3& 72.5& 78.3& 79.7& 82.5\std{1.4}& \highlight{85.1\std{0.9}}& 82.5& \highlight{86.0\std{0.7}}& \highlight{87.29\std{3.15}}\\
         \midrule
         halfcheetah-medium-replay & 36.6& 45.5& 44.2& 42.2& 39.3\std{4.1}& \highlight{47.5\std{0.3}}& \highlight{45.8}& \highlight{47.6\std{1.4}}& \highlight{48.11\std{0.56}}\\
         hopper-medium-replay & 18.1& 95.0& 94.7& \highlight{96.8}& \highlight{100.0\std{0.7}}& \highlight{100.7\std{0.6}}& 92.1& \highlight{96.9\std{2.6}}& \highlight{100.40\std{0.21}}\\
         walker2d-medium-replay & 26.0& 77.2& 73.9& 61.2& 75.0\std{4.3}& \highlight{94.3\std{1.5}}& 85.1& 84.4\std{4.1}& \highlight{93.57\std{2.63}}\\
         \bottomrule
    \end{tabular}
    }
    \label{tab:offlinerl_results}
\end{table}

\section{Other related work}\label{sec:rw}

\paragraph{Composing iterative generative processes.} \looseness=-1 Beyond the approximate posterior sampling algorithms and application-specific techniques discussed in \autoref{sec:intro} and \autoref{sec:experiments}, several recent works have explored the use of hierarchical models, such as diffusion models, as modular components in generative processes. Diffusion models can be used to sample product distributions to induce compositional structure in images \cite{liu2022compositional,du2023reduce}. 
Amortized Bayesian inference \cite{kingma2014autoencoding,rezende2014stochastic,rezende2015variational,garnelo2018neural} is another domain of sampling from product distributions where diffusion models are now being used \cite{geffner2023compositional}.
Beyond product models, \cite{garipov2023compositional} studies ways to amortize other kinds of compositions of hierarchical processes, including diffusion models, while \cite{silva2024federated} proposes methods to sample the product of many iterative processes in application to federated learning. Finally, models without hierarchical structure, such as normalizing flows, have been used to amortize intractable inference in pretrained diffusion models \cite[\eg,][]{feng2023score}. In contrast, our method performs posterior inference by \emph{fine-tuning} a prior model, developing a direction on flexible extraction of information from large pretrained models \cite{hu2023amortizing}.

\paragraph{Diffusion samplers.} Several prior works seek to amortize MCMC sampling from unnormalized densities by training diffusion models for efficient mode-mixing~\citep{berner2022optimal, zhang2021path, vargas2023denoising, richter2023improved, vargas2024transport, akhoundsadegh2024iterated}. Our work is most closely related to continuous GFlowNets~\citep{lahlou2023theory}, which offer an alternative perspective on training diffusion samplers using off-policy flow consistency objectives \cite{lahlou2023theory,zhang2023diffusion,sendera2024diffusion}. Recently, \citet{berner2025discrete} have shown connections among existing families of diffusion sampling algorithms and analyzed their continuous-time limits.

\section{Conclusions and future work}
\label{sec:conclusion}

\looseness=-1
Relative trajectory balance provides a new approach to training diffusion models to generate unbiased posterior samples given a diffusion prior and an arbitrary reward function. Through experiments on a variety of domains -- vision, language, continuous control -- we demonstrated the flexibility and general applicability of RTB. RTB can be optimized with off-policy trajectories, and future work can explore ways to leverage off-policy training, using techniques such as local search~\citep{kim2023local,sendera2024diffusion} to improve sample efficiency and mode coverage. Simulation-based objectives in the style of \cite{zhang2021path} are also applicable to the amortized sampling problems we consider and should be explored, as should simulation-free extensions, \eg, through objectives that are local in time~\citep{madan2022learning}. The ability to handle arbitrary black-box likelihoods also makes RTB a useful candidate for inverse problems in domains such as 3D object synthesis with likelihood computed via a renderer \cite[\eg,][]{poole2022dreamfusion,wang2023score}, imaging problems in astronomy~\citep[\eg,][]{adam2022posterior}, medical imaging~\citep[\eg,][]{song2022solving}, and molecular structure prediction~\citep[\eg,][]{watson2023novo}.

Moreover, RTB could facilitate a breakthrough in modeling molecular dynamics—a notoriously challenging task due to the need to sample rare-event trajectories in chemical simulations—by converting these problems into posterior inference over amplified distributions of rare-event samples. Notably, \citet{seong2024collective} have already explored a preliminary version of this concept by employing TB with a reward multiplied by the prior likelihood, which is effectively equivalent to RTB.

\paragraph{Limitations.}
RTB learns the posterior through simulation-based training, which can be slow and memory-intensive. Additionally, the RTB objective is computed on complete trajectories without any local credit-assignment signal, which can result in high variance in the gradients. 
Guarantees on the error incurred by imperfect fit of the prior model, amortization, and time discretization (analogous to \cite{berner2025discrete}'s analysis for diffusion samplers) have not been obtained and should be considered in future work.

\paragraph{Broader impact.} While our contributions focus on an algorithmic approach for learning posterior samplers with diffusion priors, we acknowledge that like other advances in generative modelling, our approach can potentially be used by nefarious actors to train generative models to produce harmful content and misinformation. At the same time, our approach can be also be used to mitigate biases captured in pretrained models and applied to various scientific problems.   

\begin{ack}
The authors thank Adam Coogan, Yashar Hezaveh, Guillaume Lajoie, and Laurence Perreault Levasseur for helpful suggestions in the course of this project and Mandana Samiei for comments on a draft of the paper.

The authors acknowledge funding from CIFAR, NSERC, IVADO, UNIQUE, FACS Acuit\'e, NRC AI4Discovery, Samsung, and Recursion.

The research was enabled in part by computational resources provided by the Digital Research
Alliance of Canada (\url{https://alliancecan.ca}), Mila (\url{https://mila.quebec}), and
NVIDIA.
\end{ack}

\bibliographystyle{plainnat}
\bibliography{references} 
\clearpage

\appendix
\counterwithin{figure}{section}
\counterwithin{table}{section}

\section{Proofs}
\label{sec:proofs}

\printProofs

\section{Relative TB as TB under the prior measure}\label{app:relative}

The theoretical foundations for continuous generative flow networks \cite{lahlou2023theory} establish the correctness of enforcing constraints such as trajectory balance (\ref{eq:tb}) for training sequential samplers, such as diffusion models, to match unnormalized target densities. While we have considered Gaussian transitions and identified transition kernels with their densities with respect to the Lebesgue measure over $\R^d$, these foundations generalize to more general \emph{reference measures}. In application to diffusion samplers, suppose that $\pi_{\rm ref}(\rvx_t)$ is a collection of Lebesgue-absolutely continuous densities over $\R^d$ for $t=0,\Delta t,\dots,1$ and that $\overrightarrow\pi_{\rm ref}(\rvx_t\mid\rvx_{t-\Delta t}),\overleftarrow\pi_{\rm ref}(\rvx_{t-\Delta t}\mid\rvx_t)$ are collections of Lebesgue-absolutely continuous transition kernels. If these densities jointly satisfy the detailed balance condition $\pi_{\rm ref}(\rvx_t)\overleftarrow\pi_{\rm ref}(\rvx_{t-\Delta t}\mid \rvx_t)=\pi_{\rm ref}(\rvx_{t-\Delta t})\overrightarrow\pi_{\rm ref}(\rvx_t\mid \rvx_{t-\Delta t})$, then they satisfy the conditions to be reference measures. A main result of \cite{lahlou2023theory} is that if a pair of forward and backward processes satisfies the trajectory balance constraint (\ref{eq:tb}) jointly with a reward density $r$, then the forward process $p$ samples from the distribution with density $r$, with all densities interpreted as \emph{relative to the reference measures} $\pi_{\rm ref},\overleftarrow\pi_{\rm ref},\overrightarrow\pi_{\rm ref}$.\footnote{Recall that the relative density (or Radon-Nikodym derivative) of a distribution with density $p$ under the Lebesgue measure relative to one with density $\pi$ is simply the ratio of densities $p/\pi$.}

If $p_\theta$ is a diffusion model that satisfies the TB constraint jointly with some reverse process $q$ and target density $q(\rvx_1)$, then one can take the reference transition kernels $\overrightarrow\pi_{\rm ref},\overleftarrow\pi_{\rm ref}$ to be $p$ and $q$, respectively. In this case, the TB constraint for a target density $\frac1Zr(\rvx_1)$ and forward transition $p_\phi\post$ is
\begin{equation}\label{eq:relative_tb}
    \frac{p_\phi\post(\rvx_0,\rvx_{\Delta t},\dots,\rvx_1)}{\overrightarrow\pi_{\rm ref}(\rvx_0,\rvx_{\Delta t},\dots,\rvx_1)}=\frac{\frac1Zr(\rvx_1)\cancel{q(\rvx_0,\rvx_{\Delta t},\dots,\rvx_{1-\Delta t}\mid\rvx_1)}}{\cancel{\overleftarrow\pi_{\rm ref}(\rvx_0,\rvx_{\Delta t},\dots,\rvx_{1-\Delta t}\mid\rvx_1)}},
\end{equation}
which is identical to the RTB constraint (\ref{eq:rtb}). If (\ref{eq:relative_tb}) holds, then $p_\phi\post$ samples from the distribution with density $\frac1Zr(\rvx_1)$ relative to $\pi_{\rm ref}(\rvx_1)$, which is exactly $\frac1Zp_\theta(\rvx_1)r(\rvx_1)$. We have thus recovered RTB as a case of TB for non-Lebesgue reference measures.

\clearpage

\section{Posterior inference on two-dimensional Gaussian mixture model}
\label{app:2d_gmm}
\paragraph{Setup} 
We conduct toy experiments in low-dimensional spaces using samples from a Gaussian mixture model with multiple modes to visually demonstrate its validity. The prior distribution \( p(\rvx_1) \) is trained on a Gaussian mixture model with 25 evenly weighted modes, while the target posterior \( p^\mathrm{post}(\rvx_1) = r(\rvx_1) p(\rvx_1) \) uses a reward $r(\rvx_1)$ to select and re-weight 9 modes from \( p(\rvx_1) \). More specifically, the resulting posterior is:  
\begin{align}
    p^\mathrm{post}(\rvx_1) &= \frac{1}{\sum_j \tilde{\pi}_j}\sum_{i} \tilde{\pi}_i \mathcal{N}(\rvx_1 \mid \mathbf{\mu}_i, \mathbf{I}) \\
    \{\mathbf{\mu}_i\} &= \{ (-10, -5), (-5, -10), (-5, 0), (10, -5), (0, 0), (0, 5),
            (5, -5), (5, 0), (5, 10)\} \\
    \{\tilde{\pi}_i\} &= \{4, 10, 4, 5, 10, 5, 4, 15, 4\}
\end{align}

Our objective is to sample from the posterior \( p^\mathrm{post}(\rvx_1) \). We compare our method with several baselines, including policy gradient reinforcement learning (RL) with KL constraint and classifier-guided diffusion models. For RL, we implemented the REINFORCE method with a mean baseline and a KL constraint, following recent work training diffusion models to optimize a reward function~\citep{black2024training}. Sampling according to the RL policy leads to a distribution $q_\theta(\rvx_1)$, which is trained with the objective:

\begin{equation}
     J(\theta) = \mathbb{E}_{q_{\theta}(\rvx_1)} \left[ r(\rvx_1) \right] + \alpha \KL \left( q_{\theta}(\rvx_1) \| p(\rvx_1) \right)
\end{equation}

While the exact computation of \( KL \left( q_{\theta}(\rvx_1) \| p(\rvx_1) \right) \) is intractable, we follow the approximation method introduced by~\citet{fan2023reinforcement}, which sums the divergence at every diffusion step. This approximation optimizes an upper bound of the marginal KL.

The other baseline is classifier (energy) guidance, which given a diffusion prior, samples using a posterior score function estimate:
\begin{equation}
    \nabla_{\rvx_t} \log p^\mathrm{post}(\rvx_t) \approx \nabla_{\rvx_t} \log p(\rvx_t) + \nabla_{\rvx_t} \log r(\rvx_t)  
\end{equation}

Note that this is a biased approximation of the true intractable score:
\begin{equation}
\nabla_{\rvx_t} \log p^\mathrm{post}(\rvx_t) = \nabla_{\rvx_t} \log p(\rvx_t) + \nabla_{\rvx_t} \log \mathbb{E}_{p(\rvx_1 \mid \rvx_t)} [r(\rvx_1)]
\end{equation}

For our experiments, we follow the source code\footnote{\url{https://github.com/GFNOrg/gfn-diffusion}} provided in recent diffusion sampler benchmarks~\citep{sendera2024diffusion}. We utilize a batch size of 500, with finetuning at 5,000 training iterations, a learning rate of 0.0001, a diffusion time scale of 5.0, 100 steps, and a log variance range of 4.0. The neural architecture employed is identical to that used in~\citep{sendera2024diffusion}. For pretraining the prior model, we use the same hyperparameters as above, but with 10,000 training iterations using maximum likelihood estimation with true samples.

\begin{figure}[t!]
  \begin{minipage}{0.13\textwidth}
        \centering
        \includegraphics[width=\textwidth]{figures/toy_exp/2d_gmm/25gmm_prior.png} %
        \subcaption{Prior}
    \end{minipage}
  \begin{minipage}{0.13\textwidth}
        \centering
        \includegraphics[width=\textwidth]{figures/toy_exp/2d_gmm/post_kde.png} %
        \subcaption{Posterior}
    \end{minipage}
  \begin{minipage}{0.13\textwidth}
        \centering
        \includegraphics[width=\textwidth]{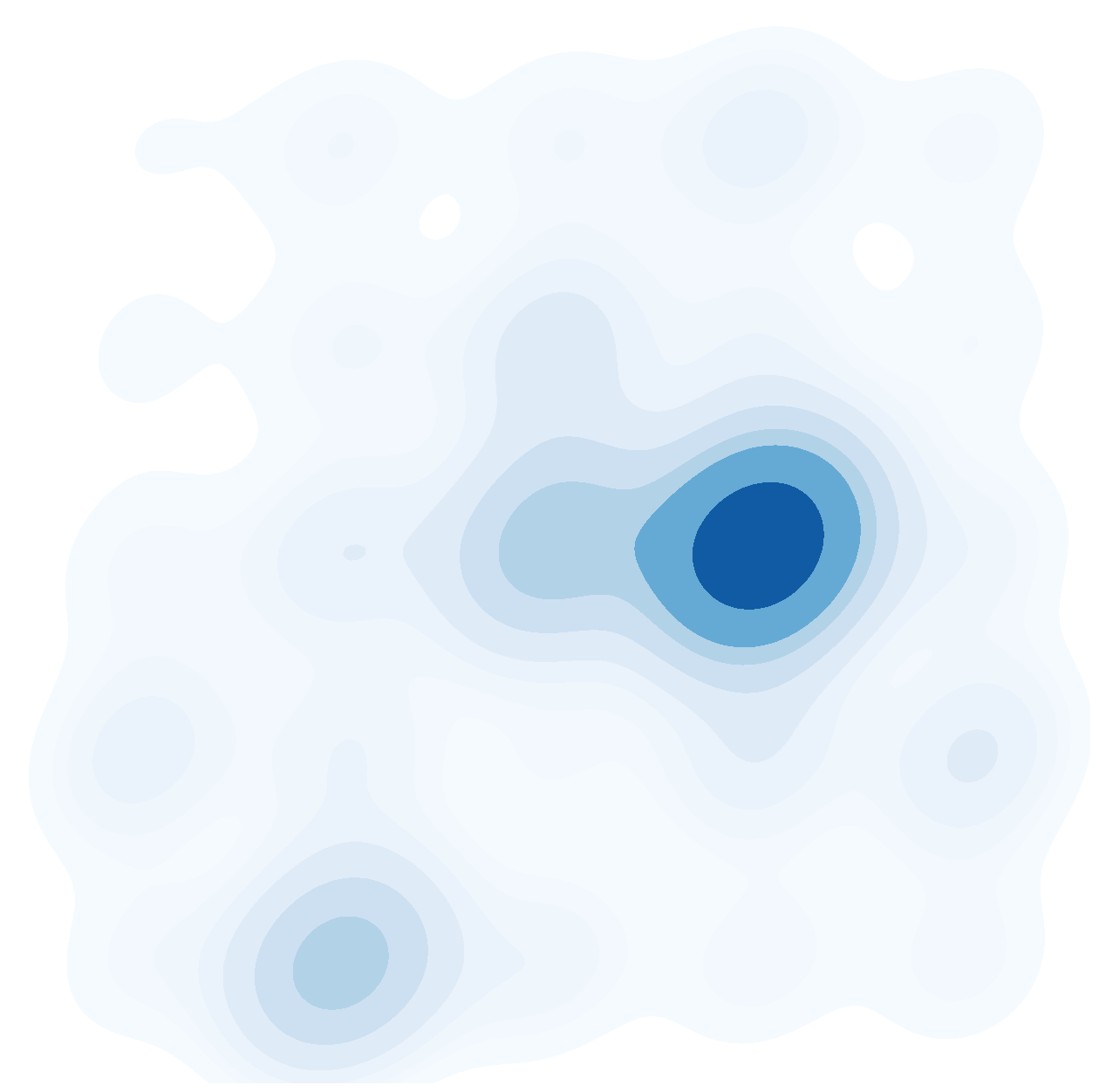} %
        \subcaption{$\alpha = 1.0$}
    \end{minipage}\hfill
  \begin{minipage}{0.13\textwidth}
        \centering
        \includegraphics[width=\textwidth]{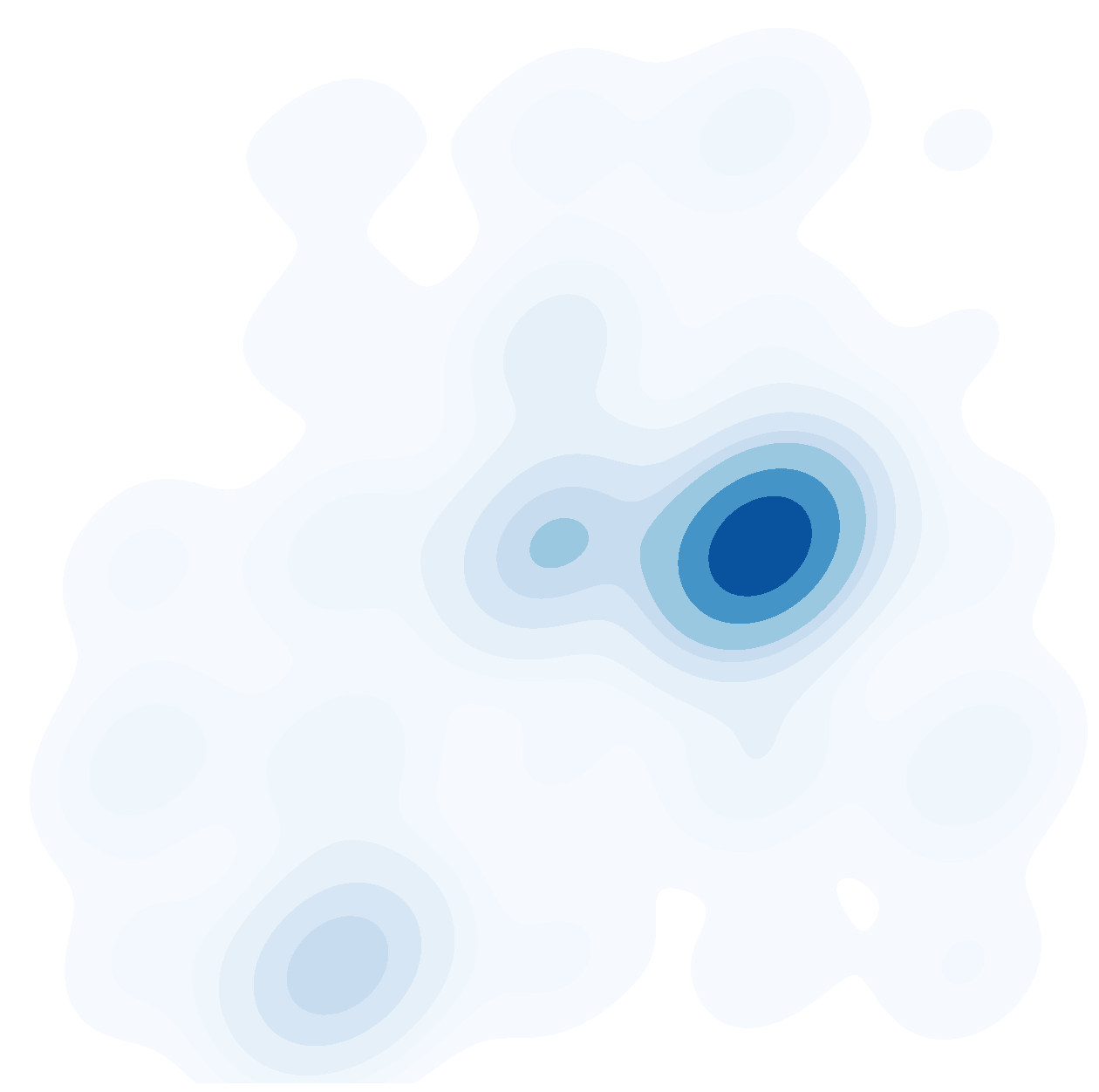} %
        \subcaption{$\alpha = 0.8$}
    \end{minipage}\hfill
  \begin{minipage}{0.13\textwidth}
        \centering
        \includegraphics[width=\textwidth]{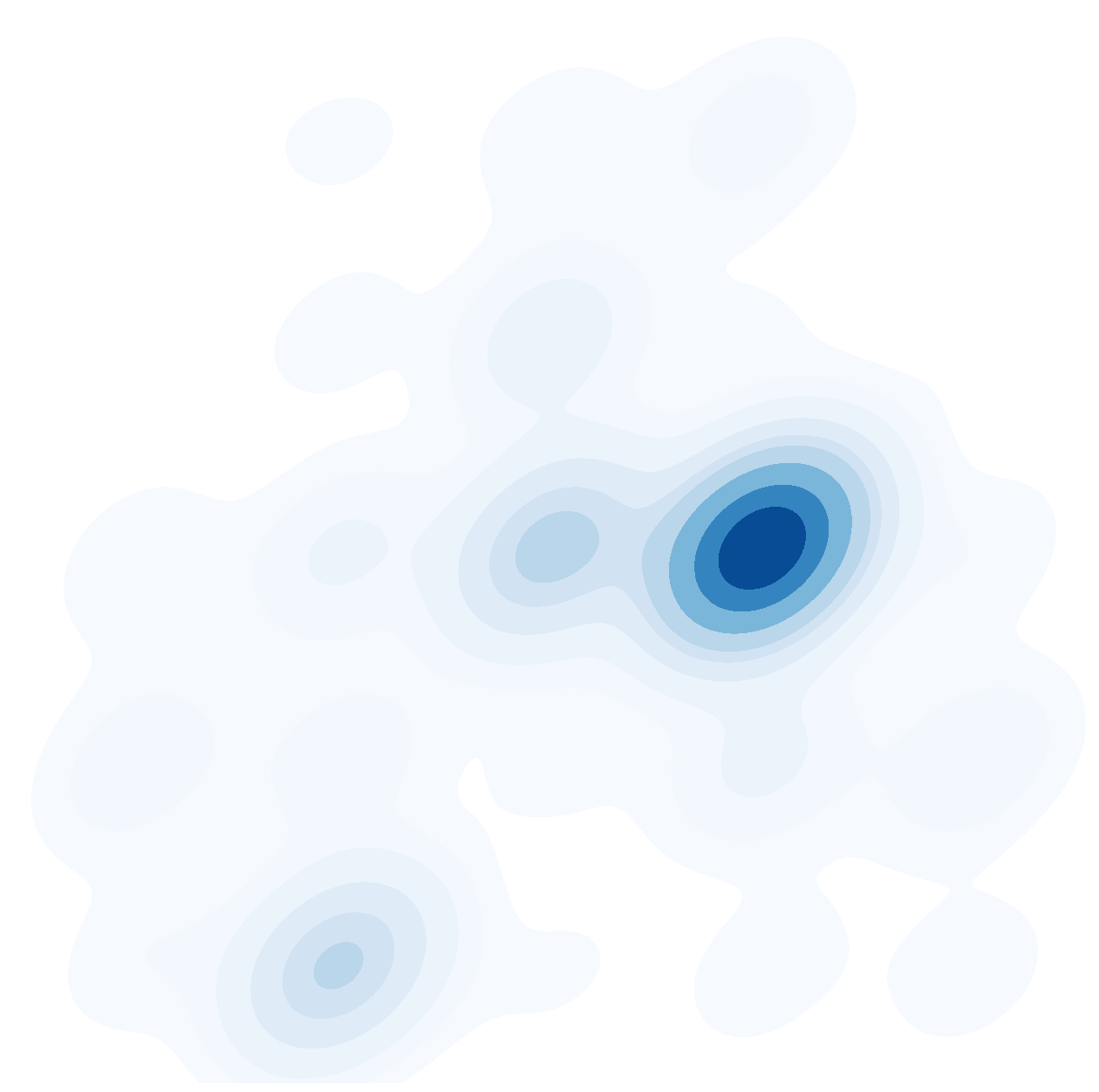} %
        \subcaption{$\alpha = 0.7$}
    \end{minipage}\hfill
  \begin{minipage}{0.13\textwidth}
        \centering
        \includegraphics[width=\textwidth]{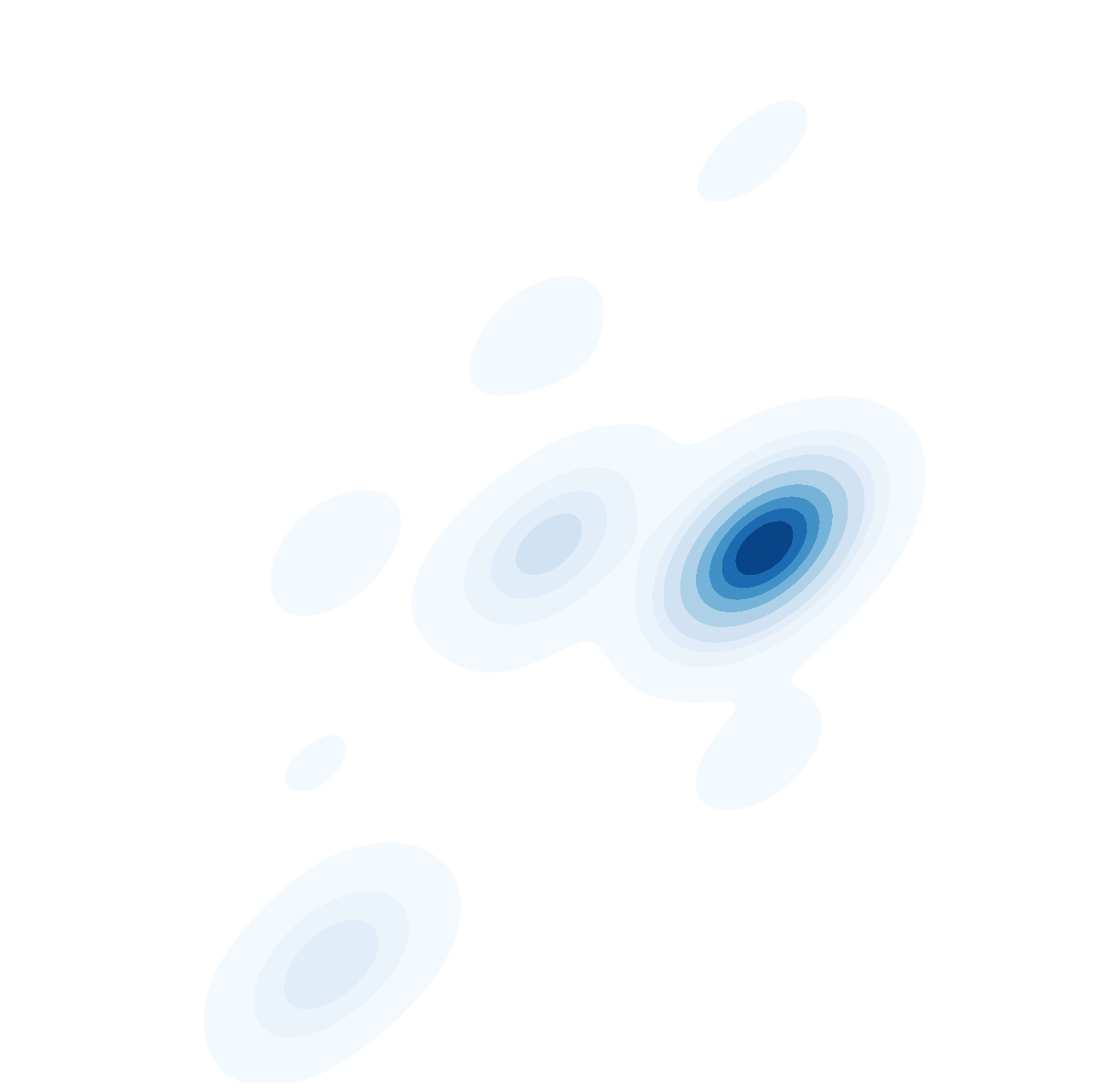} %
        \subcaption{$\alpha = 0.5$}
    \end{minipage}\hfill
  \begin{minipage}{0.13\textwidth}
        \centering
        \includegraphics[width=\textwidth]{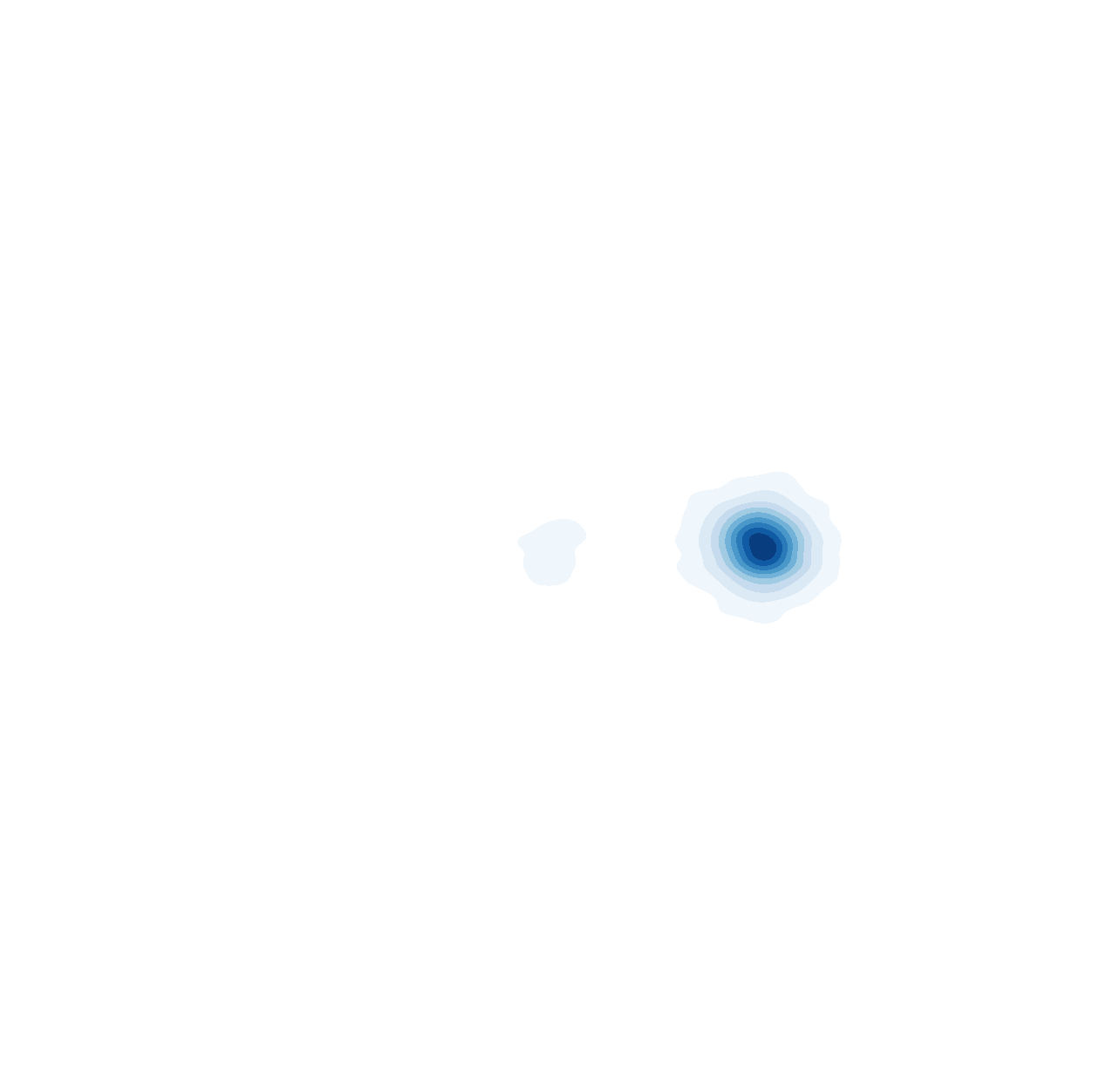} %
        \subcaption{$\alpha = 0.3$}
    \end{minipage}
    
\caption{Tuning the KL weight $\alpha$ in reinforcement learning: influences the balance between sticking to the prior distribution and moving towards the modes of the reward density. A higher $\alpha$ value maintains closer adherence to the prior, while a lower $\alpha$ allows a gradual shift towards high values of $r(\rvx)$. Setting $\alpha$ below 0.3 tends to cause mode collapse, moving too far from the prior and focusing on maximizing rewards for single modes. $\alpha = 0.5$ gives us samples that closest resembles the posterior.}
\label{fig:rl_tuning}
\end{figure}

\begin{figure}[t!]
  \begin{minipage}{0.13\textwidth}
        \centering
        \includegraphics[width=\textwidth]{figures/toy_exp/2d_gmm/25gmm_prior.png} %
        \subcaption{Prior}
    \end{minipage}\hfill
  \begin{minipage}{0.13\textwidth}
        \centering
        \includegraphics[width=\textwidth]{figures/toy_exp/2d_gmm/post_kde.png} %
        \subcaption{Posterior}
    \end{minipage}\hfill
  \begin{minipage}{0.13\textwidth}
        \centering
        \includegraphics[width=\textwidth]{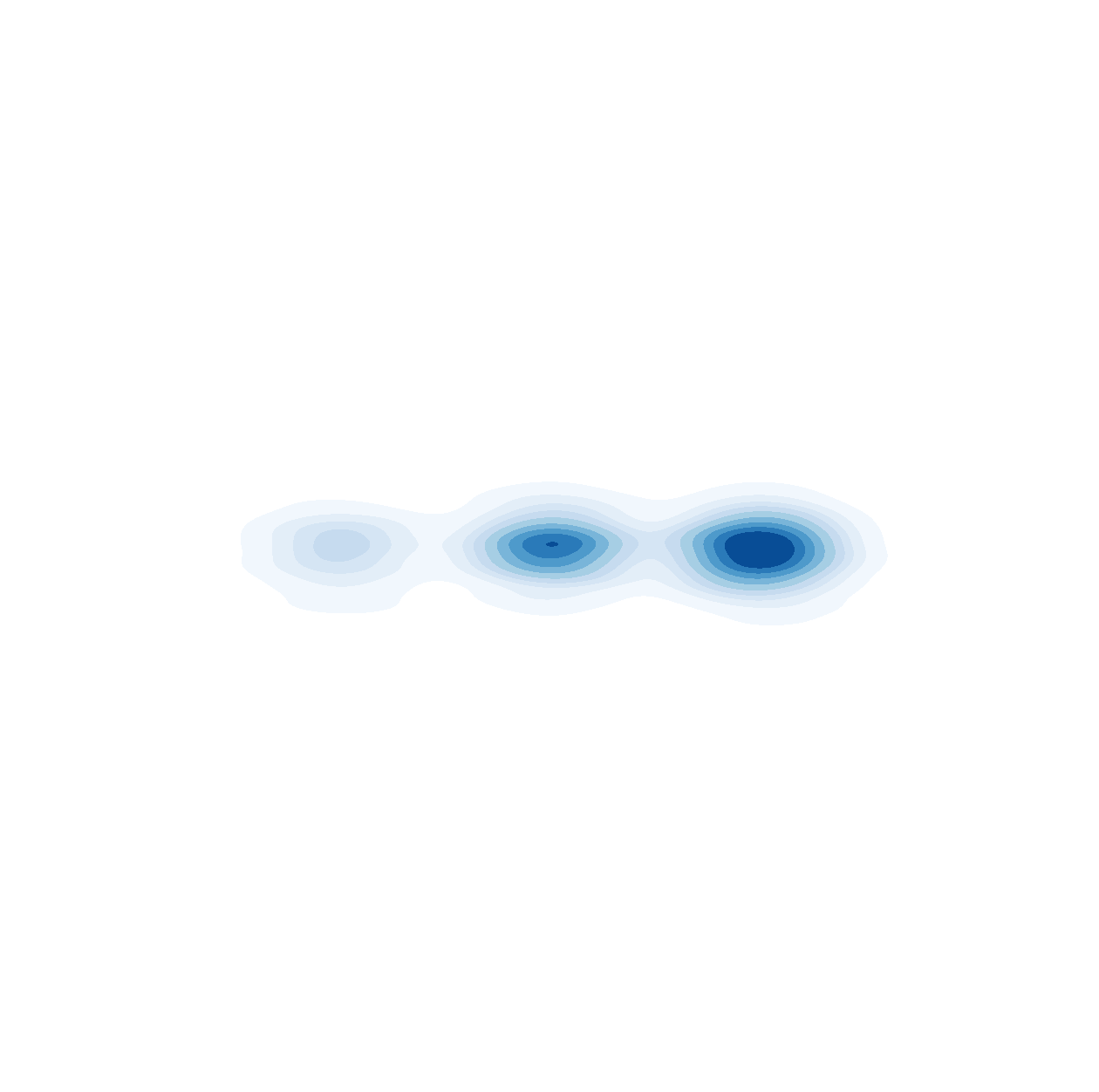} %
        \subcaption{RTB (on)}
    \end{minipage}\hfill
  \begin{minipage}{0.13\textwidth}
        \centering
        \includegraphics[width=\textwidth]{figures/toy_exp/2d_gmm/rtb_exp_0.5kde.png} %
        \subcaption{RTB (off)}
    \end{minipage}\hfill
  \begin{minipage}{0.13\textwidth}
        \centering
        \includegraphics[width=\textwidth]{figures/toy_exp/2d_gmm/ablations/rl_grid_search/rl_finetune_kl_0.5_exp_0.0kde.png} %
        \subcaption{RL (on)}
    \end{minipage}\hfill
  \begin{minipage}{0.13\textwidth}
        \centering
        \includegraphics[width=\textwidth]{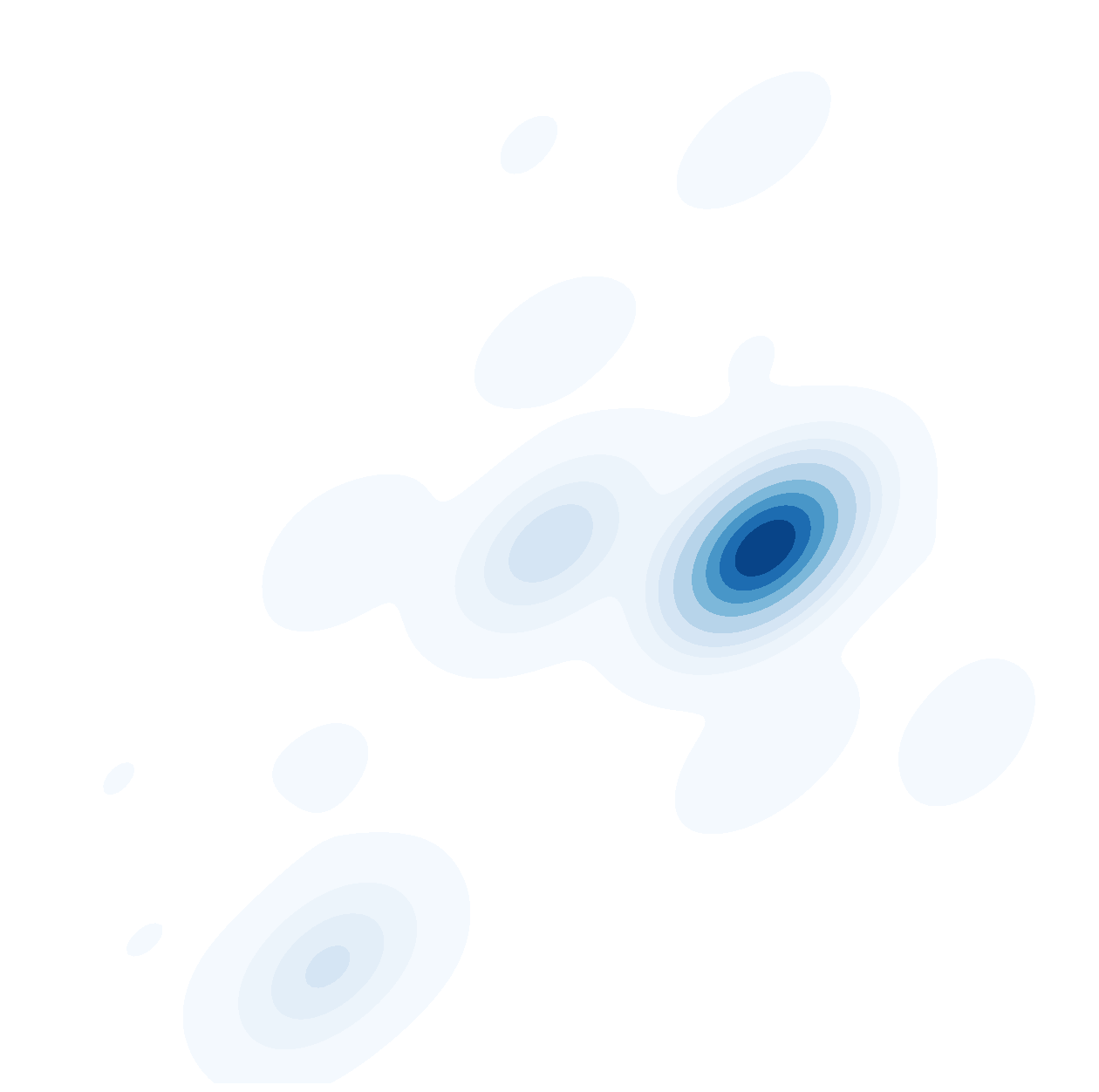} %
        \subcaption{RL (off)}
    \end{minipage} 
\caption{Off-policy exploration benefits for RTB training. RTB, with simple off-policy exploration techniques that increase randomness in the diffusion process, significantly improves mode coverage. On the other hand, policy gradient RL methods which are typically used to finetune diffusion models are on-policy, and hence prone to mode collapse. }
\label{fig:off-policy}
\end{figure}

\paragraph{Results.} As we reported in the main text, in \autoref{fig:2d_gmm}, we present illustrative results. The classifier-guided diffusion model shows biased posterior sampling (\autoref{fig:2d_classifier}), failing to provide accurate inference. RL with a per step KL constraint cannot exactly optimize for the posterior distribution, making the tuning of the KL weight \(\alpha\) crucial to achieving desirable output \autoref{fig:rl_tuning}. RTB asymptotically achieves the true posterior without introducing a balancing hyperparameter \(\alpha\). %
Another advantage of our approach is off-policy exploration for efficient mode coverage. RL methods for fine-tuning diffusion models (e.g., DPOK~\cite{fan2023reinforcement}, DDPO~\cite{black2024training}) typically use policy gradient style methods that are on-policy. By using a simple off-policy trick introduced by~\cite{malkin2023gflownets,lahlou2023theory} and demonstrated by~\citet{sendera2024diffusion}, we can introduce randomness into the exploration process in diffusion by adding \(\frac{\epsilon^2}{T}\), where \(\epsilon\) is a noise hyperparameter and \(T\) is the diffusion timestep, into the variances and annealing it to zero over training iterations. We set \(\epsilon = 0.5\) for off-policy exploration. As shown in \autoref{fig:off-policy}, RTB with off-policy exploration gives very close posterior inferences, whereas off-policy exploration in RL with \(\alpha = 0.5\) (which is a carefully selected hyperparameter) does not improve performance due to its on-policy nature.

\section{Code}\label{app:code}
Code for all experiments is available at \thegithuburl{} and will
continue to be maintained and extended.

\section{On classifier guidance and RTB posterior sampling}\label{app:cls_rtb}

\begin{figure}[h!]
\centering
\includegraphics[width=.98\linewidth]{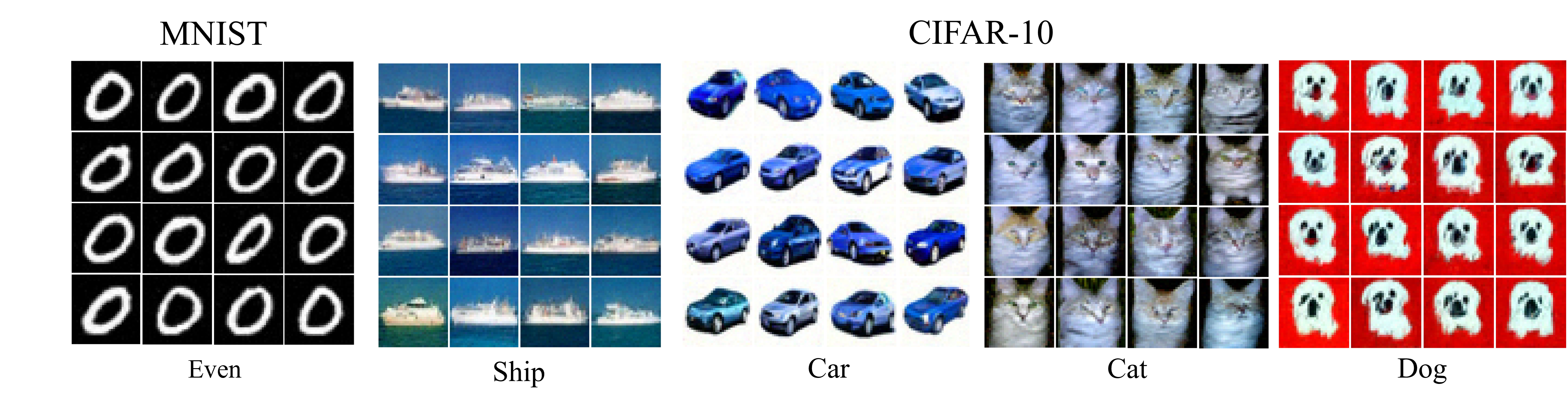}
\caption{Samples from a posterior model fine-tuned with RL (no KL). We observe early mode collapse, showcasing high-reward samples with minimal diversity.} 
\label{app:fig:cls_guidance_collapse}
\end{figure}

\subsection{Experimental Details}\label{app:cls_rtb:experimental_details}
In our experiments, we fine-tune pretrained unconditional diffusion models with our RTB objective, to sample from a posterior distribution in the form $p\post(x \mid y) = p(y \mid x) p(x)$. In this section, we detail the experimental settings for RTB as well as the compared baselines.

\paragraph{Experiments setting.} For MNIST, we pretrain a noise-predicting diffusion model on $28\times28$ (upscaled to $32\times32$) single channel images of digits from the MNIST datasets. We discretize the forward and backward processes into 200 steps and train our model until convergence. For CIFAR-10, we use a pretrained model from \cite{ho2020ddpm}, trained to generate $32\times32$ 3-channel images from the CIFAR-10 dataset, while discretizing the noising/denoising processes into 1000 steps. For fine-tuning the prior, we parametrize the posterior with LoRA weights~\cite{hu2022lora}, with the number of parameters equal to about 3\% of the prior model's parameter count. We train our models on a single NVIDA V100 GPU.

We compute FID as a similarity score estimate of the \emph{true} posterior distribution from the data. As such, the computation is limited to the total number of per-class-samples present in the data, (between 5k and 6k for CIFAR-10 and MNIST digits, and 30k for the even/odd task).

\paragraph{RTB.} For RTB fine-tuning, we finetune a diffusion model following the objective in \autoref{eq:rtb_objective}. We impose the objective while sampling denoising paths following a DDPM sampling scheme, with only 20\% to 50\% of the original trained steps. We employ loss clipping at $0.1$, to account for imperfect constraints in the pretrained prior, and train each of our models for 1500 training iterations, well into convergence trends.

\paragraph{RL ~\cite{fan2023reinforcement}.} We implement two RL-based fine-tuning techniques derived from DPOK~\cite{fan2023reinforcement} and DDPO~\citep{black2024training}, respectively with and without KL regularization. These implementations use a reinforcement learning baseline similar to the one in our experiments described in \autoref{sec:experiments:text2image}. By following the same sampling scheme as in our RTB experiments, we enable a direct comparison with RTB. To fine-tune the KL weight, we perform a search over \(\alpha \in \{0.01, 0.1, 1.0\}\).

\paragraph{DP \cite{chung2023diffusion}.}  We implement and adapt the Gaussian version of the posterior sampling scheme in \cite{chung2023diffusion}, originally devised for noisy inverse problems. This method relaxes some of our experimental constraints, as it requires a differentiable reward $r(\rvx)$.  We perform a sweep over ten values of the suggest parameter range for the step size $\zeta\in[.1, 1.]$ on MNIST single-digit sampling, and choose $\zeta=0.1$ for our experiments.

\paragraph{LGD-MC~\cite{song2023loss}.} We adapt the implementation of the algorithm in \cite{song2023loss} to sample from the classifier-based posteriors in CIFAR-10 and MNIST. Similarly to the DP baseline, we use our pretrained classifier to perform measurements at each sampling step, and use a Monte Carlo estimate of the gradient correction to guide the denoising process. We choose $\zeta=0.1$ following the DP experiments and default the number of particles to 10 as per the authors' guidelines.

\subsection{Additional findings.} \label{app:cls_rtb:additional_findings}

\paragraph{Classifier-guidance baselines.} We find that the DP and LGD-MC classifier-guidance based baselines struggle to sample from the true posterior distribution in our experimental settings. The baselines achieve the lowers classifier average rewards in all tested settings. Despite choosing $\zeta=0.1$ as the validate best performing hyperparameter, we also also observe the posterior samples from DP and LGD-MC to be close to the prior. As such, DP and LGD-MC score high in diversity, and low in FID for the Even/Odd experimental scenario, as expected from prior sampling benchmarks, but failing to appropriately model the posterior distribution.

\paragraph{RL and mode collapse.} In the pure Reinforcement Learning objective imposed for the experiments in \autoref{sec:experiments:cls_guidance} (no KL), we observe a significantly higher reward than other baseline methods, while showcasing increased FID and lower diversity. In \autoref{app:fig:cls_guidance_collapse} we show a random set of 16 samples for posterior models trained on 4 different classes of the CIFAR-10 datasets, as well as the \emph{Even} objective from the MNIST dataset, after 500 training iterations. In the figure, we observe early mode collapse and reward exploitation, visually evident from the little to no variation amongst samples for each class class, and single-digit collapse in the multi-modal \emph{even} digits objective (see samples in \autoref{fig:cls_finetuning_samples} for comparison with our RTB-finetuned models).

\paragraph{Conditional architectures.}

\begin{wraptable}[11]{r}{0.5\linewidth}
\vspace*{-1.5em}
\caption{\looseness=-1 Conditional experiment for MNIST even/odd posterior. Note that the posterior model in the conditional experiment is different from that in the baselines because it uses a different architecture that includes an additional input channel.\label{tab:conditional_mnist}}
\vspace*{-0.5em}
\centering
\resizebox{1\linewidth}{!}{
\begin{tabular}{@{}lcc}
\toprule
Dataset $\rightarrow$ & \multicolumn{2}{c}{MNIST even/odd} \\
\cmidrule(lr){2-3}
Algorithm $\downarrow$ Metric $\rightarrow$ & $\mathbb{E}[\log r(\mathbf{x})]$ ($\uparrow$) & FID ($\downarrow$) \\
\midrule
DPS    & $-1.2270$\std{$0.202$} & \highlight{$1.1498$\std{$0.182$}} \\
LGD$-$MC   & $-1.1720$\std{$0.199$} & \highlight{$1.1445$\std{$0.184$}} \\
DDPO   & \highlight{$-8.6$\std{$12.3$}$\times10^{-11}$} & $1.8024$\std{$0.423$}  \\
DPOK    &$-0.0783$\std{$0.082$} & $1.2536$\std{$0.206$}  \\
\textbf{RTB (unconditional)} & $-0.1816$\std{$0.175$} & \highlight{$1.1794$\std{$0.171$}} \\
\midrule
\textbf{RTB (conditional)} & -0.1236 & \highlight{0.9112} \\
\bottomrule
\end{tabular}
}
\end{wraptable}

We repeat the even/odd posterior sampling experiment of \autoref{sec:experiments:cls_guidance} in a conditional setting, where the condition is an input to the posterior model. For the posterior architecture, we use a naive modification of the prior with an extra input channel, which is populated a full mask of 0 or 1 for conditioning on the even and odd classes, respectively. The results are shown in \autoref{tab:conditional_mnist}. We look forward to future work which develops more specialized architectures for handling conditional constraints.

\section{Infilling with discrete diffusion}
\label{app:infilling}
\paragraph{Additional details.}

We illustrate some examples from the ROC Stories dataset used for training in~
\autoref{tab:examples_infilling}. For the prior we use the \texttt{sedd-small}\footnote{\url{https://huggingface.co/louaaron/sedd-small}} model, which uses an absorbing noising process~\citep{austin2021structured} with a log linear noise schedule, as the diffusion prior $p(\rvz\mid\rvx)$. The posterior model is parameterized as a copy of the prior. To condition the diffusion model on the beginning $\rvx$ we set the tokens at the appropriate location in the state in the initial time step, \ie $t=0$. Our implementation is based on the original SEDD codebase~\footnote{\url{https://github.com/louaaron/Score-Entropy-Discrete-Diffusion}}. Training this model is computationally expensive (in terms of memory and speed) so we utilize the stochastic TB trick, only propagating the gradients through a subset of the steps of the trajectory. We also use the loss clipping trick as discussed in \autoref{sec:training}. Specifically, we clip the loss below a certain threshold to 0, resulting in updates only when the loss is larger. This threshold -- referred to as the loss clipping coefficient -- is a hyperparameter. As this is a conditional problem we also use the relative VarGrad objective. We also use some tempering on the reward likelihood which helps in learning (\ie, $p_\text{reward}(\rvy\mid\rvx,\rvz)^{\beta}$) where $\beta$ is the inverse temperature parameter. %
We perform all experiments on an NVIDIA A100-Large GPU. 
Note that we also tried a baseline of simply fine-tuning the diffusion model on the data but encountered some training instabilities that we could not fix. 
The hyperparameters used for training RTB in our experiments are detailed in \autoref{tab:hps_infill}. 

\paragraph{Reward.}
For training $p_\text{reward}$ we follow the training procedure and implementation from~\citep{hu2023amortizing}\footnote{\url{https://github.com/GFNOrg/gfn-lm-tuning}}. Specifically, we fine-tune a \texttt{GPT-2 Large} model~\citep{radford2019language} on the stories dataset with full parameter fine-tuning using the \texttt{trl} library~\citep{vonwerra2022trl}.
We trained for 20 epochs with a batch size of 64 and 32 gradient accumulation steps and a learning rate of 0.0005.

\paragraph{Baselines.}
For the baselines, we adopt the implementations from~\citep{hu2023amortizing}. A critical difference in our experiments compared to~\citep{hu2023amortizing} is that the posterior model is not initialized with a base model that is fine-tuned on the stories dataset. To condition the model on $X$ and $Y$, as well as for the prompting baseline, we use the following prompt:

\texttt{"Beginning: \{X\}\textbackslash n End: \{Y\}\textbackslash n Middle: "}

During training for the autoregressive GFlowNet fine-tuning, a $(\rvx, \rvy)$ pair is sampled from the dataset and then sample (batch size) $\rvx$s for every $(X, Y)$, and $p_{\rm reward}(XZY)$ is used as the reward. Both the GFlowNet fine-tuning and supervised fine-tuning baseline use LoRA fine-tuning. We use the default hyperparamteres from~\citep{hu2023amortizing}.
At test time, we sample $100$ infills for each example in the test set from all the models at temperature $0.9$, and average over 5 such draws.

\begin{table}[]
    \centering
    \caption{Examples of training samples for the language infilling task.}
    \begin{tabular}{p{0.4\textwidth} p{0.25\textwidth}p{0.25\textwidth}}
        \toprule
        \textbf{Beginning} ($\rvx$) & \textbf{Middle} ($\rvz$) & \textbf{End} ($\rvy$)\\\midrule
        I was going to a Halloween party. I looked through my clothes but could not find a costume. I cut up my old clothes and constructed a costume. & I put my costume on and went to the party. & My friends loved my costume. \\ \midrule
        Allen thought he was a very talented poet. He attended college to study creative writing. In college, he met a boy named Carl. & Carl told him that he wasn't very good. & Because of this, Allen swore off poetry forever.\\
        
         \bottomrule
    \end{tabular}
    \label{tab:examples_infilling}
\end{table}

\begin{table}[]
    \centering
    \caption{Hyperparameters for the story infilling task.}
    \begin{tabular}{lc}
        \toprule
         Batch size & 16 \\
         Gradient accumulation steps & 8 \\
         Learning rate & 1e-5 \\
         Warmup Step & 20 \\
         Optimizer & AdamW \\
         Reward temperature start & 1.2 \\
         Reward inverse temperature end & 0.9 \\
         Reward inverse temperature horizon & 5000 \\
         Number of training steps & 1500 \\
         Loss clipping coefficient & 0.1 \\
         Discretization steps $T$ & 15 \\
         \bottomrule
    \end{tabular}
    \label{tab:hps_infill}
\end{table}

\paragraph{Additional results.}

\autoref{tab:infill_gfn}, \autoref{tab:infill_prior_x} and \autoref{tab:infill_prior_xy} illustrates some examples of the infills generated by the diffusion models. We note that the general quality of the samples is poor, due to a relatively weak prior. At the same time we can observe that the prompting baselines often generate infills that are unrelated to the current story. We also note that the RTB fine-tuned model can sometimes generate repitions as the reward model tends to assign high likelihood to repititions~\citep{welleck-etal-2020-consistency}. 
We also attempted a LLMEval~\cite{liu2023gpteval} for evaluating the coherence of the stories but did not obtain statistically significant results. 
\begin{table}
    \centering
    \caption{{Examples of infills generated by the posterior trained with RTB along with \textbf{reference infills} for the stories infilling task.} %The reference infill from the dataset is in \textbf{bold}.
    }
\begin{tabular}{p{0.3\textwidth} p{0.4\textwidth}p{0.2\textwidth}}
\toprule
\textbf{Beginning} ($\rvx$) & \textbf{Middle} ($\rvz$) & \textbf{End} ($\rvy$) \\ \midrule
\multirow{5}{0.3\textwidth}{David noticed he had put on a lot of weight recently. He examined his habits to try and figure out the reason. He realized he'd been eating too much fast food lately.} & \textbf{He stopped going to burger places and started a vegetarian diet.} & \multirow{5}{0.2\textwidth}{After a few weeks, he started to feel much better.} \\ 
& He reviewed his habits to try to figure out how to change & \\ 
& He asked he thought try to cut down on the amount amount. & \\ 
& He examined his habits to try and figure out the reason. & \\ 
& He realized he had been eating too much fast food recently. & \\  
\midrule\multirow{5}{0.3\textwidth}{Robbie was competing in a cross country meet. He was halfway through when his leg cramped up. Robbie wasn't sure he could go on.} & \textbf{He stopped for a minute and stretched his bad leg.} & \multirow{5}{0.2\textwidth}{Robbie began to run again and finished the race in second place.} \\ 
& Robbie was sure he could go on. Robbie was sure. & \\ 
& He was floating and twisting his leg in half then. & \\ 
& His body just caught up with his legs. Robbie was.  & \\ 
&He held his leg forward as he went through and his & \\ 
\bottomrule\end{tabular}
    \label{tab:infill_gfn}
\end{table}

\begin{table}
    \centering
    \caption{{Examples of infills generated by Prompt $(\rvx,\rvy)$ along with \textbf{reference infills} for the stories infilling task.} %The reference infill from the dataset is in \textbf{bold}.
    }
\begin{tabular}{p{0.3\textwidth} p{0.4\textwidth}p{0.2\textwidth}}
\toprule
\textbf{Beginning} ($\rvx$) & \textbf{Middle} ($\rvz$) & \textbf{End} ($\rvy$) \\ \midrule
\multirow{5}{0.3\textwidth}{David noticed he had put on a lot of weight recently. He examined his habits to try and figure out the reason. He realized he'd been eating too much fast food lately.} & \textbf{He stopped going to burger places and started a vegetarian diet.} & \multirow{5}{0.2\textwidth}{After a few weeks, he started to feel much better.} \\ 
& He'd had less opportunities to eat properly all of last. & \\ 
& Doctors made the note of the situation. He was treated. & \\ 
& He told him the guy for a mic replacement.\verb|\n\n| & \\ 
& He felt empty for one reason and new fresh, too. & \\ 
\midrule\multirow{5}{0.3\textwidth}{Robbie was competing in a cross country meet. He was halfway through when his leg cramped up. Robbie wasn't sure he could go on.} & \textbf{He stopped for a minute and stretched his bad leg.} & \multirow{5}{0.2\textwidth}{Robbie began to run again and finished the race in second place.} \\ 
& Robbie wasn't sure Robbie's fuel tank was full. & \\ 
& Robbie took a photograph with a close friend.\verb|\n\n| & \\ 
& Only Stacey Ebers and Rand were out there. & \\ 
& Robbie got bigger as the position got better.\verb|\n\n| & \\ 
\bottomrule\end{tabular}
    \label{tab:infill_prior_x}
\end{table}

\begin{table}
    \centering
    \caption{{Examples of infills generated by Prompt $(\rvx)$ along with \textbf{reference infills} for the stories infilling task.} %The reference infill from the dataset is in \textbf{bold}.
    }
\begin{tabular}{p{0.3\textwidth} p{0.4\textwidth}p{0.2\textwidth}}
\toprule
\textbf{Beginning} ($\rvx$) & \textbf{Middle} ($\rvz$) & \textbf{End} ($\rvy$) \\ \midrule
\multirow{5}{0.3\textwidth}{David noticed he had put on a lot of weight recently. He examined his habits to try and figure out the reason. He realized he'd been eating too much fast food lately.} & \textbf{He stopped going to burger places and started a vegetarian diet.} & \multirow{5}{0.2\textwidth}{After a few weeks, he started to feel much better.} \\ 
& David, "All I had told eat was a problem. & \\ 
& He got the backside what about that and he made the, & \\ 
& He made just good of fast food and spliced it down. & \\ 
& He explained everything to them, reached them out, the problem. & \\ 
\midrule\multirow{5}{0.3\textwidth}{Robbie was competing in a cross country meet. He was halfway through when his leg cramped up. Robbie wasn't sure he could go on.} & \textbf{He stopped for a minute and stretched his bad leg.} & \multirow{5}{0.2\textwidth}{Robbie began to run again and finished the race in second place.} \\ 
& Robbie and Robbie was piling. Robbie and Robbie fistfight. & \\ 
& I said goodbye. Robbie at dinner. Robbie agreed with. & \\ 
& He cut away a little to Robbie's pace fleetingly. & \\ 
& He held off all the police and place. Robbie. & \\ 
\bottomrule\end{tabular}
    \label{tab:infill_prior_xy}
\end{table}

\section{Offline RL}
\label{app:offline_rl}

\subsection{Training details}

Our method requires first training a diffusion-based behavior policy $\pi_{\theta}$ and a Q-function $Q_{\psi}$. Once $\pi_{\theta}$ and $Q_{\psi}$ are trained, The posterior policy $\pi_{\gamma}$ is trained using RTB, with its weights initialized to the trained behavior policy weights $\theta$. 

The behavior policy $\pi_{\theta}$ is parametrized as a state-conditioned noise-predicting denoising diffusion probabilistic model (DDPM) \cite{ho2020ddpm} with a linear schedule, and 75 denoising steps. The diffusion model takes as input a state $s$, a noised action $a_t$ and a noise level $t$ and predicts the source noise $\epsilon$. The state $s$ and noised action $a_t$ are concatenated with Fourier features computed on the noise level $t$, which are then fed through a 3-layer MLP of hidden dimensionality 256, with layer normalization and a GeLU activation after each hidden layer. The behavior policy is trained using the Adam optimizer with batch size 512 and learning rate 5e-4 for 10000 epochs. The Q-function $Q_{\psi}$ is trained using IQL. We use the same IQL experimental configurations and training hyperparameters as in \cite{kostrikov2022offline}. That is, we set $\tau = 0.7$. The architecture for $Q_{\psi}$ is a 3-layer MLP with hidden dimensionality 256 and  ReLU activations, which is trained using the Adam optimizer with a learning rate 3e-4 and batch size 256 for 750000 gradient steps. The task rewards are normalized as in \citep{kostrikov2022offline} and the target network is updated with soft updates of $m = 0.005$. The posterior policy $\pi_{\gamma}$ is trained using the relative trajectory balance objective. $\pi_{\gamma}$ is also parametrized as a state-conditioned noise-predicting DDPM, initialized as a copy of the prior. We additionally use the Langevin dynamics inductive bias (\ref{eq:langevin}), and learn an additional MLP for the energy scaling network. The posterior noise prediction network also outputs an additive correction to the output of the prior noise prediction network. That is, the predicted noise of the posterior diffusion model is defined as $\epsilon(s, a_t, t) := \epsilon(s, a_t, t; \theta) + \epsilon(s, a_t, t; \gamma)$, where $\epsilon(\cdot; \theta)$ is the output of the prior noise prediction network and $\epsilon(\cdot; \gamma)$ is the output of the posterior noise prediction network. We train all models on a single NVIDIA A100-Large GPU. The only hyperparameter tuned per task is the temperature $\alpha$ which we show in \autoref{tab:offline_beta}.

Note that in these experiments both the prior and constraint are conditioned on the state $\rvs$. To prevent having to learn a neural network for $\log Z_{\phi}(\rvs)$, we employ a variant of VarGrad objective~\citep{richter2020vargrad}. For each state $\rvs$ sampled in the minibatch, we further generate $k=64$ on-policy trajectories ${\tau^{(i)}}_{i=1}^k$ with $\pi_{\gamma}$. Each of these trajectories can be used to implicitly estimate $\log Z(\rvs)$:
\begin{equation}
    \hat{\log Z(\rvs)}^{(i)} = \log \pi_{\theta}(\tau^{(i)} \mid \rvs) + Q_{\psi}(\rvs, \rva_1^{(i)}) - \log \pi_{\gamma}(\tau^{(i)} \mid \rvs)
\end{equation}
We then minimize the sample variance across the batch:
\begin{equation}\label{eq:vargrad_rtb}
    \LRTB^{\rm VarGrad}(\gamma) = \frac{1}{k}\sum_{i=1}^k\Bigl(\hat{\log Z(\rvs)}^{(i)} - \frac{1}{k}\sum_{j=1}^k \hat{\log Z(\rvs)}^{(j)}\Bigr)^2
\end{equation}

\begin{table}[]
    \centering
    \caption{Mixed vs. online training on DR4L Tasks. We report mean\std{\text{std}} over 5 random seeds.}
    \begin{tabular}{lcc}
        \toprule
         Task & RTB (Online) & RTB (Mixed) \\
         \midrule
         halfcheetah-medium-replay & 46.88\std{0.51} & \highlight{48.11\std{0.56}} \\
         hopper-medium-replay & 99.23\std{3.22} & \highlight{100.40\std{0.21}} \\
         walker2d-medium-replay & \highlight{94.01\std{0.28}} & 93.57\std{2.63} \\
         \bottomrule
    \end{tabular}
    \label{tab:mixedvsonline}
\end{table}

RTB allows off-policy training so we are not restricted to train with samples generated on-policy. We thus also leverage the offline dataset, which are samples from the prior and noise them with the DDPM noising process to generate off-policy trajectories with high density under the prior. Since there are actions in the replay buffer from high reward episodes in the tasks, this can help training efficiency compared to purely online training. We ran 5 seeds of training each with mixed training (off-policy and on-policy) and pure on-policy training on the medium-replay tasks, with results shown in Table \ref{tab:mixedvsonline}, where mixed training outperforms pure online training on two of the three tasks. 

\begin{table}[]
    \centering
    \caption{Temperature $\alpha = \frac{1}{\beta}$ for D4RL tasks}
    \begin{tabular}{lc}
        \toprule
         Task & $\alpha$ \\
         \midrule
         halfcheetah-medium-expert & 0.1 \\
         hopper-medium-expert & 0.5 \\
         walker2d-medium-expert & 0.1 \\
         halfcheetah-medium & 0.05 \\
         hopper-medium & 0.1 \\
         walker2d-medium & 0.05 \\
         halfcheetah-medium-replay & 0.05 \\
         hopper-medium-replay & 0.05 \\
         walker2d-medium-replay & 0.1 \\
         \bottomrule
    \end{tabular}
    \label{tab:offline_beta}
\end{table}

\subsection{Baseline details}
As is standard in offline RL, we use the reported performance numbers from the previous papers. CQL, IQL are reported from the IQL paper. Diffuser (D), DD, D-QL and QGPO are reported from the QGPO paper. Their implementation improved the performance of D and D-QL compared to their original papers. IDQL results are reported from the IDQL paper. We follow the evaluation protocol of previous work, and report the mean performance over 10 episodes, averaged across 5 random seeds at the end of training (150k training steps).

\clearpage
\section{Fine-tuning text-to-image diffusion models}
\label{app:text2image}
We build off the DPOK implementation\footnote{\url{https://github.com/google-research/google-research/tree/master/dpok}}, which fine-tunes stable-diffusion-v1-5 with ImageReward function. The posterior model to be fine-tuned is initialized as a copy of the prior model. We use LoRA~\citep{hu2022lora} since it is significantly more efficient than fine-tuning the entire model. Sampling of images is done with 50 steps of DDIM~\citep{song2021denoising}. Even with LoRA, it is still difficult to fit gradients of all steps in the diffusion trajectory in memory. To help with this, we use a ``stochastic subsampling'' trick (\autoref{app:efficient}). 

We train all models on a single NVIDIA A100-Large GPU. For the main experiments, we use the default parameters for DPOK of reward weight $\beta=10$ and KL weight $= 0.01$. For RTB we fix $\beta=1.0$ for all prompts. We next perform an ablation over different values of $\beta$.

We plot in \autoref{tab:betaablation} the final average reward and diversity score for models trained with different values of reward weight $\beta$ for the prompt ``A green colored rabbit.''. As expected, we find that increasing $\beta$ increases reward at the cost of diversity for RTB and DPOK. The exception is $\beta=10$ for RTB which has slightly lower final reward than $\beta=1$, which we could attribute to more difficult optimization due to the peaky distribution associated with higher reward weight.

\begin{table}[htbp]
\centering
\caption{Ablation of reward weights $\beta$ for ``A green colored rabbit.''.}
\label{tab:table1}
\resizebox{0.6\linewidth}{!}{
\begin{tabular}{llcccc}
    \toprule
     Model $\downarrow$ & $\beta$ $\downarrow$ Metric $\rightarrow$ & Reward ($\uparrow$) & diversity ($\uparrow$)\\
    \midrule
    \multirow{1}{*}{Prior}& - &  -0.113 & 0.597 \\
    \midrule
     \multirow{4}{*}{DPOK (KL weight=0.01)} & $\beta=0.01$ & -0.27 & 0.1488 \\
      & $\beta=0.1$ & -0.06 & 0.1486 \\
      & $\beta=1.0$ & 0.638 & 0.1362  \\
       & $\beta=10.0$ & 1.492 & 0.076 \\
    \midrule
    \multirow{1}{*}{DDPO (KL weight=0.0)} & $\beta=10.0$ & 1.795 & 0.0493 \\
    \midrule
    \multirow{4}{*}{RTB} &
    $\beta=0.01$ & 0.485 & 0.1431  \\
      & $\beta=0.1$ & 1.525 & 0.0721  \\
      & $\beta=1.0$ & 1.756 & 0.0436 \\
       & $\beta=10.0$ & 1.568 & 0.0689 \\
     \bottomrule
     \label{tab:betaablation}
\end{tabular}
}
\end{table}

\subsection{Memory-efficient learning}
\label{app:efficient}

We propose two methods to reduce the memory requirement of RTB fine-tuning.

\paragraph{Stochastic subsampling.} The expected gradient of the RTB objective (\ref{eq:rtb_objective}) is unaffected by propagating gradient to a randomly sampled subset of the timesteps in a trajectory and rescaling by the inverse proportion of timesteps sampled. Stochastically subsampling timesteps for gradient propagation in this way can significantly decrease memory consumption because computation graphs for the remaining timesteps do not need to be maintained; however, such subsampling increases gradient variance, so it is preferable to keep gradients for as many timesteps as possible to fit in memory. For our text-to-image experiments, we found sampling 8 timesteps out of 50 to keep gradients was sufficient.

\paragraph{Batched gradient computation.}

An important property of the RTB objective is that computing its gradient does not require storing the computation graph of all timesteps. The gradient of the RTB objective for a single trajectory is just the sum of per-step log-likelihood gradients scaled by the RTB residual:
\[
 \nabla_{\phi}\mathcal{L}_{\rm RTB}(\tau;\phi)= 2\left(\log\frac{Z_{\phi}}{r(\rvx_1)} + \sum_{i=1}^T\log\frac{p_{\phi}^{post}(\rvx_{i\Delta t} \mid \rvx_{(i-1)\Delta t})}{p_{\theta}(\rvx_{i\Delta t} \mid \rvx_{(i-1)\Delta t})}\right) \cdot \nabla_{\phi}\sum_{i=1}^T\log p_{\phi}^{\rm post}(\rvx_{i\Delta t} \mid \rvx_{(i-1)\Delta t}).
\]
Because the likelihood gradients can be accumulated during the forward pass, this allows for a batched gradient accumulation version of the update. For trajectory length (number of diffusion steps) $T$ and accumulation batch size (number of time steps receiving a gradient signal in each backward pass) $B$, the number of batched forward passes required scales as $\frac TB$.

Only the accumulation batch size $B$, not the trajectory length $T$, is constrained by the memory budget. This means we can easily scale training with large number of diffusion steps without increasing the variance of the gradient through stochastic subsampling, with training time growing linearly with number of time steps under a fixed memory budget. Although this method is not used in the main experiments presented here, preliminary experiments confirm these observations.

We highlight that both methods are not applicable to diffusion samplers based on differentiable simulation (\eg, PIS and DDS), which need to store the entire computation graph of SDE integration. For these methods, the memory requirement scales linearly with the trajectory length.

\clearpage

\subsection{Generated images}
\label{app:promptimages}

\subsubsection{A green-colored rabbit}

\begin{figure}[h!]
\vspace{-1em}
    \centering
    \begin{minipage}{0.09\textwidth}
        \centering
        \includegraphics[width=\textwidth]{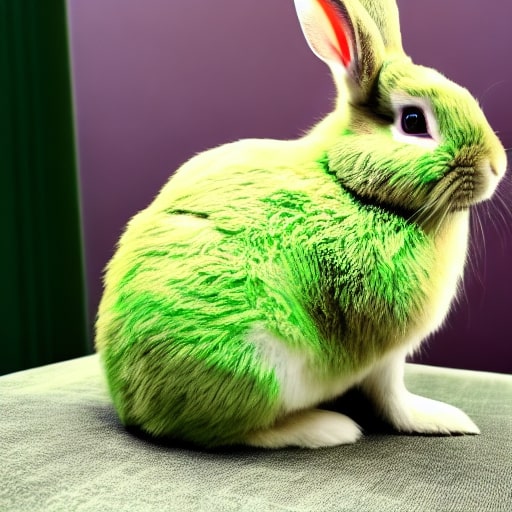} %
    \end{minipage}
    \begin{minipage}{0.09\textwidth}
        \centering
         \includegraphics[width=\textwidth]{figures/text-to-image-jpg/green_rabbit/prior/image1.jpg} %
    \end{minipage}
    \begin{minipage}{0.09\textwidth}
        \centering
         \includegraphics[width=\textwidth]{figures/text-to-image-jpg/green_rabbit/prior/image2.jpg} %
    \end{minipage}
    \begin{minipage}{0.09\textwidth}
        \centering
        \includegraphics[width=\textwidth]{figures/text-to-image-jpg/green_rabbit/prior/image3.jpg} 
    \end{minipage}
    \begin{minipage}{0.09\textwidth}
        \centering
         \includegraphics[width=\textwidth]{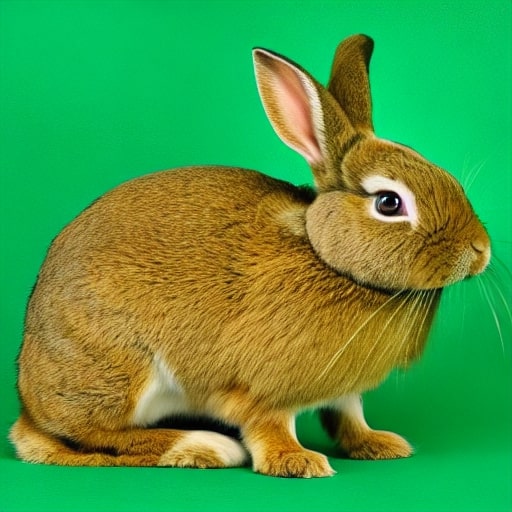}
    \end{minipage}
    \begin{minipage}{0.09\textwidth}
        \centering
       \includegraphics[width=\textwidth]{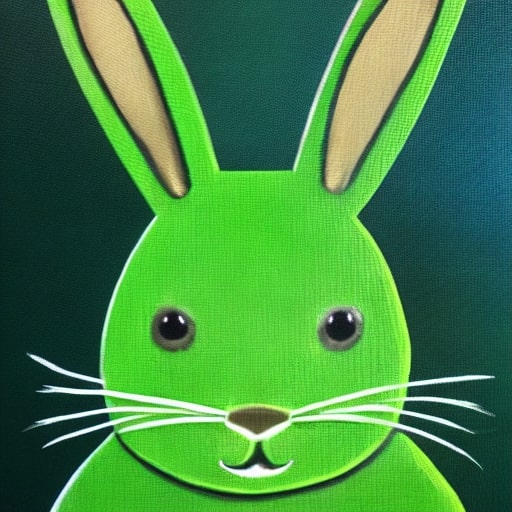}
    \end{minipage}
    \begin{minipage}{0.09\textwidth}
        \centering
    \includegraphics[width=\textwidth]{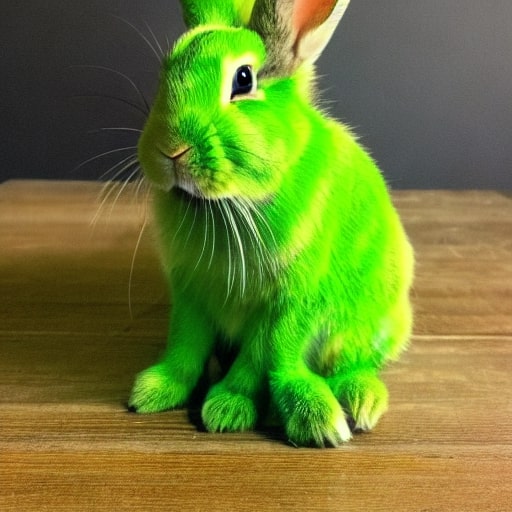}
    \end{minipage}
    \begin{minipage}{0.09\textwidth}
        \centering
 \includegraphics[width=\textwidth]{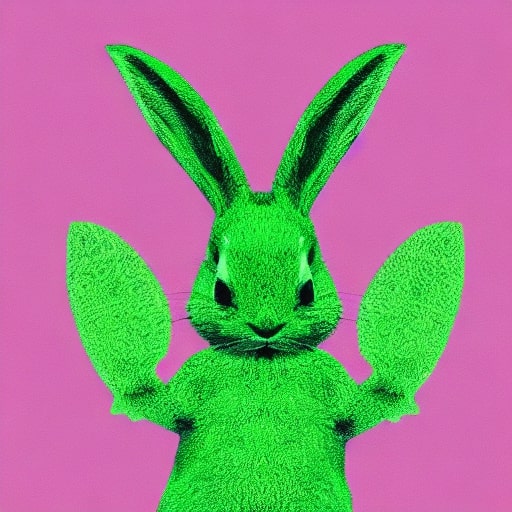}
    \end{minipage}
    \begin{minipage}{0.09\textwidth}
        \centering
 \includegraphics[width=\textwidth]{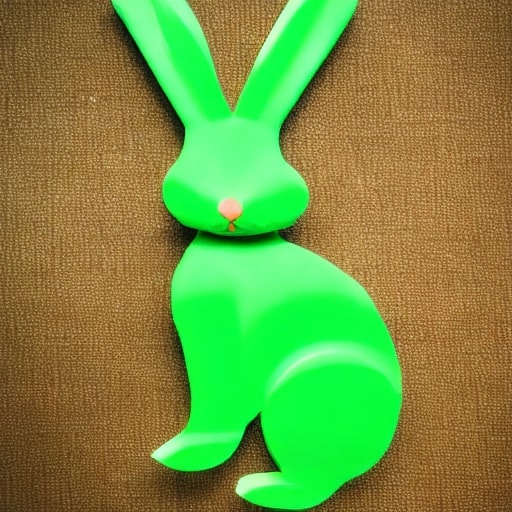}
    \end{minipage}
    \begin{minipage}{0.09\textwidth}
        \centering
  \includegraphics[width=\textwidth]{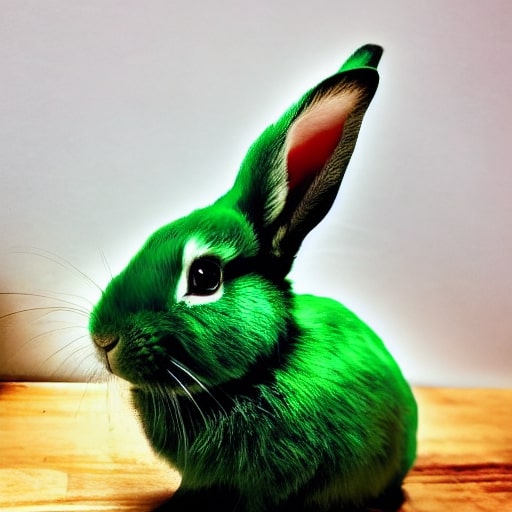}
    \end{minipage}
    \caption{Prior}
\end{figure}

\begin{figure}[h!]
\vspace{-1em}
    \centering
    \begin{minipage}{0.09\textwidth}
        \centering
        \includegraphics[width=\textwidth]{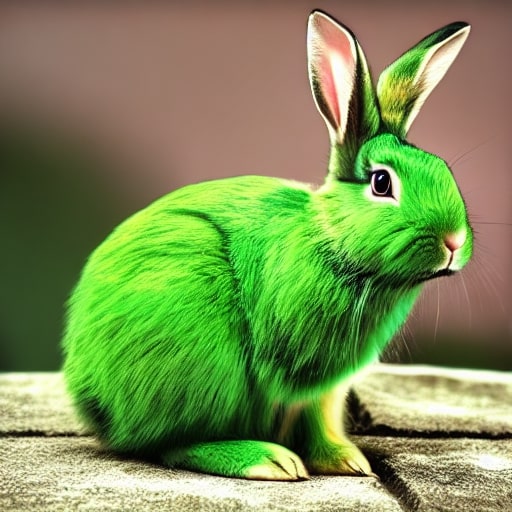} %
    \end{minipage}
    \begin{minipage}{0.09\textwidth}
        \centering
         \includegraphics[width=\textwidth]{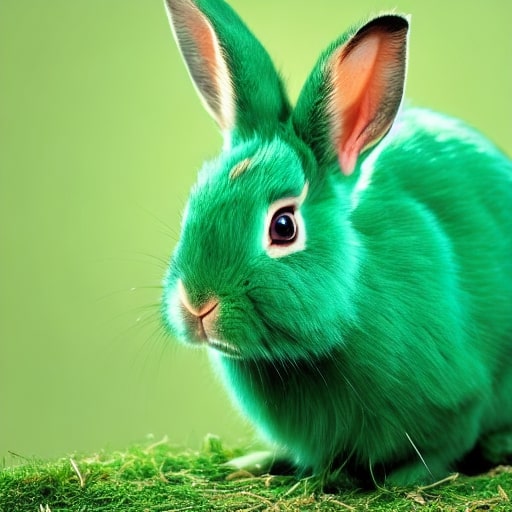} %
    \end{minipage}
    \begin{minipage}{0.09\textwidth}
        \centering
         \includegraphics[width=\textwidth]{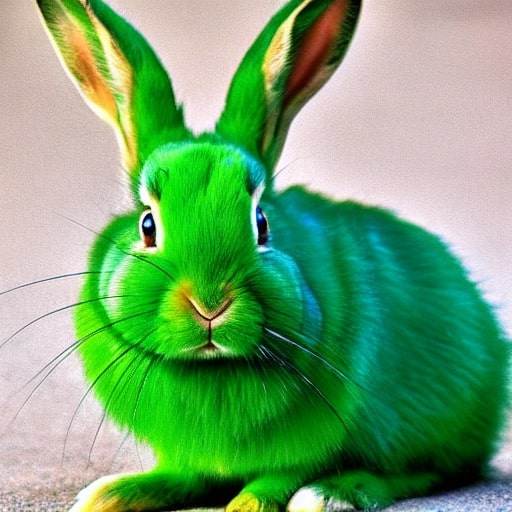} %
    \end{minipage}
    \begin{minipage}{0.09\textwidth}
        \centering
        \includegraphics[width=\textwidth]{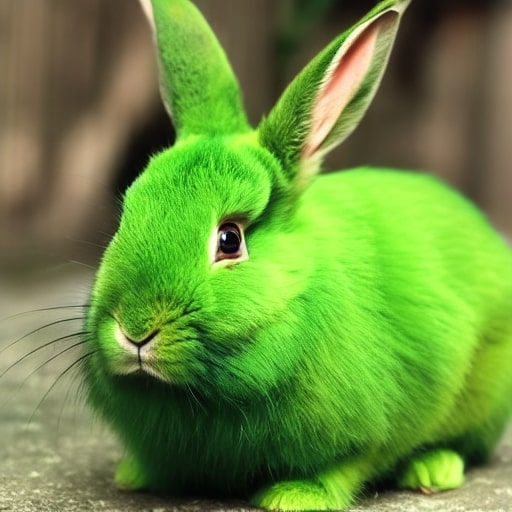} 
    \end{minipage}
    \begin{minipage}{0.09\textwidth}
        \centering
         \includegraphics[width=\textwidth]{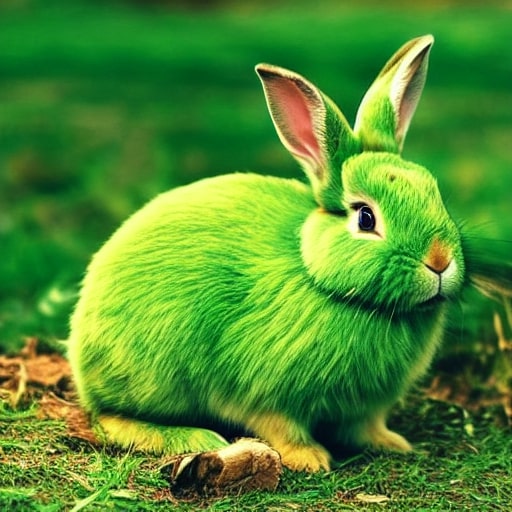}
    \end{minipage}
    \begin{minipage}{0.09\textwidth}
        \centering
       \includegraphics[width=\textwidth]{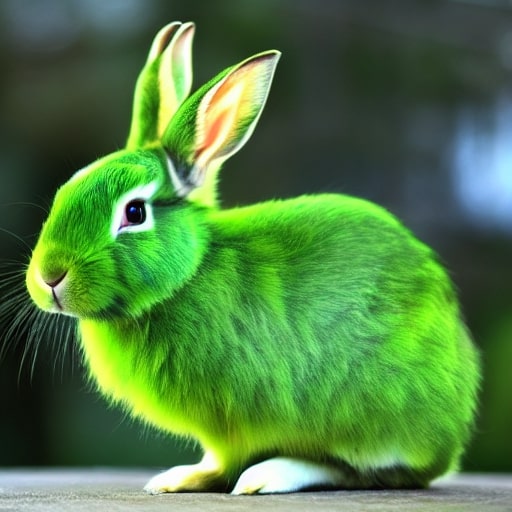}
    \end{minipage}
    \begin{minipage}{0.09\textwidth}
        \centering
    \includegraphics[width=\textwidth]{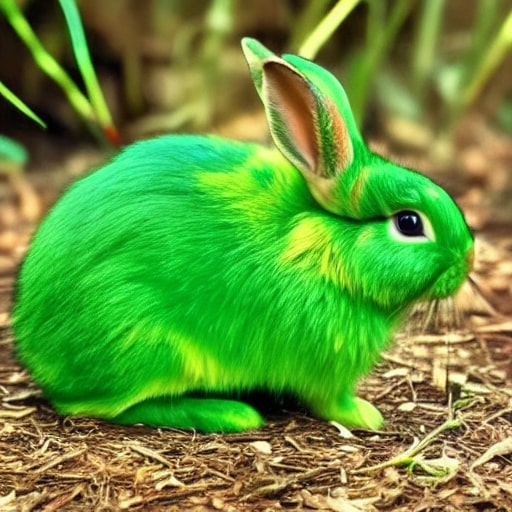}
    \end{minipage}
    \begin{minipage}{0.09\textwidth}
        \centering
 \includegraphics[width=\textwidth]{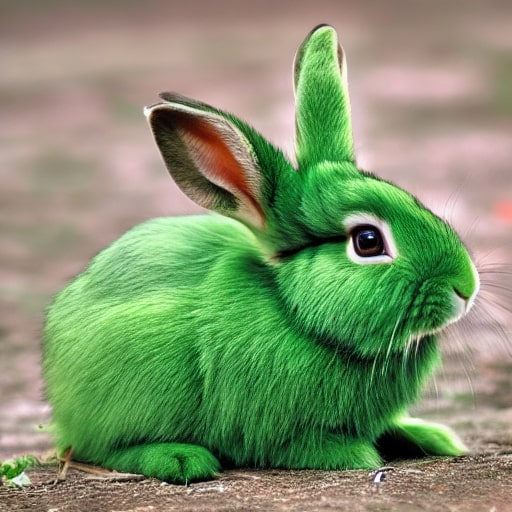}
    \end{minipage}
    \begin{minipage}{0.09\textwidth}
        \centering
 \includegraphics[width=\textwidth]{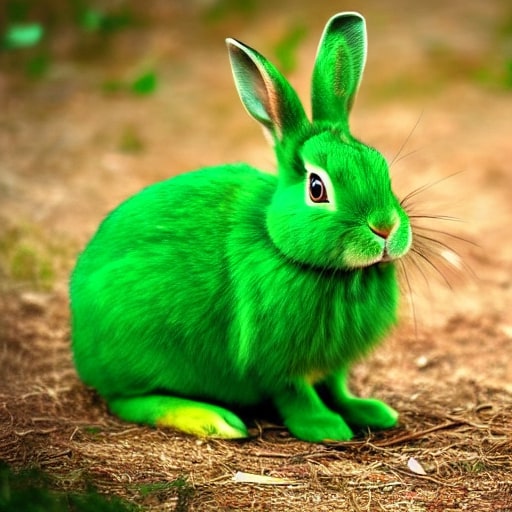}
    \end{minipage}
    \begin{minipage}{0.09\textwidth}
        \centering
  \includegraphics[width=\textwidth]{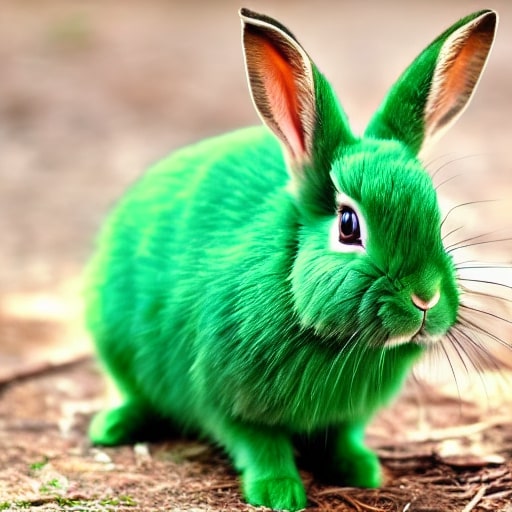}
    \end{minipage}
    \caption{DDPO}
\end{figure}

\begin{figure}[h!]
\vspace{-1em}
    \centering
    \begin{minipage}{0.09\textwidth}
        \centering
        \includegraphics[width=\textwidth]{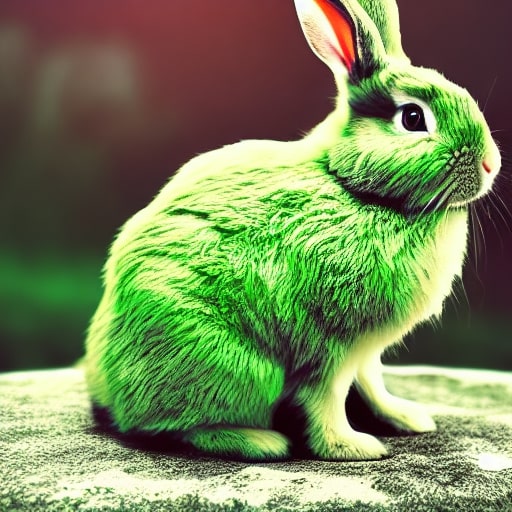} %
    \end{minipage}
    \begin{minipage}{0.09\textwidth}
        \centering
         \includegraphics[width=\textwidth]{figures/text-to-image-jpg/green_rabbit/dpok/image_1.jpg} %
    \end{minipage}
    \begin{minipage}{0.09\textwidth}
        \centering
         \includegraphics[width=\textwidth]{figures/text-to-image-jpg/green_rabbit/dpok/image_2.jpg} %
    \end{minipage}
    \begin{minipage}{0.09\textwidth}
        \centering
        \includegraphics[width=\textwidth]{figures/text-to-image-jpg/green_rabbit/dpok/image_3.jpg} 
    \end{minipage}
    \begin{minipage}{0.09\textwidth}
        \centering
         \includegraphics[width=\textwidth]{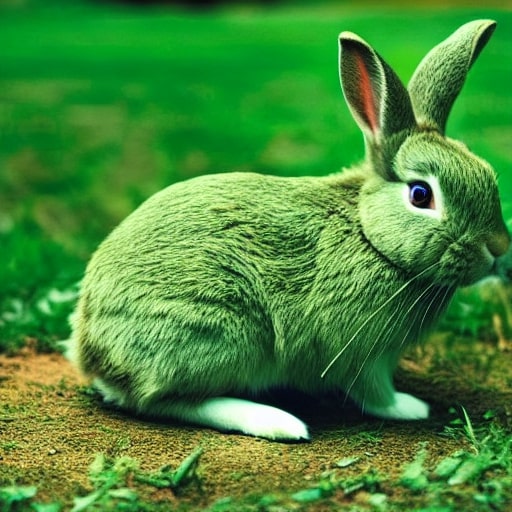}
    \end{minipage}
    \begin{minipage}{0.09\textwidth}
        \centering
       \includegraphics[width=\textwidth]{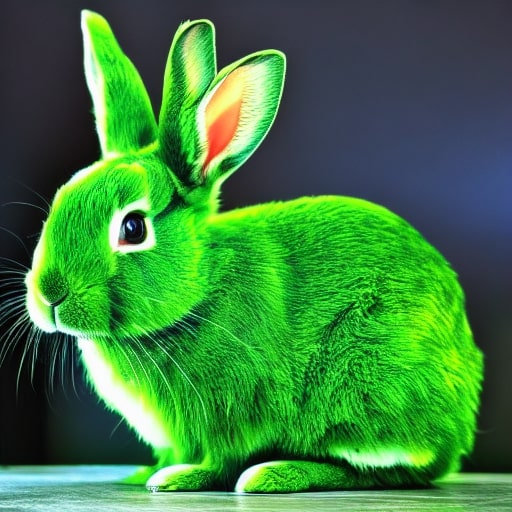}
    \end{minipage}
    \begin{minipage}{0.09\textwidth}
        \centering
    \includegraphics[width=\textwidth]{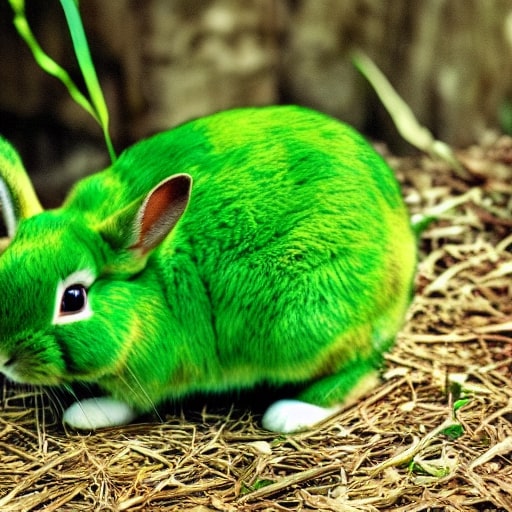}
    \end{minipage}
    \begin{minipage}{0.09\textwidth}
        \centering
 \includegraphics[width=\textwidth]{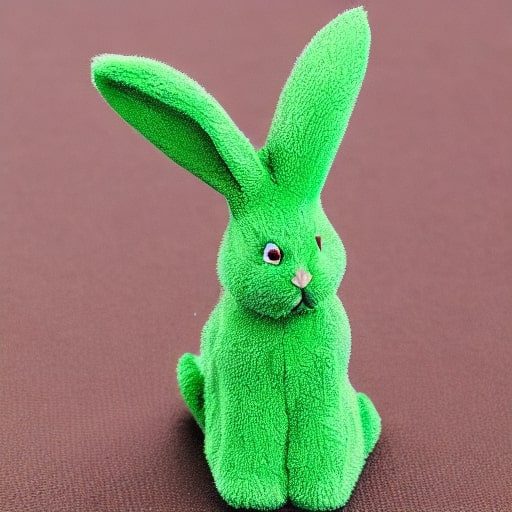}
    \end{minipage}
    \begin{minipage}{0.09\textwidth}
        \centering
 \includegraphics[width=\textwidth]{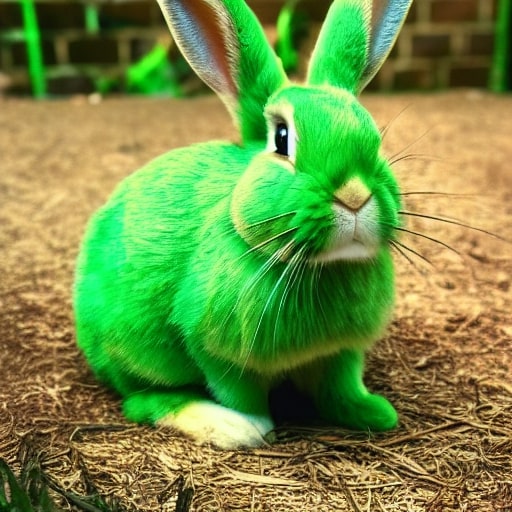}
    \end{minipage}
    \begin{minipage}{0.09\textwidth}
        \centering
  \includegraphics[width=\textwidth]{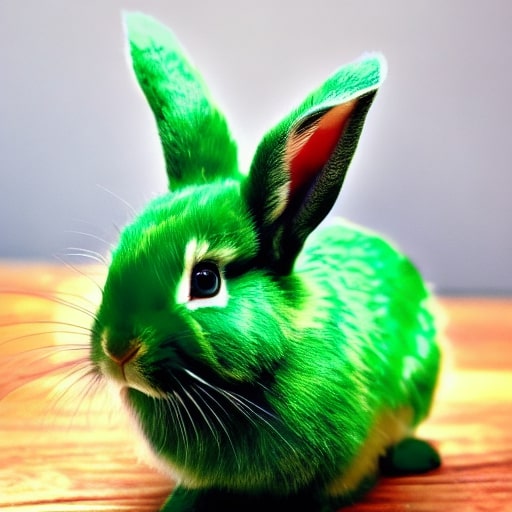}
    \end{minipage}
    \caption{DPOK}
\end{figure}

\begin{figure}[h!]
\vspace{-1em}
    \centering
    \begin{minipage}{0.09\textwidth}
        \centering
        \includegraphics[width=\textwidth]{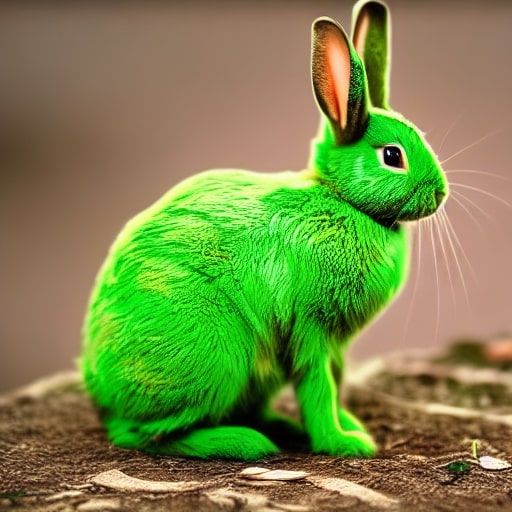} %
    \end{minipage}
    \begin{minipage}{0.09\textwidth}
        \centering
         \includegraphics[width=\textwidth]{figures/text-to-image-jpg/green_rabbit/gfn/image1.jpg} %
    \end{minipage}
    \begin{minipage}{0.09\textwidth}
        \centering
         \includegraphics[width=\textwidth]{figures/text-to-image-jpg/green_rabbit/gfn/image2.jpg} %
    \end{minipage}
    \begin{minipage}{0.09\textwidth}
        \centering
        \includegraphics[width=\textwidth]{figures/text-to-image-jpg/green_rabbit/gfn/image3.jpg} 
    \end{minipage}
    \begin{minipage}{0.09\textwidth}
        \centering
         \includegraphics[width=\textwidth]{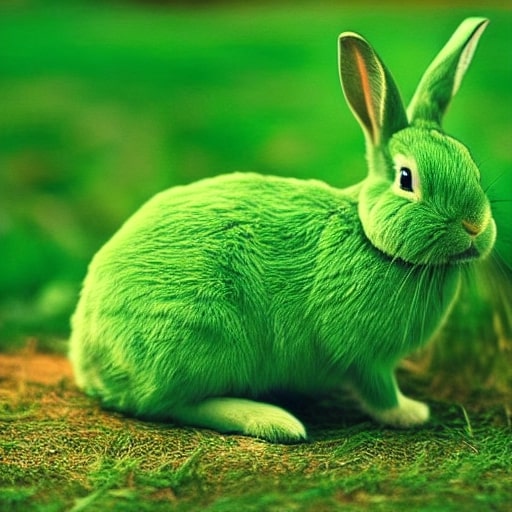}
    \end{minipage}
    \begin{minipage}{0.09\textwidth}
        \centering
       \includegraphics[width=\textwidth]{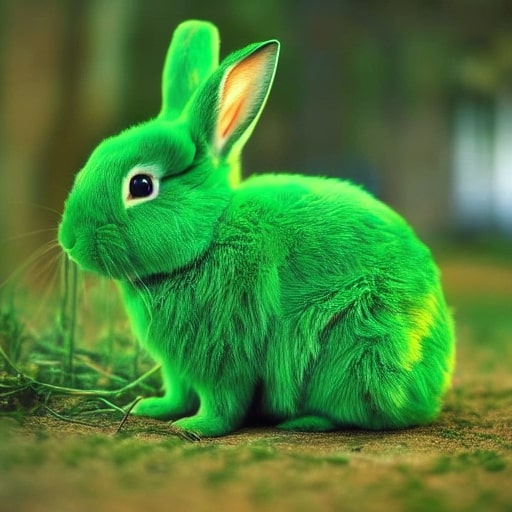}
    \end{minipage}
    \begin{minipage}{0.09\textwidth}
        \centering
    \includegraphics[width=\textwidth]{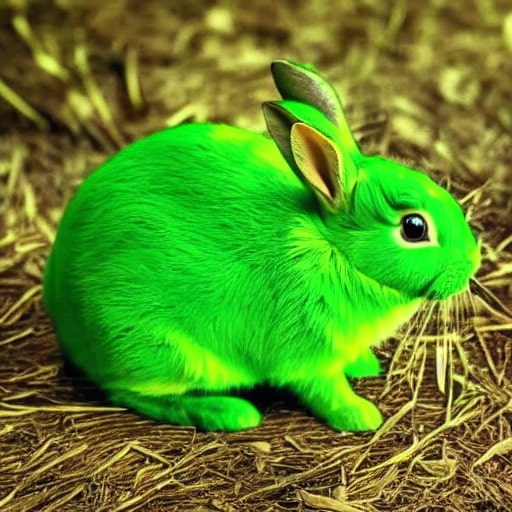}
    \end{minipage}
    \begin{minipage}{0.09\textwidth}
        \centering
 \includegraphics[width=\textwidth]{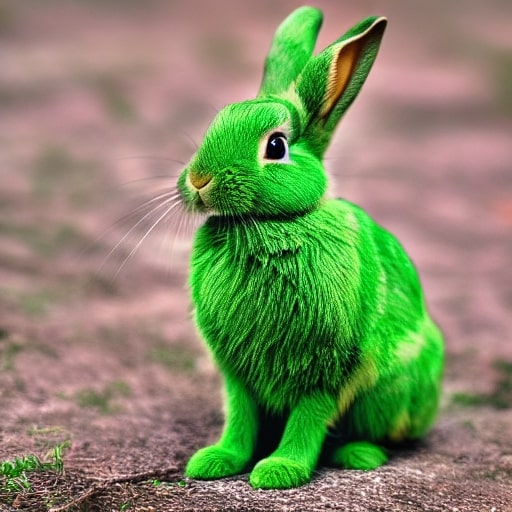}
    \end{minipage}
    \begin{minipage}{0.09\textwidth}
        \centering
 \includegraphics[width=\textwidth]{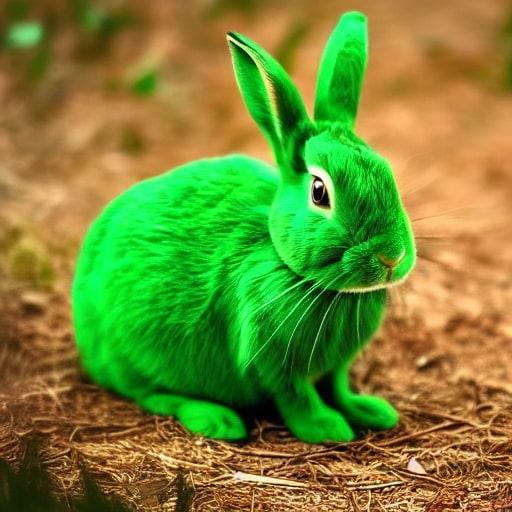}
    \end{minipage}
    \begin{minipage}{0.09\textwidth}
        \centering
  \includegraphics[width=\textwidth]{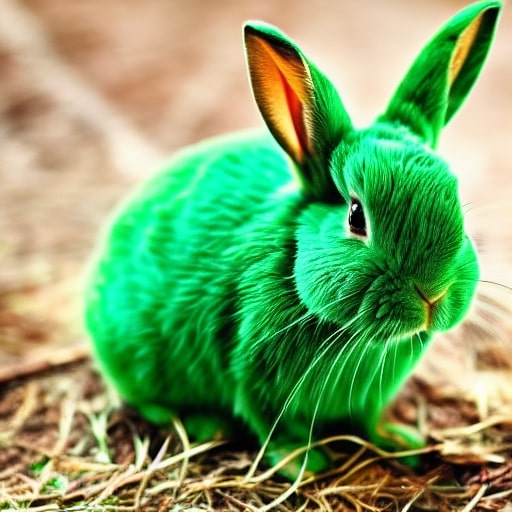}
    \end{minipage}
    \caption{RTB}
\end{figure}

\vspace{-2em}
\subsubsection{Four roses}

\begin{figure}[h!]
\vspace{-1em}
    \centering
    \begin{minipage}{0.09\textwidth}
        \centering
        \includegraphics[width=\textwidth]{figures/text-to-image-jpg/four_roses/prior/image_0.jpg} %
    \end{minipage}
    \begin{minipage}{0.09\textwidth}
        \centering
         \includegraphics[width=\textwidth]{figures/text-to-image-jpg/four_roses/prior/image_1.jpg} %
    \end{minipage}
    \begin{minipage}{0.09\textwidth}
        \centering
         \includegraphics[width=\textwidth]{figures/text-to-image-jpg/four_roses/prior/image_2.jpg} %
    \end{minipage}
    \begin{minipage}{0.09\textwidth}
        \centering
        \includegraphics[width=\textwidth]{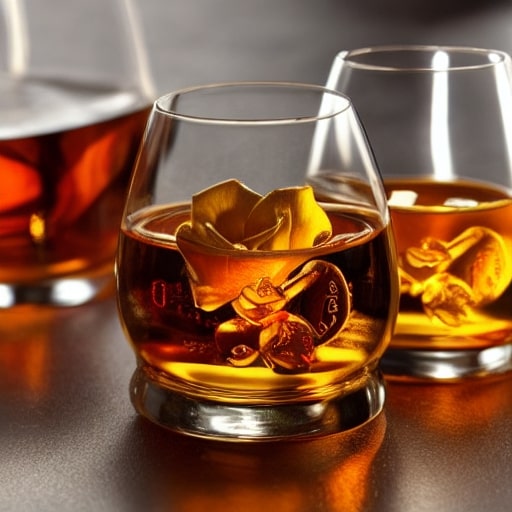} 
    \end{minipage}
    \begin{minipage}{0.09\textwidth}
        \centering
         \includegraphics[width=\textwidth]{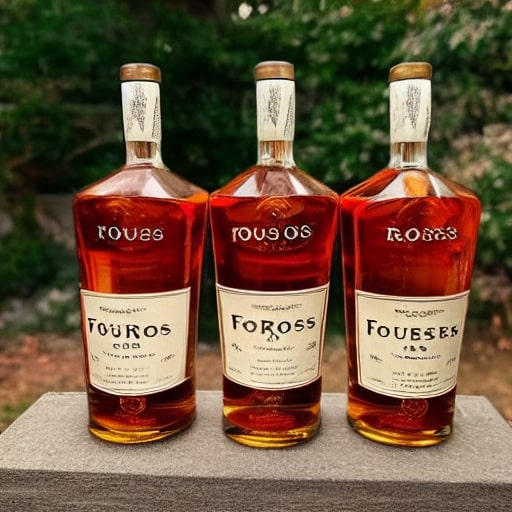}
    \end{minipage}
    \begin{minipage}{0.09\textwidth}
        \centering
       \includegraphics[width=\textwidth]{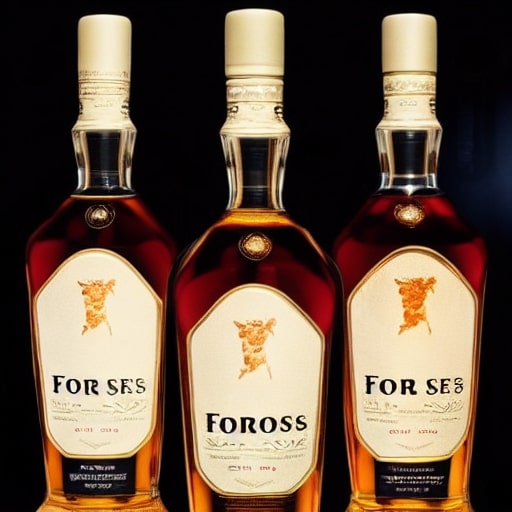}
    \end{minipage}
    \begin{minipage}{0.09\textwidth}
        \centering
    \includegraphics[width=\textwidth]{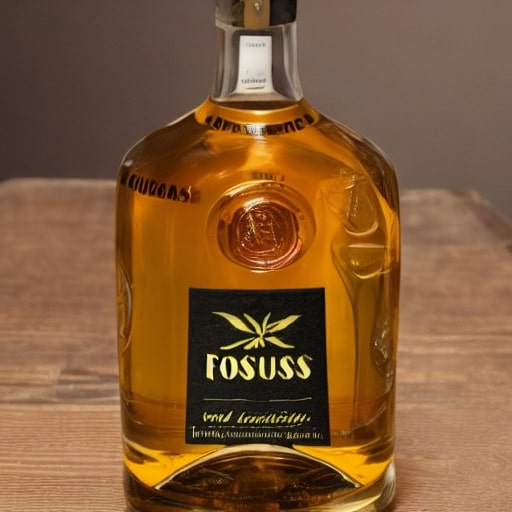}
    \end{minipage}
    \begin{minipage}{0.09\textwidth}
        \centering
 \includegraphics[width=\textwidth]{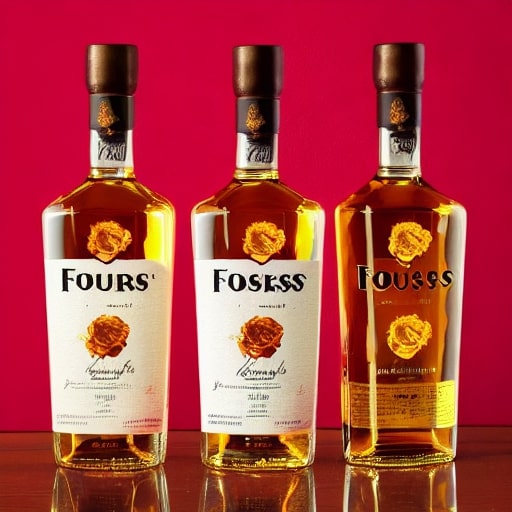}
    \end{minipage}
    \begin{minipage}{0.09\textwidth}
        \centering
 \includegraphics[width=\textwidth]{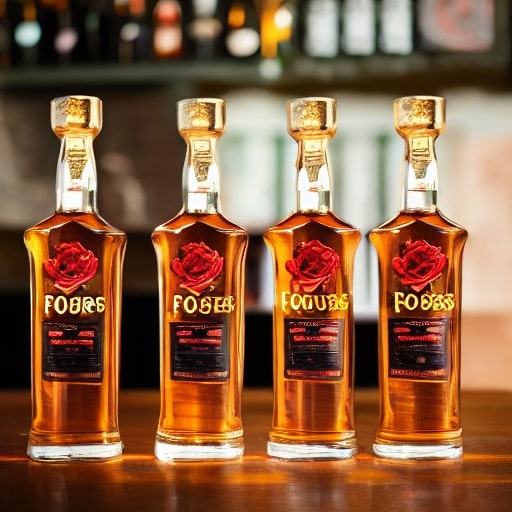}
    \end{minipage}
    \begin{minipage}{0.09\textwidth}
        \centering
  \includegraphics[width=\textwidth]{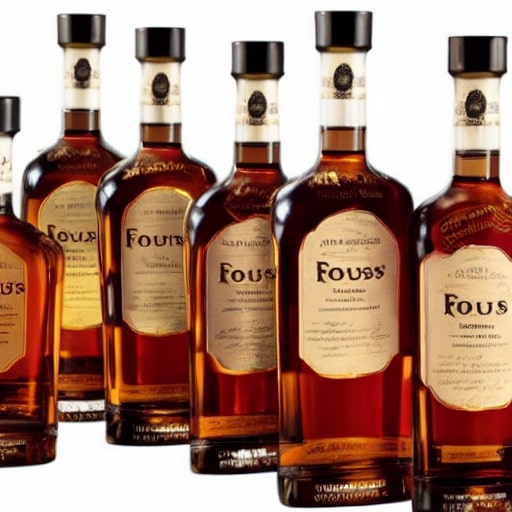}
    \end{minipage}
    \caption{Prior}
\end{figure}

 \begin{figure}[h!]
 \vspace{-1em}
     \centering
     \begin{minipage}{0.09\textwidth}
         \centering
         \includegraphics[width=\textwidth]{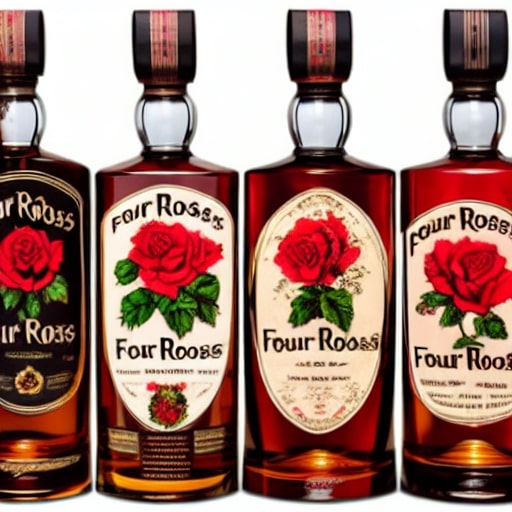} %
     \end{minipage}
     \begin{minipage}{0.09\textwidth}
         \centering
          \includegraphics[width=\textwidth]{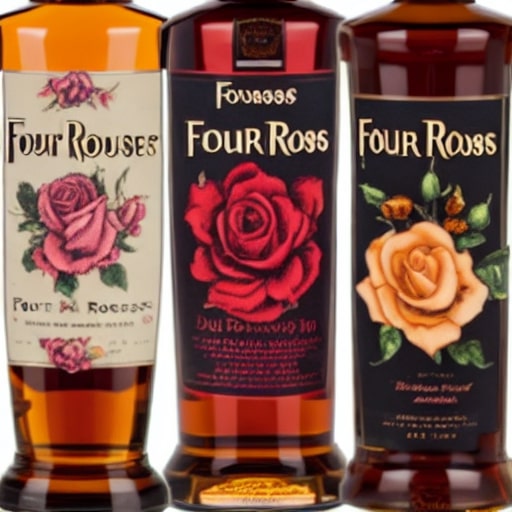} %
     \end{minipage}
     \begin{minipage}{0.09\textwidth}
         \centering
          \includegraphics[width=\textwidth]{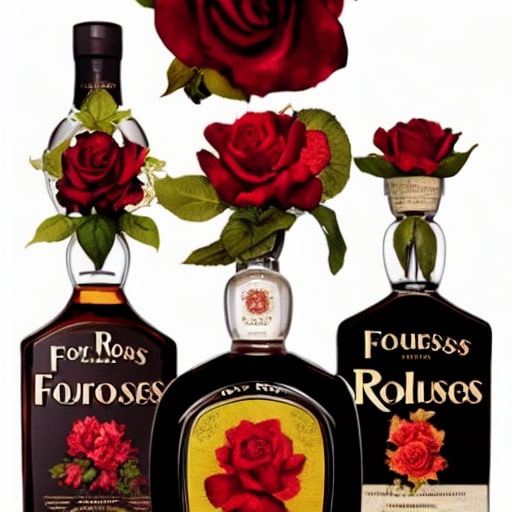} %
     \end{minipage}
     \begin{minipage}{0.09\textwidth}
         \centering
         \includegraphics[width=\textwidth]{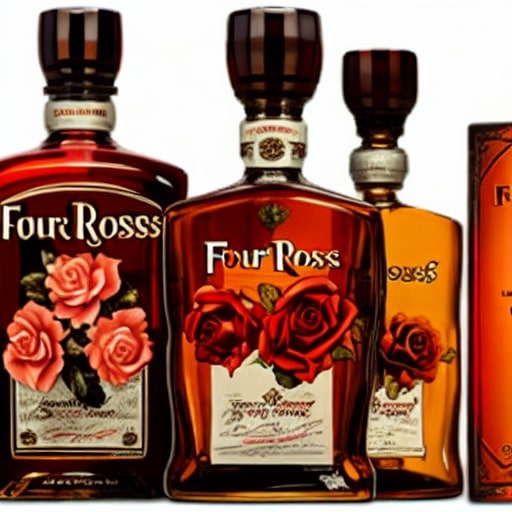} 
     \end{minipage}
     \begin{minipage}{0.09\textwidth}
         \centering
          \includegraphics[width=\textwidth]{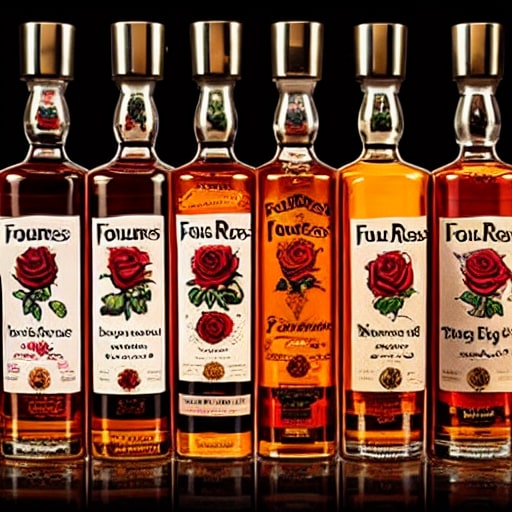}
     \end{minipage}
     \begin{minipage}{0.09\textwidth}
         \centering
        \includegraphics[width=\textwidth]{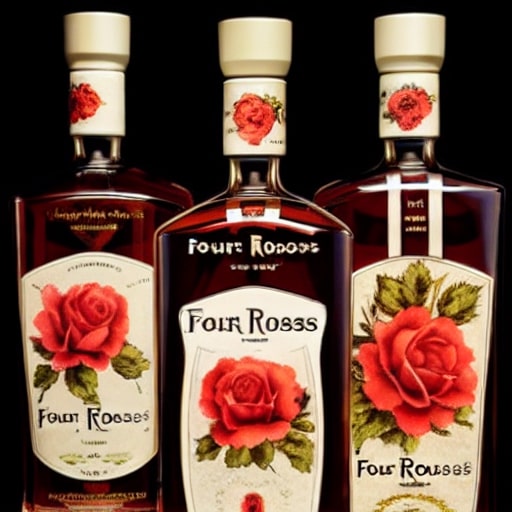}
     \end{minipage}
     \begin{minipage}{0.09\textwidth}
         \centering
     \includegraphics[width=\textwidth]{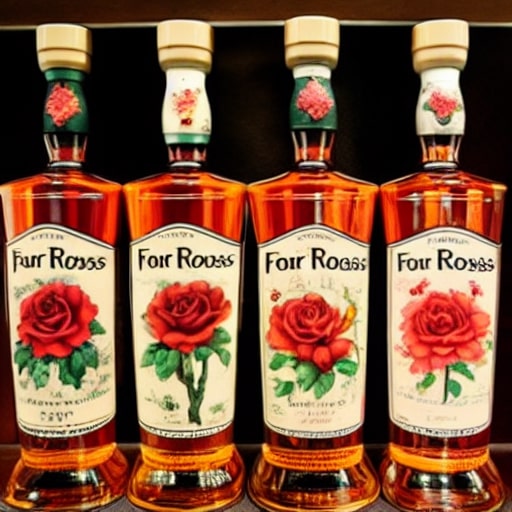}
     \end{minipage}
     \begin{minipage}{0.09\textwidth}
         \centering
  \includegraphics[width=\textwidth]{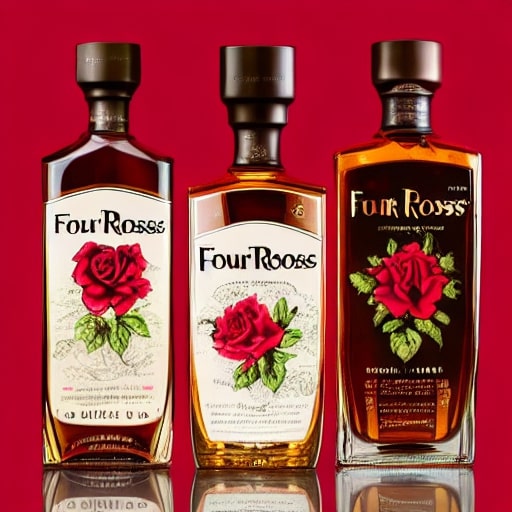}
     \end{minipage}
     \begin{minipage}{0.09\textwidth}
         \centering
  \includegraphics[width=\textwidth]{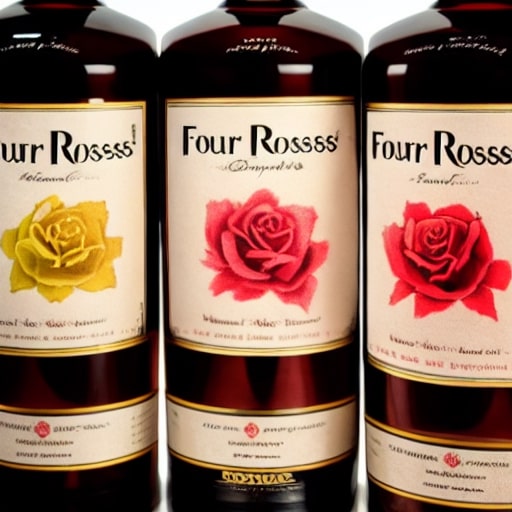}
     \end{minipage}
     \begin{minipage}{0.09\textwidth}
         \centering
   \includegraphics[width=\textwidth]{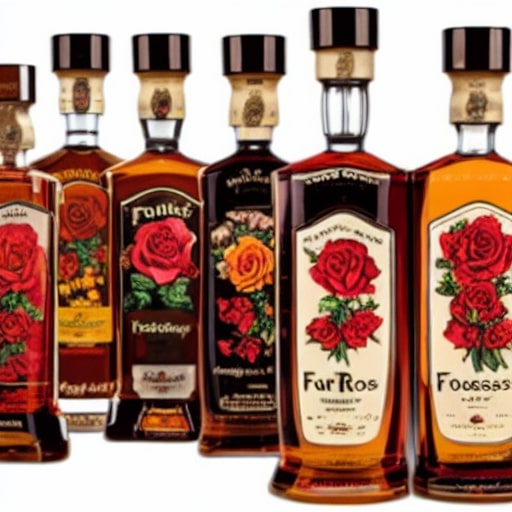}
     \end{minipage}
     \caption{DDPO}
 \end{figure}

\begin{figure}[h!]
\vspace{-1em}
    \centering
    \begin{minipage}{0.09\textwidth}
        \centering
        \includegraphics[width=\textwidth]{figures/text-to-image-jpg/four_roses/dpok/image_0.jpg} %
    \end{minipage}
    \begin{minipage}{0.09\textwidth}
        \centering
         \includegraphics[width=\textwidth]{figures/text-to-image-jpg/four_roses/dpok/image_1.jpg} %
    \end{minipage}
    \begin{minipage}{0.09\textwidth}
        \centering
         \includegraphics[width=\textwidth]{figures/text-to-image-jpg/four_roses/dpok/image_2.jpg} %
    \end{minipage}
    \begin{minipage}{0.09\textwidth}
        \centering
        \includegraphics[width=\textwidth]{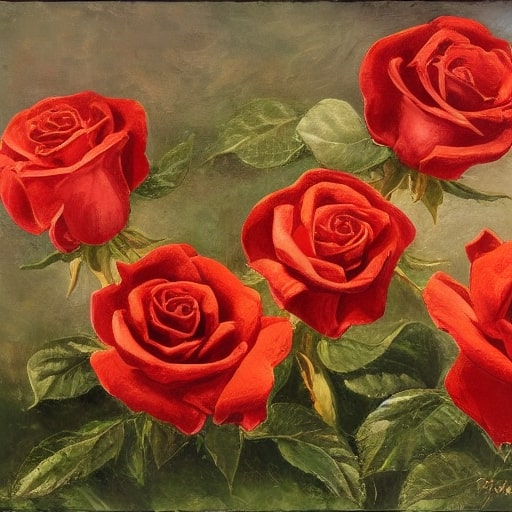} 
    \end{minipage}
    \begin{minipage}{0.09\textwidth}
        \centering
         \includegraphics[width=\textwidth]{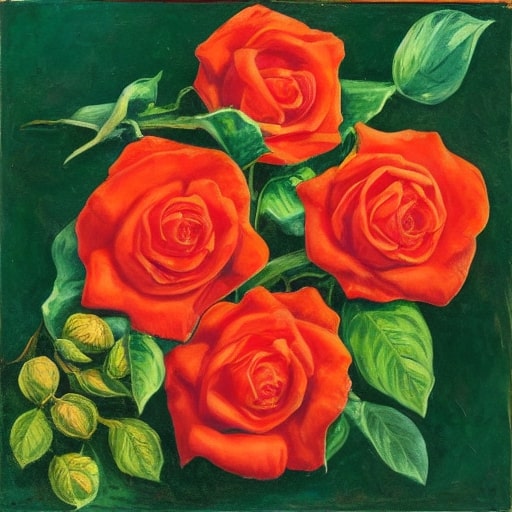}
    \end{minipage}
    \begin{minipage}{0.09\textwidth}
        \centering
       \includegraphics[width=\textwidth]{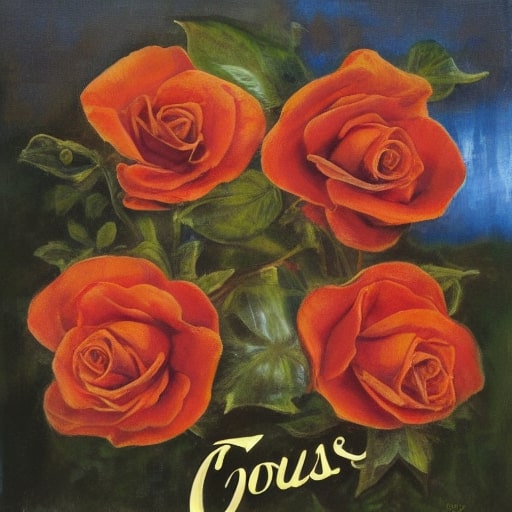}
    \end{minipage}
    \begin{minipage}{0.09\textwidth}
        \centering
    \includegraphics[width=\textwidth]{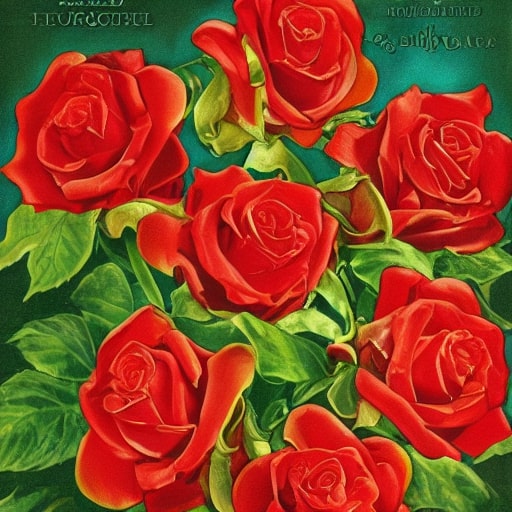}
    \end{minipage}
    \begin{minipage}{0.09\textwidth}
        \centering
 \includegraphics[width=\textwidth]{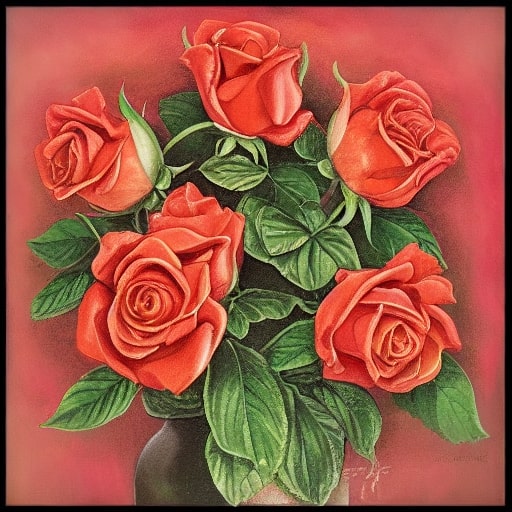}
    \end{minipage}
    \begin{minipage}{0.09\textwidth}
        \centering
 \includegraphics[width=\textwidth]{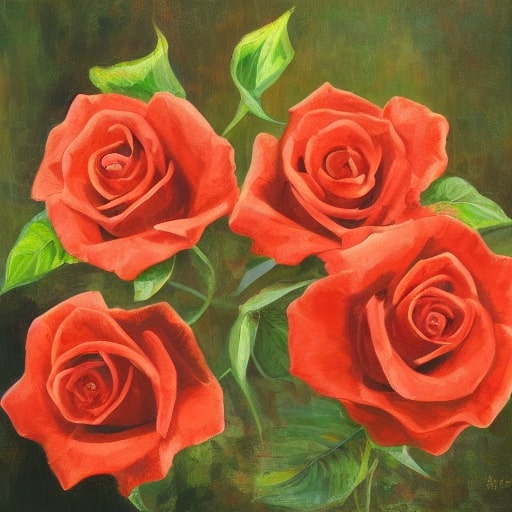}
    \end{minipage}
    \begin{minipage}{0.09\textwidth}
        \centering
  \includegraphics[width=\textwidth]{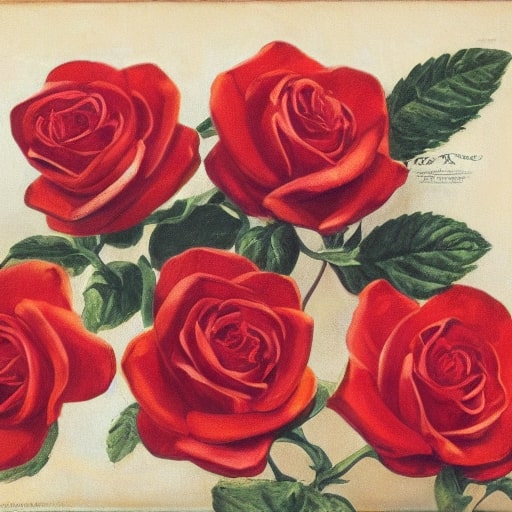}
    \end{minipage}
    \caption{DPOK}
\end{figure}

\begin{figure}[h!]
\vspace{-1em}
    \centering
    \begin{minipage}{0.09\textwidth}
        \centering
        \includegraphics[width=\textwidth]{figures/text-to-image-jpg/four_roses/gfn/image_0.jpg} %
    \end{minipage}
    \begin{minipage}{0.09\textwidth}
        \centering
         \includegraphics[width=\textwidth]{figures/text-to-image-jpg/four_roses/gfn/image_1.jpg} %
    \end{minipage}
    \begin{minipage}{0.09\textwidth}
        \centering
         \includegraphics[width=\textwidth]{figures/text-to-image-jpg/four_roses/gfn/image_2.jpg} %
    \end{minipage}
    \begin{minipage}{0.09\textwidth}
        \centering
        \includegraphics[width=\textwidth]{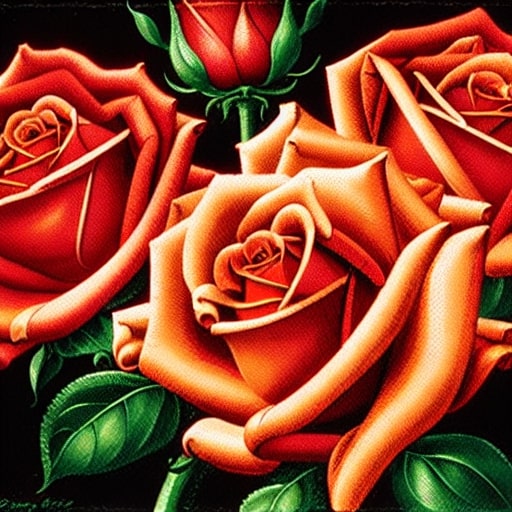} 
    \end{minipage}
    \begin{minipage}{0.09\textwidth}
        \centering
         \includegraphics[width=\textwidth]{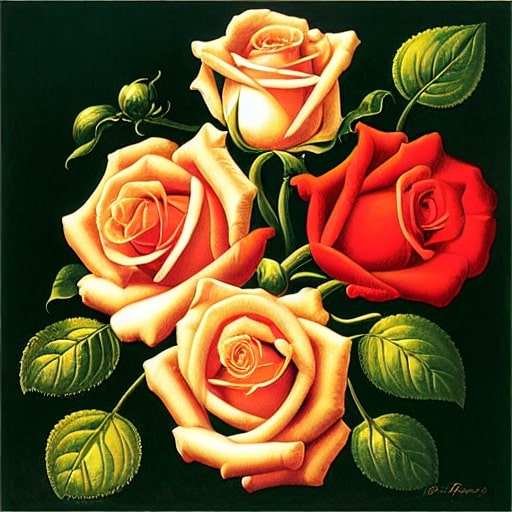}
    \end{minipage}
    \begin{minipage}{0.09\textwidth}
        \centering
       \includegraphics[width=\textwidth]{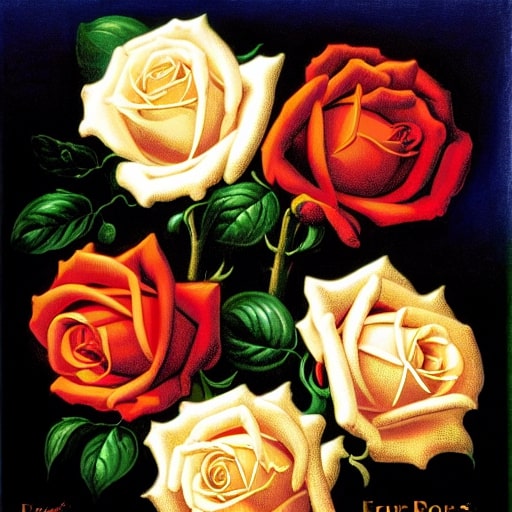}
    \end{minipage}
    \begin{minipage}{0.09\textwidth}
        \centering
    \includegraphics[width=\textwidth]{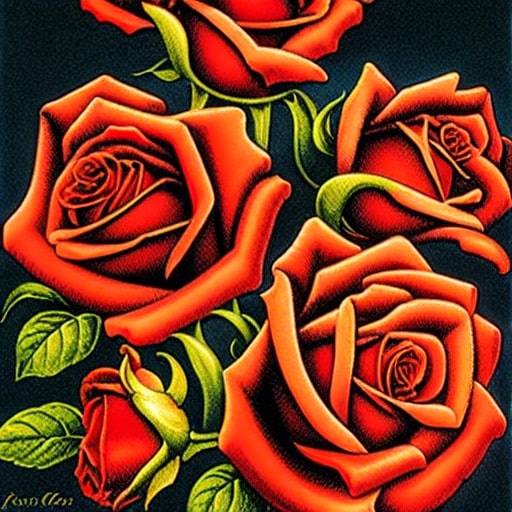}
    \end{minipage}
    \begin{minipage}{0.09\textwidth}
        \centering
 \includegraphics[width=\textwidth]{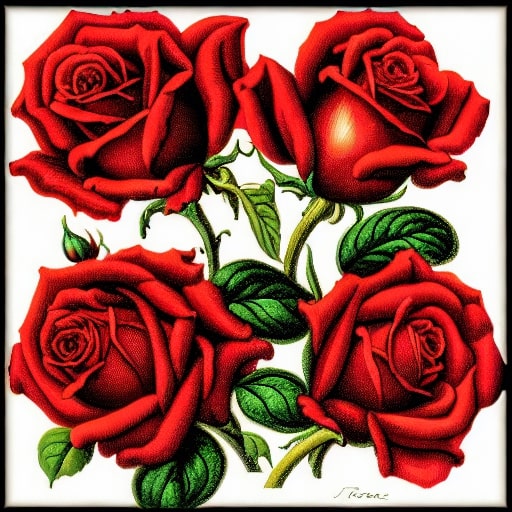}
    \end{minipage}
    \begin{minipage}{0.09\textwidth}
        \centering
 \includegraphics[width=\textwidth]{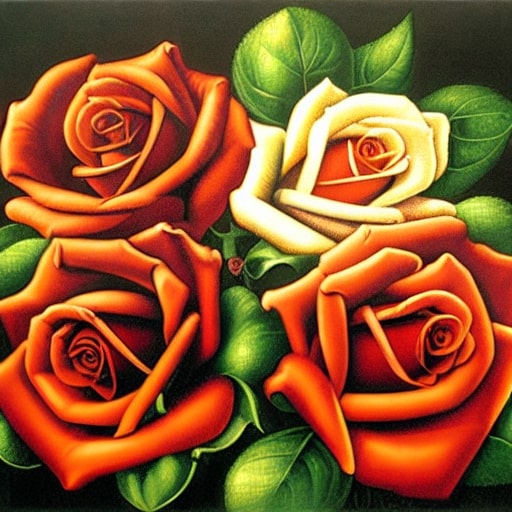}
    \end{minipage}
    \begin{minipage}{0.09\textwidth}
        \centering
  \includegraphics[width=\textwidth]{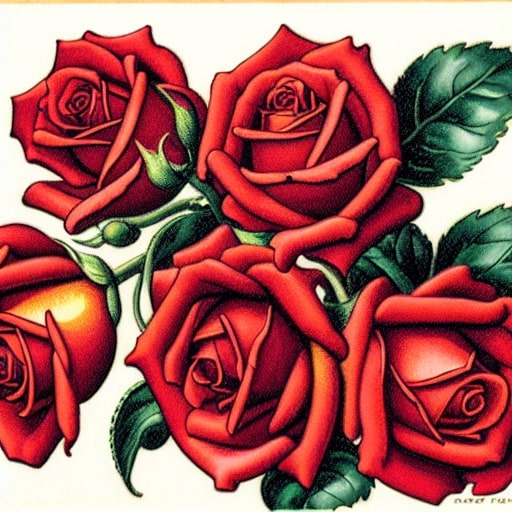}
    \end{minipage}
    \caption{RTB}
\end{figure}

\clearpage

\subsubsection{A cat and a dog}
\begin{figure}[h!]
\vspace{-1em}
    \centering
    \begin{minipage}{0.09\textwidth}
        \centering
        \includegraphics[width=\textwidth]{figures/text-to-image-jpg/cat_dog/prior/image_0.jpg} %
    \end{minipage}
    \begin{minipage}{0.09\textwidth}
        \centering
         \includegraphics[width=\textwidth]{figures/text-to-image-jpg/cat_dog/prior/image_1.jpg} %
    \end{minipage}
    \begin{minipage}{0.09\textwidth}
        \centering
         \includegraphics[width=\textwidth]{figures/text-to-image-jpg/cat_dog/prior/image_2.jpg} %
    \end{minipage}
    \begin{minipage}{0.09\textwidth}
        \centering
        \includegraphics[width=\textwidth]{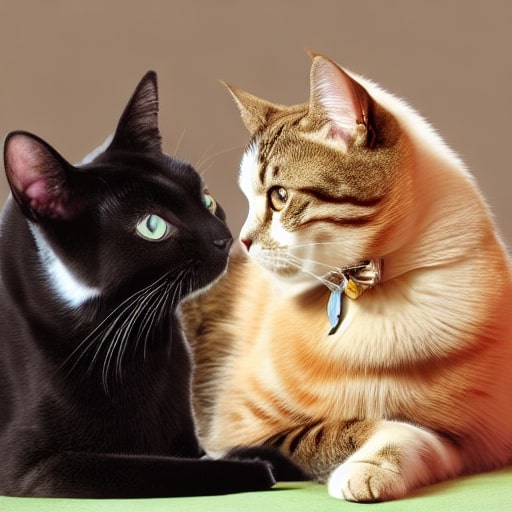} 
    \end{minipage}
    \begin{minipage}{0.09\textwidth}
        \centering
         \includegraphics[width=\textwidth]{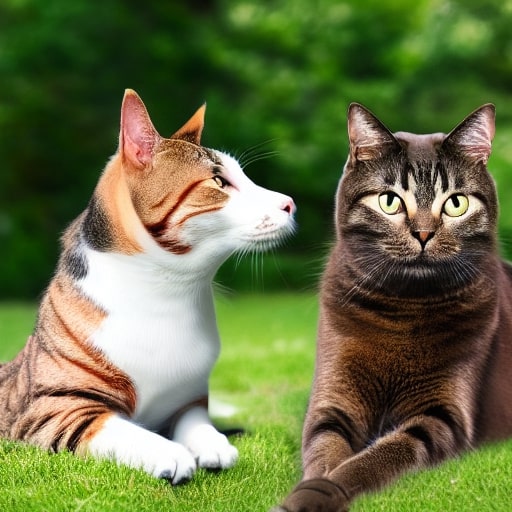}
    \end{minipage}
    \begin{minipage}{0.09\textwidth}
        \centering
       \includegraphics[width=\textwidth]{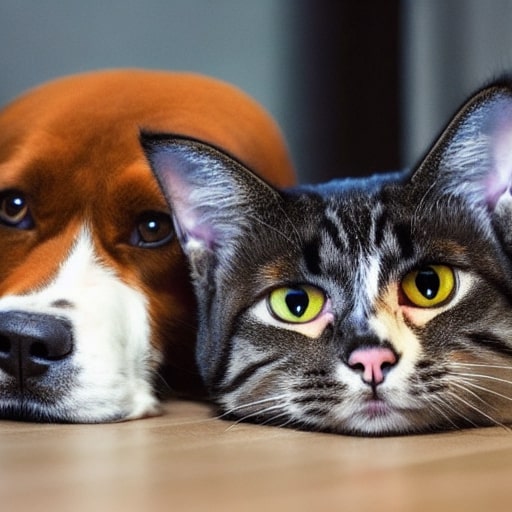}
    \end{minipage}
    \begin{minipage}{0.09\textwidth}
        \centering
    \includegraphics[width=\textwidth]{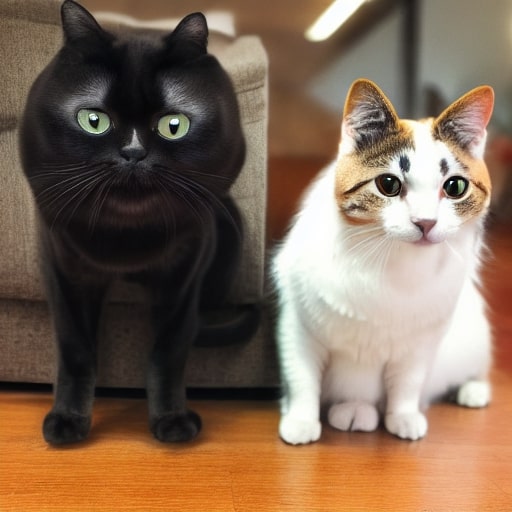}
    \end{minipage}
    \begin{minipage}{0.09\textwidth}
        \centering
 \includegraphics[width=\textwidth]{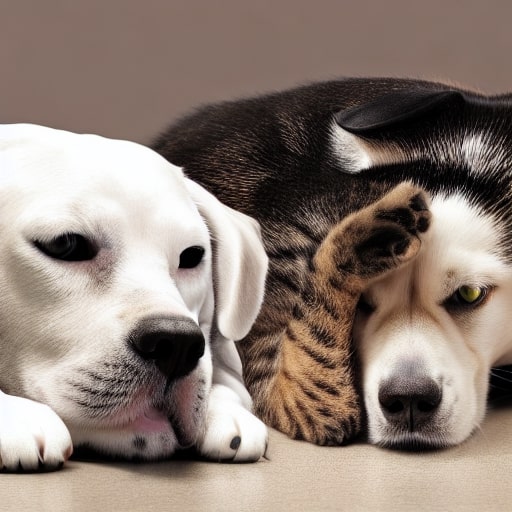}
    \end{minipage}
    \begin{minipage}{0.09\textwidth}
        \centering
 \includegraphics[width=\textwidth]{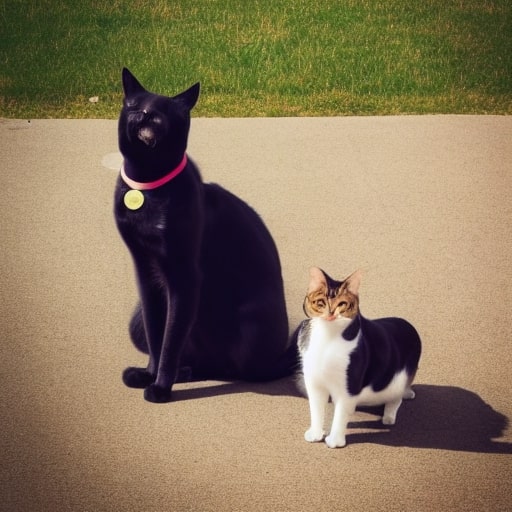}
    \end{minipage}
    \begin{minipage}{0.09\textwidth}
        \centering
  \includegraphics[width=\textwidth]{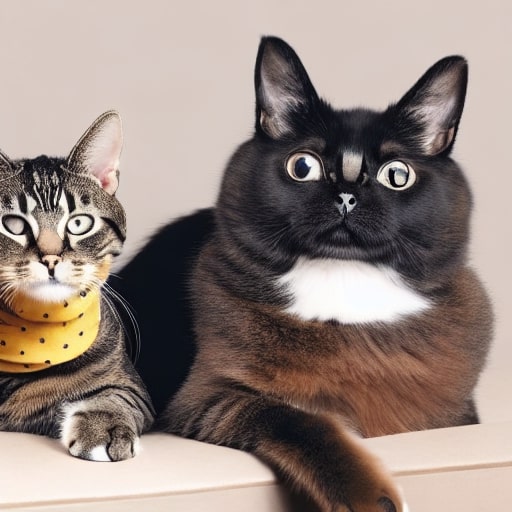}
    \end{minipage}
    \caption{Prior}
\end{figure}

 \begin{figure}[h!]
 \vspace{-1em}
     \centering
     \begin{minipage}{0.09\textwidth}
         \centering
         \includegraphics[width=\textwidth]{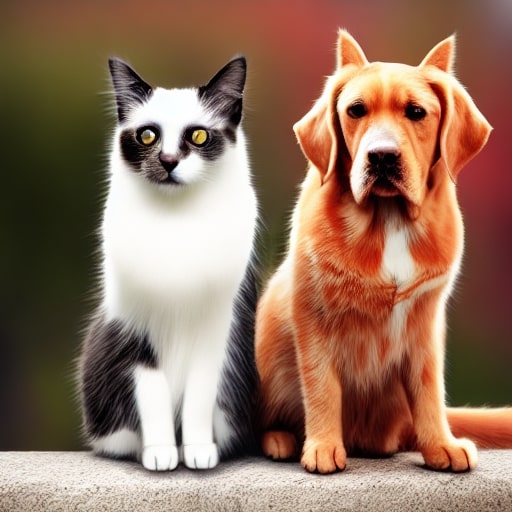} %
     \end{minipage}
     \begin{minipage}{0.09\textwidth}
         \centering
          \includegraphics[width=\textwidth]{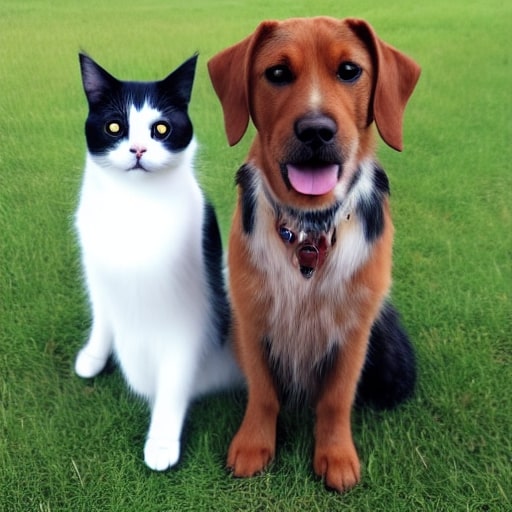} %
     \end{minipage}
     \begin{minipage}{0.09\textwidth}
         \centering
          \includegraphics[width=\textwidth]{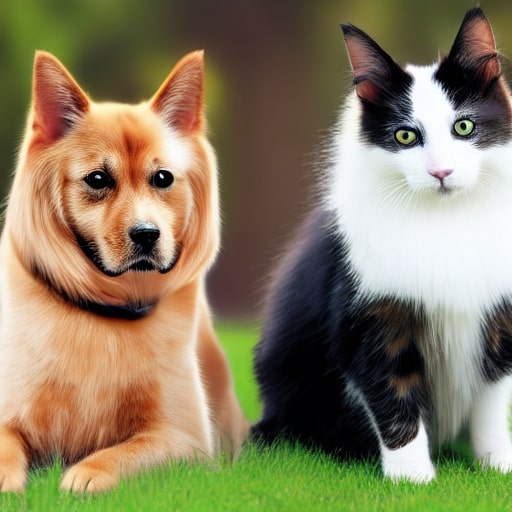} %
     \end{minipage}
     \begin{minipage}{0.09\textwidth}
         \centering
         \includegraphics[width=\textwidth]{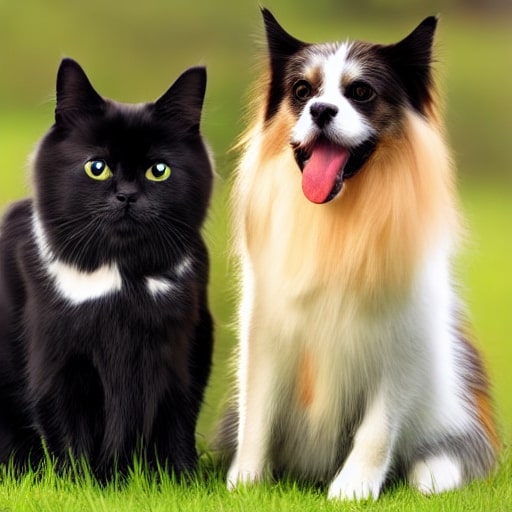} 
     \end{minipage}
     \begin{minipage}{0.09\textwidth}
         \centering
          \includegraphics[width=\textwidth]{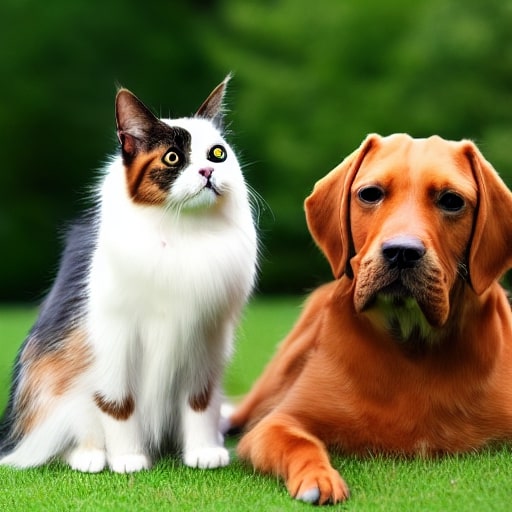}
     \end{minipage}
     \begin{minipage}{0.09\textwidth}
         \centering
        \includegraphics[width=\textwidth]{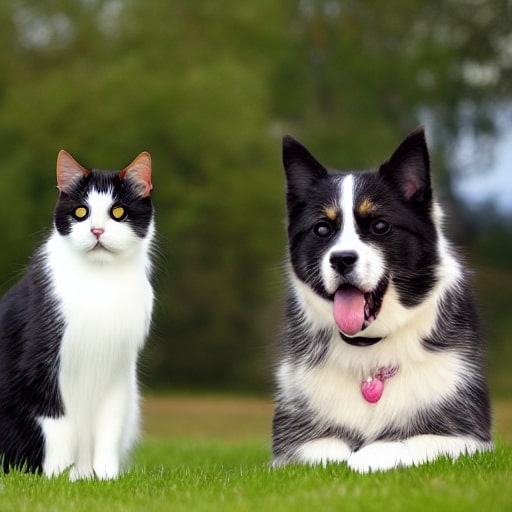}
     \end{minipage}
     \begin{minipage}{0.09\textwidth}
         \centering
     \includegraphics[width=\textwidth]{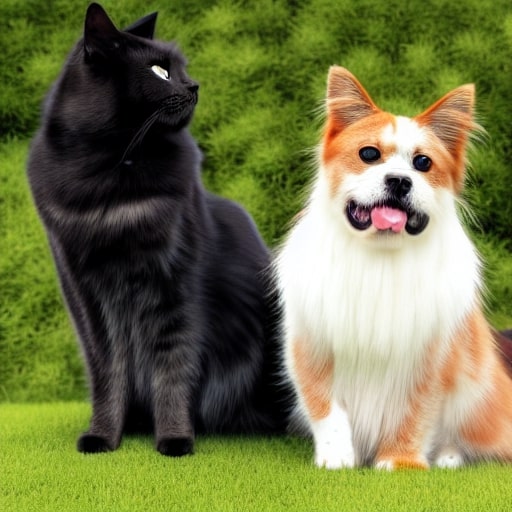}
     \end{minipage}
     \begin{minipage}{0.09\textwidth}
         \centering
  \includegraphics[width=\textwidth]{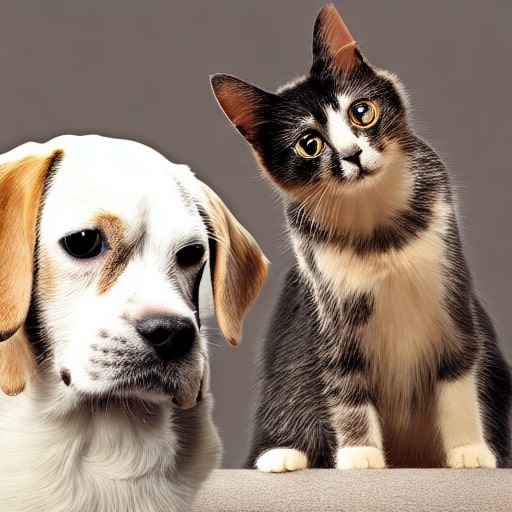}
     \end{minipage}
     \begin{minipage}{0.09\textwidth}
         \centering
  \includegraphics[width=\textwidth]{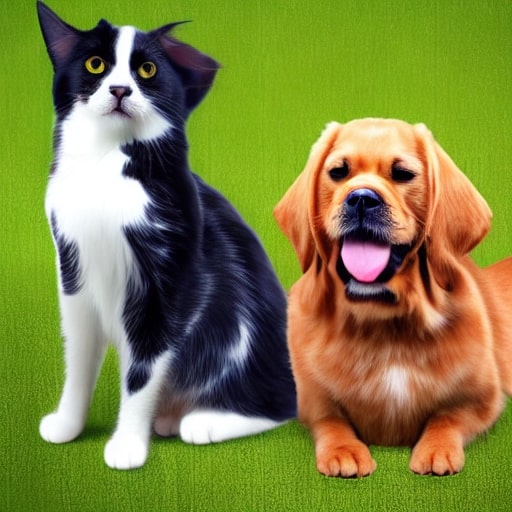}
     \end{minipage}
     \begin{minipage}{0.09\textwidth}
         \centering
   \includegraphics[width=\textwidth]{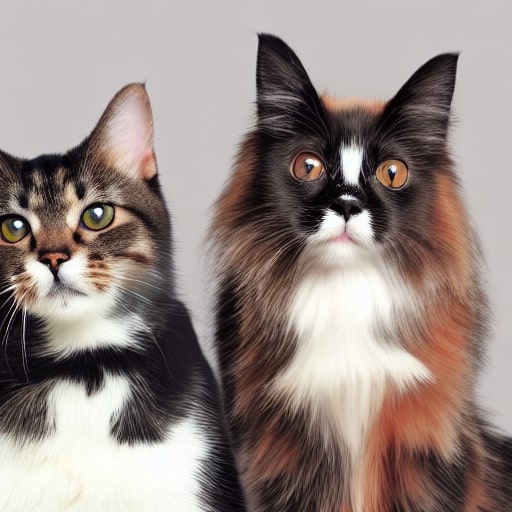}
     \end{minipage}
     \caption{DDPO}
 \end{figure}

\begin{figure}[h!]
\vspace{-1em}
    \centering
    \begin{minipage}{0.09\textwidth}
        \centering
        \includegraphics[width=\textwidth]{figures/text-to-image-jpg/cat_dog/dpok/image_0.jpg} %
    \end{minipage}
    \begin{minipage}{0.09\textwidth}
        \centering
         \includegraphics[width=\textwidth]{figures/text-to-image-jpg/cat_dog/dpok/image_1.jpg} %
    \end{minipage}
    \begin{minipage}{0.09\textwidth}
        \centering
         \includegraphics[width=\textwidth]{figures/text-to-image-jpg/cat_dog/dpok/image_2.jpg} %
    \end{minipage}
    \begin{minipage}{0.09\textwidth}
        \centering
        \includegraphics[width=\textwidth]{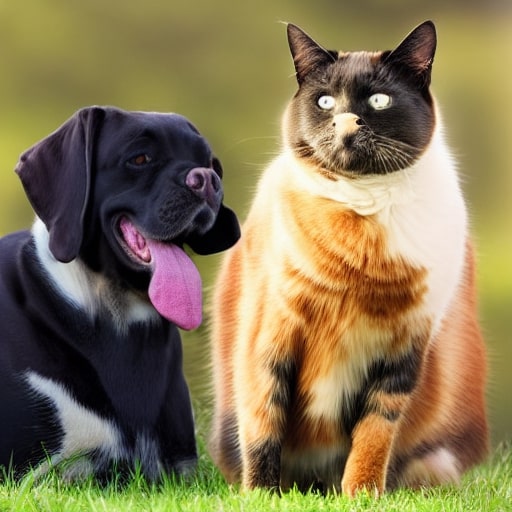} 
    \end{minipage}
    \begin{minipage}{0.09\textwidth}
        \centering
         \includegraphics[width=\textwidth]{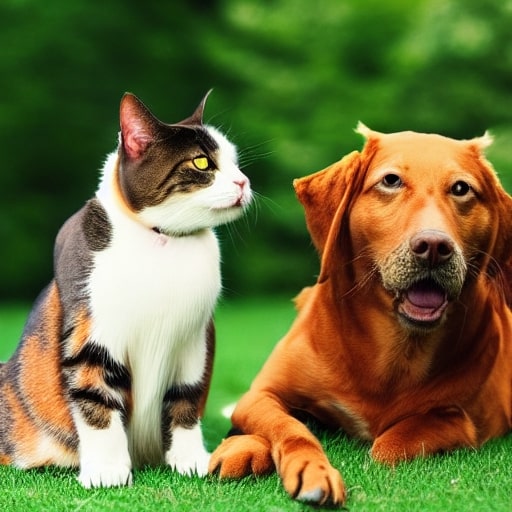}
    \end{minipage}
    \begin{minipage}{0.09\textwidth}
        \centering
       \includegraphics[width=\textwidth]{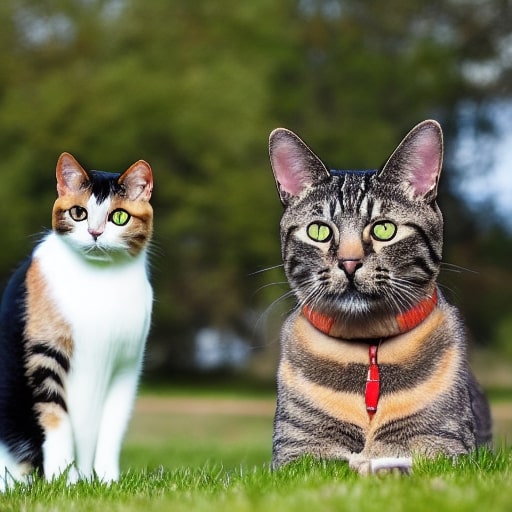}
    \end{minipage}
    \begin{minipage}{0.09\textwidth}
        \centering
    \includegraphics[width=\textwidth]{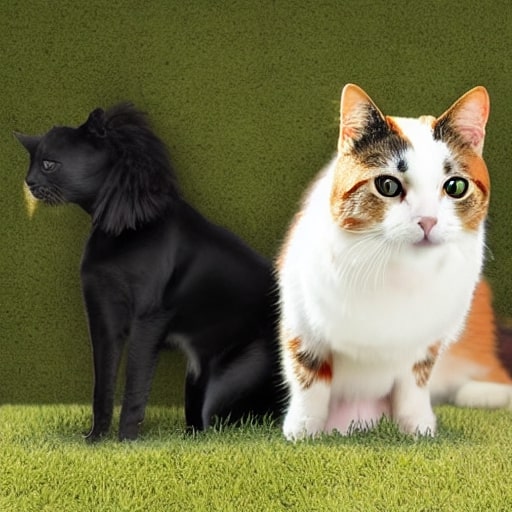}
    \end{minipage}
    \begin{minipage}{0.09\textwidth}
        \centering
 \includegraphics[width=\textwidth]{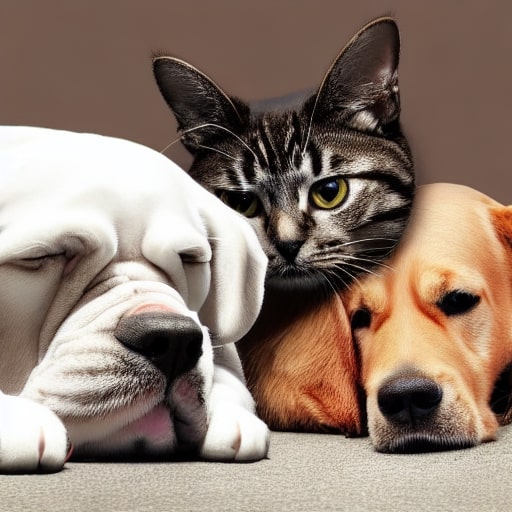}
    \end{minipage}
    \begin{minipage}{0.09\textwidth}
        \centering
 \includegraphics[width=\textwidth]{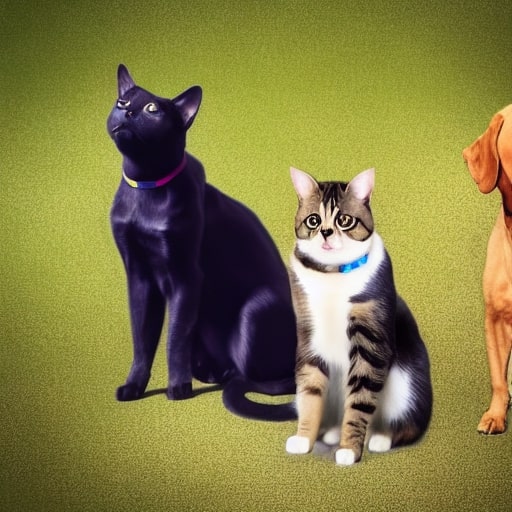}
    \end{minipage}
    \begin{minipage}{0.09\textwidth}
        \centering
  \includegraphics[width=\textwidth]{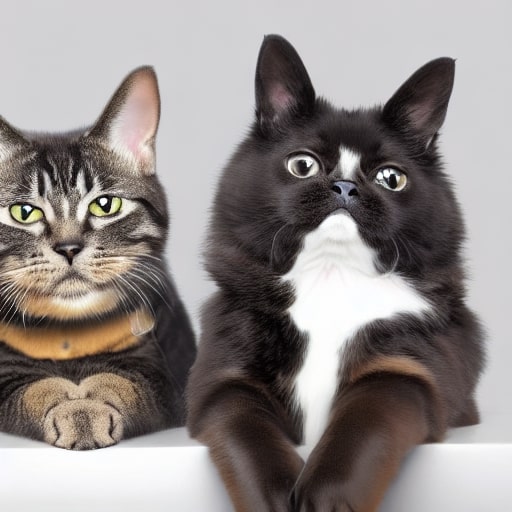}
    \end{minipage}
    \caption{DPOK}
\end{figure}

\begin{figure}[h!]
\vspace{-1em}
    \centering
    \begin{minipage}{0.09\textwidth}
        \centering
        \includegraphics[width=\textwidth]{figures/text-to-image-jpg/cat_dog/gfn/image_0.jpg} %
    \end{minipage}
    \begin{minipage}{0.09\textwidth}
        \centering
         \includegraphics[width=\textwidth]{figures/text-to-image-jpg/cat_dog/gfn/image_1.jpg} %
    \end{minipage}
    \begin{minipage}{0.09\textwidth}
        \centering
         \includegraphics[width=\textwidth]{figures/text-to-image-jpg/cat_dog/gfn/image_2.jpg} %
    \end{minipage}
    \begin{minipage}{0.09\textwidth}
        \centering
        \includegraphics[width=\textwidth]{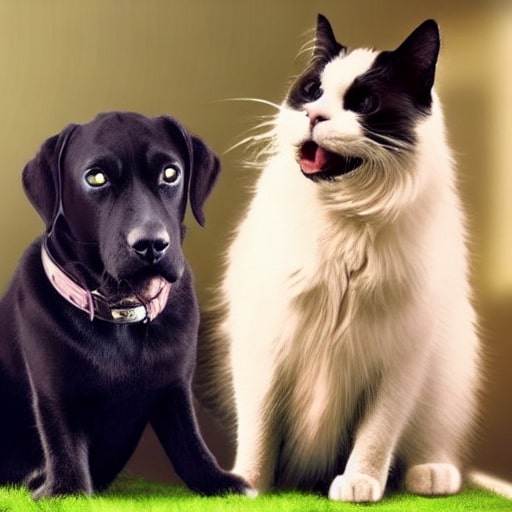} 
    \end{minipage}
    \begin{minipage}{0.09\textwidth}
        \centering
         \includegraphics[width=\textwidth]{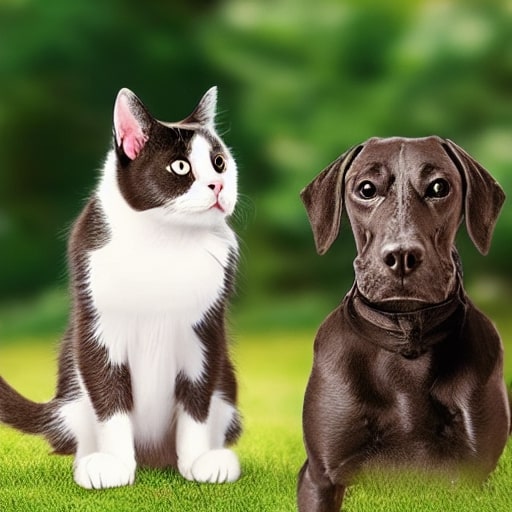}
    \end{minipage}
    \begin{minipage}{0.09\textwidth}
        \centering
       \includegraphics[width=\textwidth]{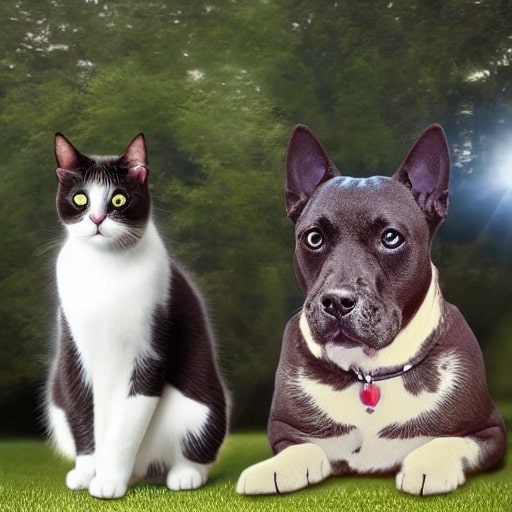}
    \end{minipage}
    \begin{minipage}{0.09\textwidth}
        \centering
    \includegraphics[width=\textwidth]{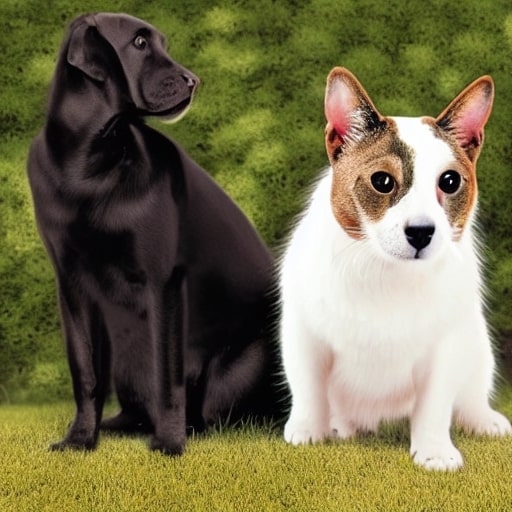}
    \end{minipage}
    \begin{minipage}{0.09\textwidth}
        \centering
 \includegraphics[width=\textwidth]{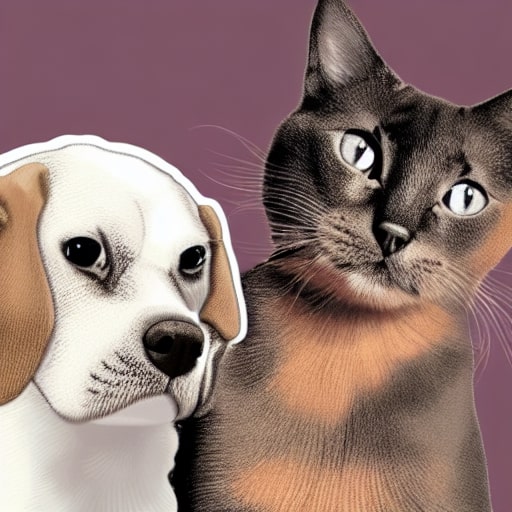}
    \end{minipage}
    \begin{minipage}{0.09\textwidth}
        \centering
 \includegraphics[width=\textwidth]{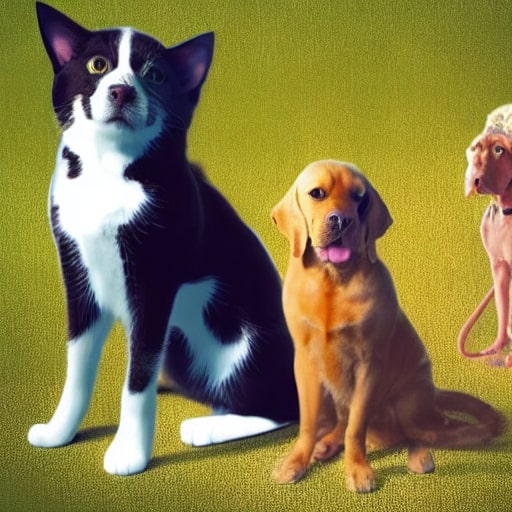}
    \end{minipage}
    \begin{minipage}{0.09\textwidth}
        \centering
  \includegraphics[width=\textwidth]{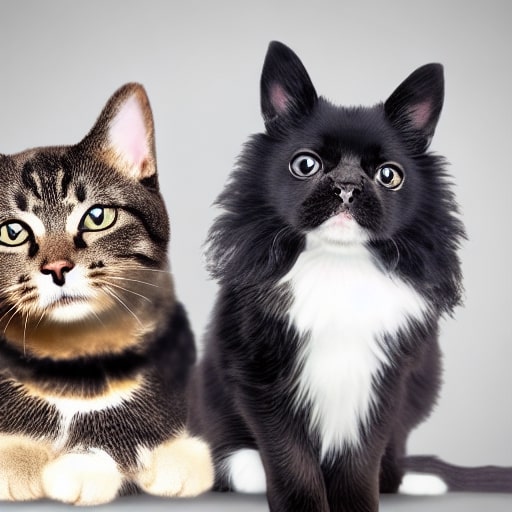}
    \end{minipage}
    \caption{RTB}
\end{figure}

\subsubsection{A half - masked rugged laboratory engineer man with cybernetic enhancements as seen from a distance, scifi character portrait by greg rutkowski, esuthio, craig mullins.}

\begin{figure}[h!]
    \vspace*{-1em}
    \centering
    \begin{minipage}{0.09\textwidth}
        \centering
        \includegraphics[width=\textwidth]{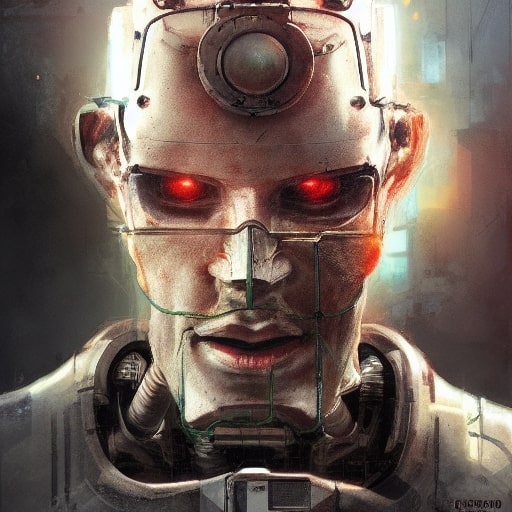} %
    \end{minipage}
    \begin{minipage}{0.09\textwidth}
        \centering
         \includegraphics[width=\textwidth]{figures/text-to-image-jpg/cybernetic_man/prior/image1.jpg} %
    \end{minipage}
    \begin{minipage}{0.09\textwidth}
        \centering
         \includegraphics[width=\textwidth]{figures/text-to-image-jpg/cybernetic_man/prior/image2.jpg} %
    \end{minipage}
    \begin{minipage}{0.09\textwidth}
        \centering
        \includegraphics[width=\textwidth]{figures/text-to-image-jpg/cybernetic_man/prior/image3.jpg} 
    \end{minipage}
    \begin{minipage}{0.09\textwidth}
        \centering
         \includegraphics[width=\textwidth]{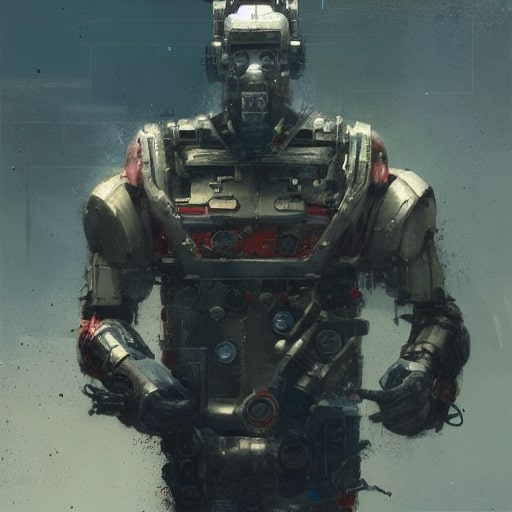}
    \end{minipage}
    \begin{minipage}{0.09\textwidth}
        \centering
       \includegraphics[width=\textwidth]{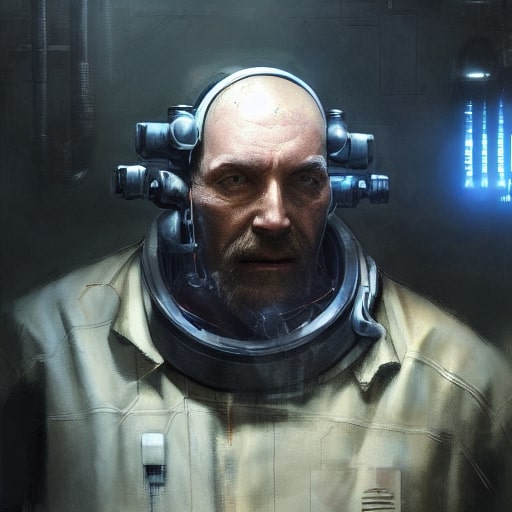}
    \end{minipage}
    \begin{minipage}{0.09\textwidth}
        \centering
    \includegraphics[width=\textwidth]{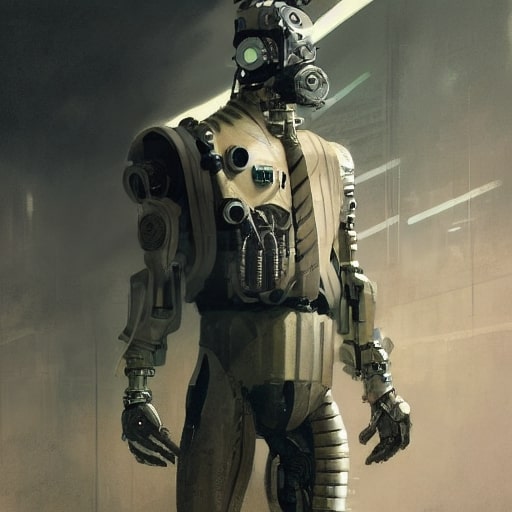}
    \end{minipage}
    \begin{minipage}{0.09\textwidth}
        \centering
 \includegraphics[width=\textwidth]{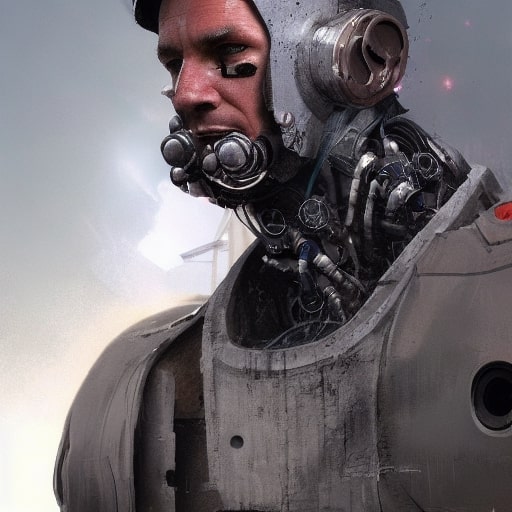}
    \end{minipage}
    \begin{minipage}{0.09\textwidth}
        \centering
 \includegraphics[width=\textwidth]{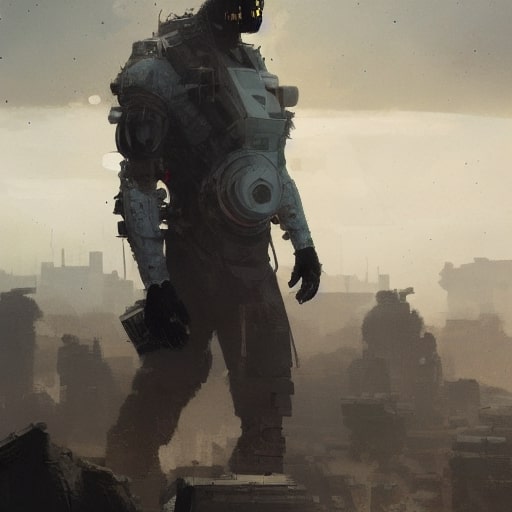}
    \end{minipage}
    \begin{minipage}{0.09\textwidth}
        \centering
  \includegraphics[width=\textwidth]{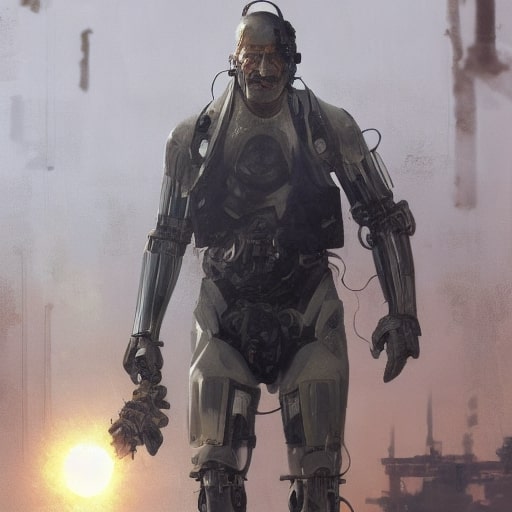}
    \end{minipage}
    \caption{Prior}
\end{figure}

\begin{figure}[h!]
    \vspace*{-1em}
    \centering
    \begin{minipage}{0.09\textwidth}
        \centering
        \includegraphics[width=\textwidth]{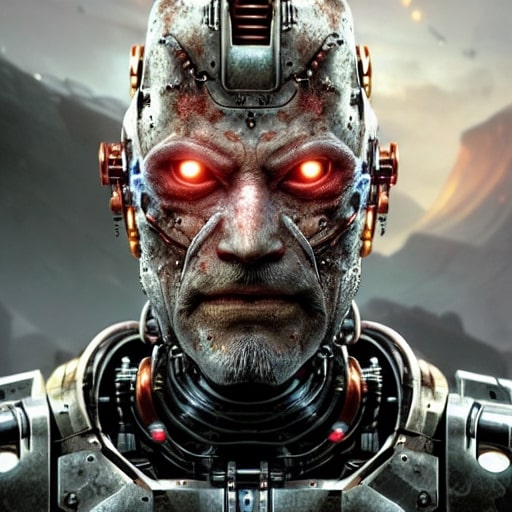} %
    \end{minipage}
    \begin{minipage}{0.09\textwidth}
        \centering
         \includegraphics[width=\textwidth]{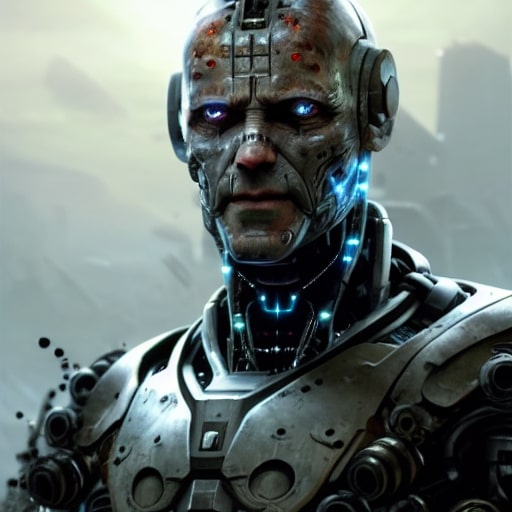} %
    \end{minipage}
    \begin{minipage}{0.09\textwidth}
        \centering
         \includegraphics[width=\textwidth]{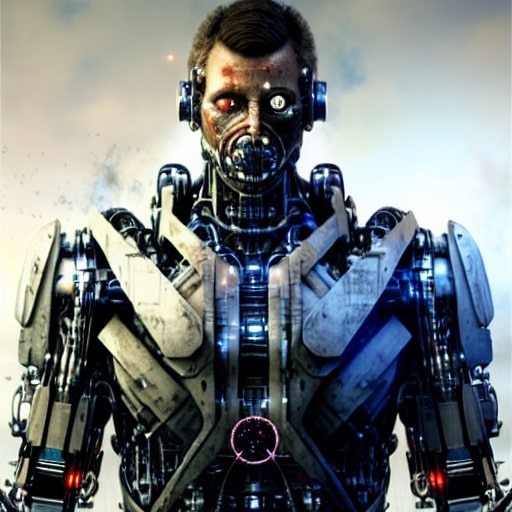} %
    \end{minipage}
    \begin{minipage}{0.09\textwidth}
        \centering
        \includegraphics[width=\textwidth]{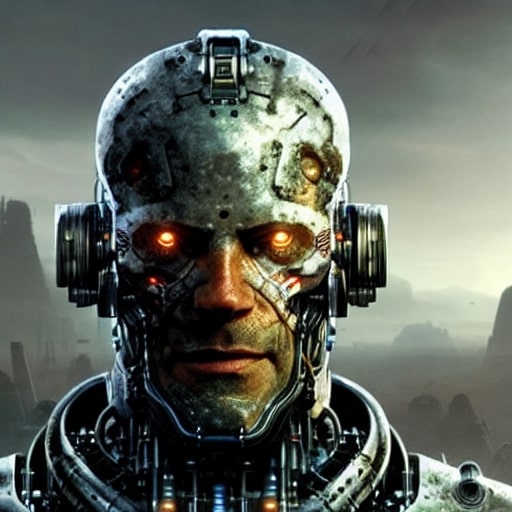} 
    \end{minipage}
    \begin{minipage}{0.09\textwidth}
        \centering
         \includegraphics[width=\textwidth]{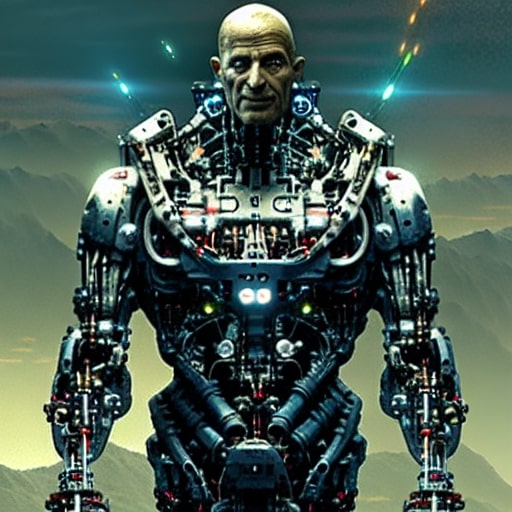}
    \end{minipage}
    \begin{minipage}{0.09\textwidth}
        \centering
       \includegraphics[width=\textwidth]{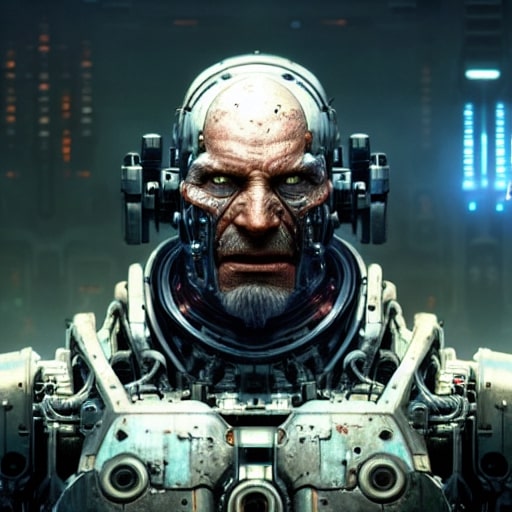}
    \end{minipage}
    \begin{minipage}{0.09\textwidth}
        \centering
    \includegraphics[width=\textwidth]{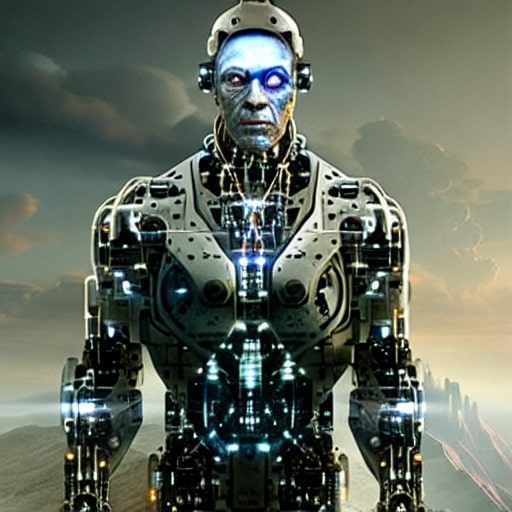}
    \end{minipage}
    \begin{minipage}{0.09\textwidth}
        \centering
 \includegraphics[width=\textwidth]{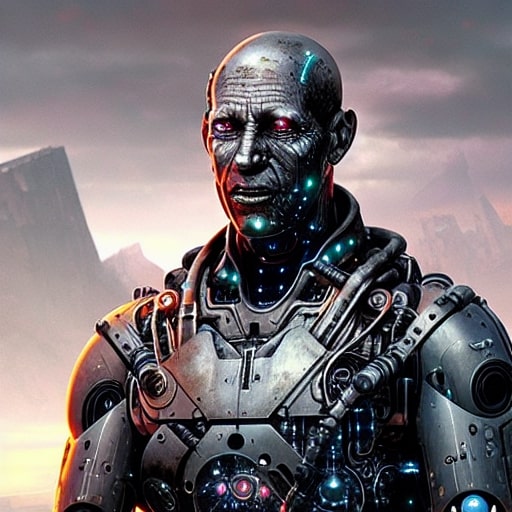}
    \end{minipage}
    \begin{minipage}{0.09\textwidth}
        \centering
 \includegraphics[width=\textwidth]{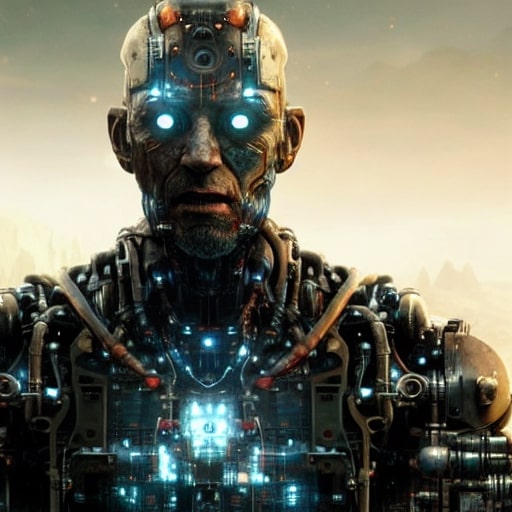}
    \end{minipage}
    \begin{minipage}{0.09\textwidth}
        \centering
  \includegraphics[width=\textwidth]{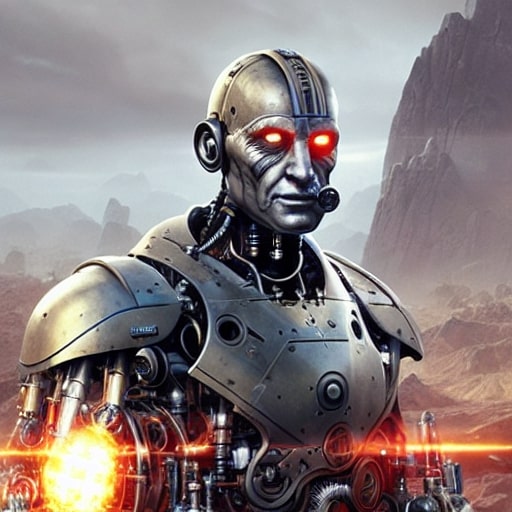}
    \end{minipage}
    \caption{DDPO}
\end{figure}

\begin{figure}[h!]
    \vspace*{-1em}
    \centering
    \begin{minipage}{0.09\textwidth}
        \centering
        \includegraphics[width=\textwidth]{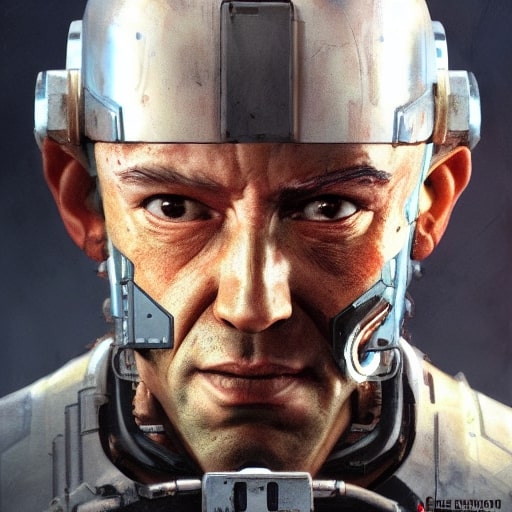} %
    \end{minipage}
    \begin{minipage}{0.09\textwidth}
        \centering
         \includegraphics[width=\textwidth]{figures/text-to-image-jpg/cybernetic_man/dpok/image1.jpg} %
    \end{minipage}
    \begin{minipage}{0.09\textwidth}
        \centering
         \includegraphics[width=\textwidth]{figures/text-to-image-jpg/cybernetic_man/dpok/image2.jpg} %
    \end{minipage}
    \begin{minipage}{0.09\textwidth}
        \centering
        \includegraphics[width=\textwidth]{figures/text-to-image-jpg/cybernetic_man/dpok/image3.jpg} 
    \end{minipage}
    \begin{minipage}{0.09\textwidth}
        \centering
         \includegraphics[width=\textwidth]{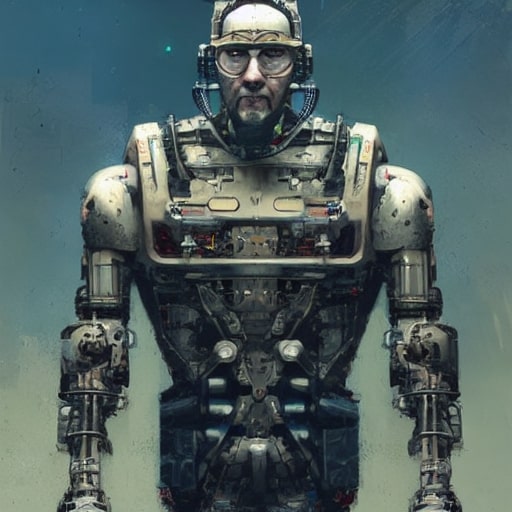}
    \end{minipage}
    \begin{minipage}{0.09\textwidth}
        \centering
       \includegraphics[width=\textwidth]{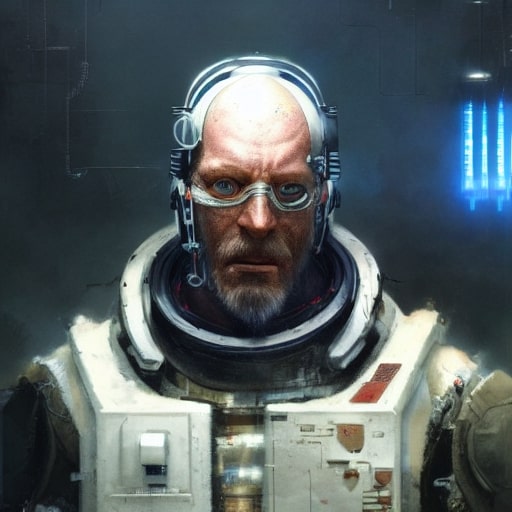}
    \end{minipage}
    \begin{minipage}{0.09\textwidth}
        \centering
    \includegraphics[width=\textwidth]{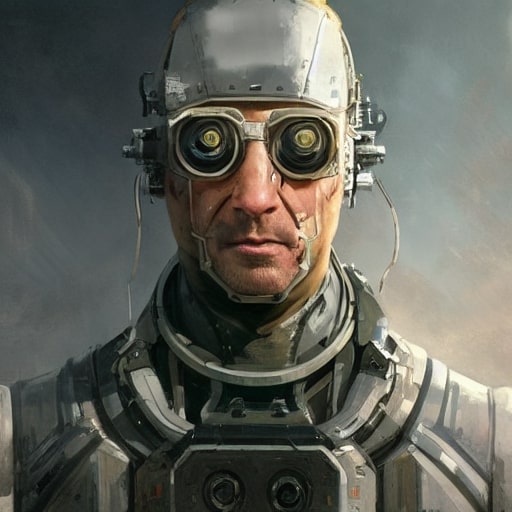}
    \end{minipage}
    \begin{minipage}{0.09\textwidth}
        \centering
 \includegraphics[width=\textwidth]{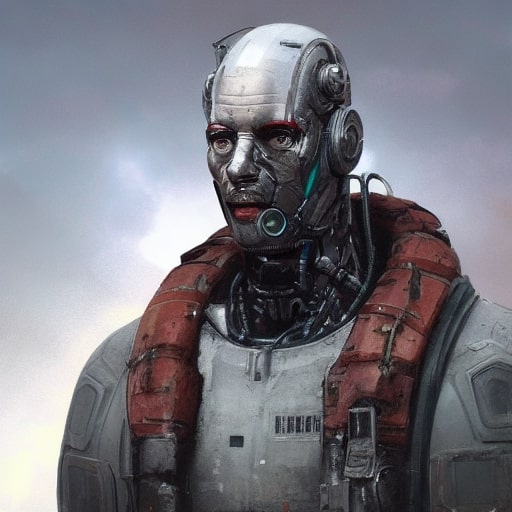}
    \end{minipage}
    \begin{minipage}{0.09\textwidth}
        \centering
 \includegraphics[width=\textwidth]{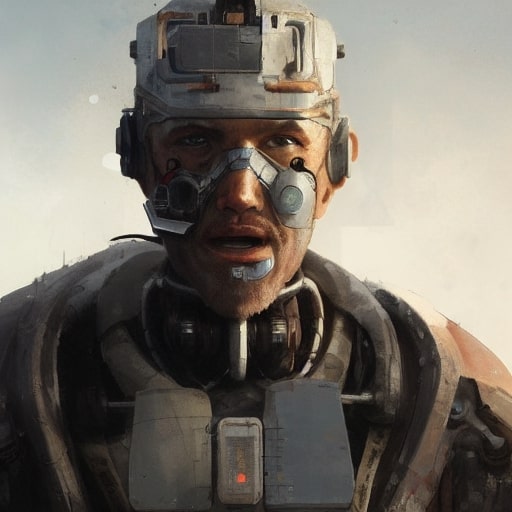}
    \end{minipage}
    \begin{minipage}{0.09\textwidth}
        \centering
  \includegraphics[width=\textwidth]{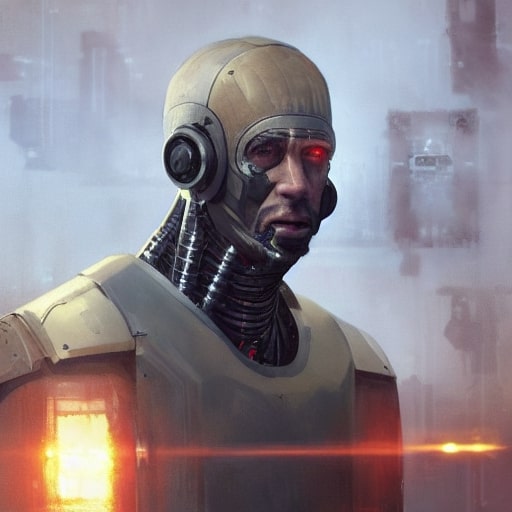}
    \end{minipage}
    \caption{DPOK}
\end{figure}

\begin{figure}[h!]
    \vspace*{-1em}
    \centering
    \begin{minipage}{0.09\textwidth}
        \centering
        \includegraphics[width=\textwidth]{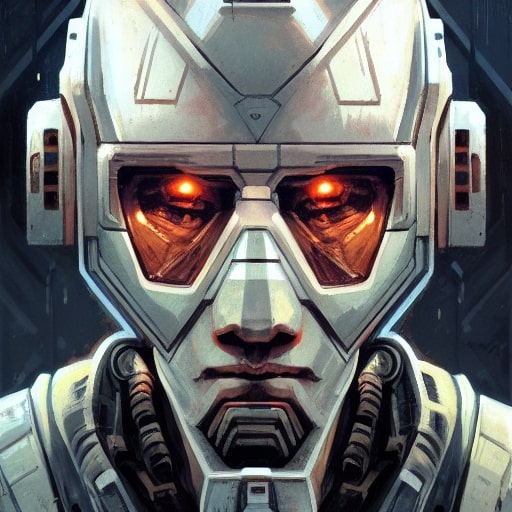} %
    \end{minipage}
    \begin{minipage}{0.09\textwidth}
        \centering
         \includegraphics[width=\textwidth]{figures/text-to-image-jpg/cybernetic_man/gfn/image1.jpg} %
    \end{minipage}
    \begin{minipage}{0.09\textwidth}
        \centering
         \includegraphics[width=\textwidth]{figures/text-to-image-jpg/cybernetic_man/gfn/image2.jpg} %
    \end{minipage}
    \begin{minipage}{0.09\textwidth}
        \centering
        \includegraphics[width=\textwidth]{figures/text-to-image-jpg/cybernetic_man/gfn/image3.jpg} 
    \end{minipage}
    \begin{minipage}{0.09\textwidth}
        \centering
         \includegraphics[width=\textwidth]{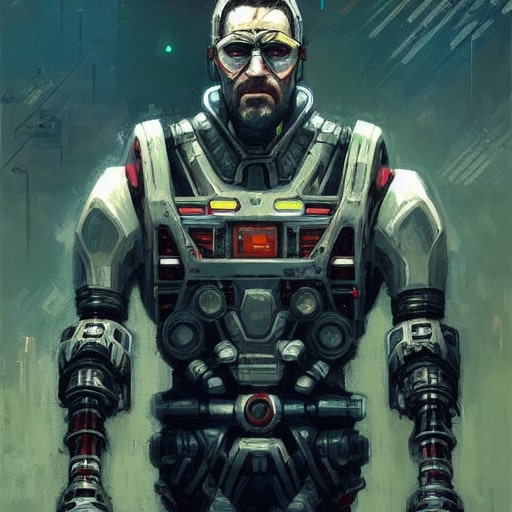}
    \end{minipage}
    \begin{minipage}{0.09\textwidth}
        \centering
       \includegraphics[width=\textwidth]{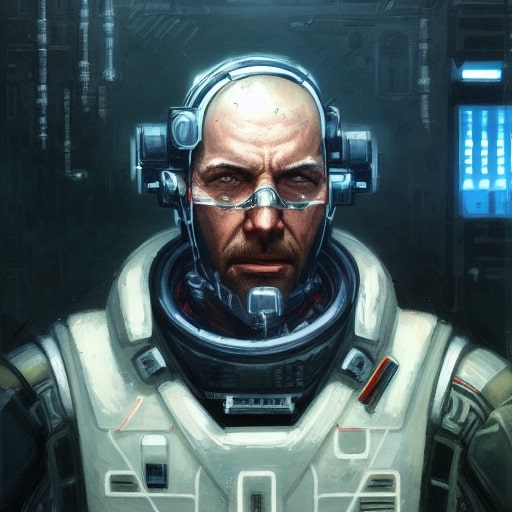}
    \end{minipage}
    \begin{minipage}{0.09\textwidth}
        \centering
    \includegraphics[width=\textwidth]{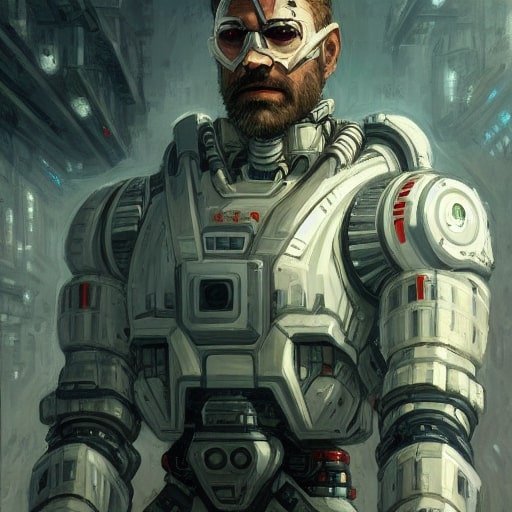}
    \end{minipage}
    \begin{minipage}{0.09\textwidth}
        \centering
 \includegraphics[width=\textwidth]{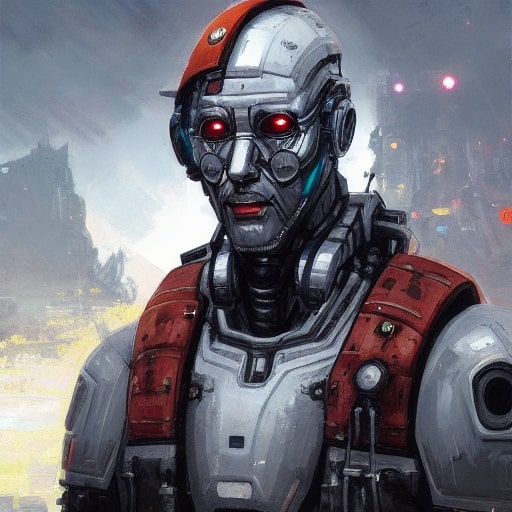}
    \end{minipage}
    \begin{minipage}{0.09\textwidth}
        \centering
 \includegraphics[width=\textwidth]{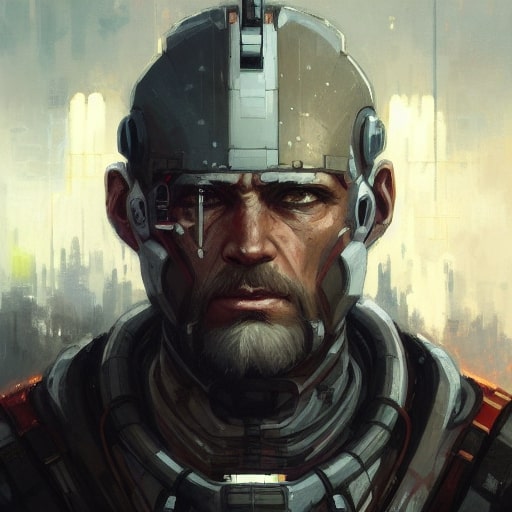}
    \end{minipage}
    \begin{minipage}{0.09\textwidth}
        \centering
  \includegraphics[width=\textwidth]{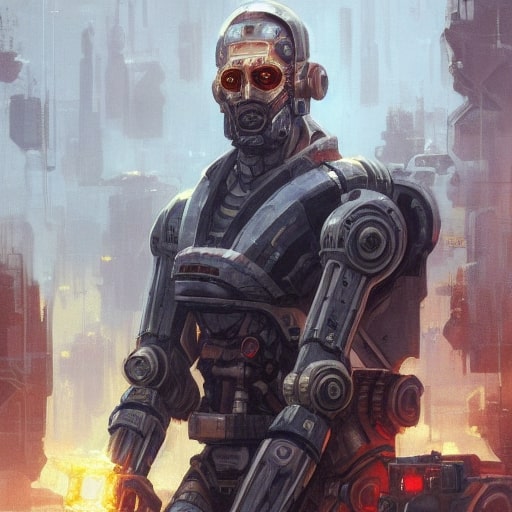}
    \end{minipage}
    \caption{RTB}
\end{figure}

\section{Compute resources}
\label{app:compute}
For classifier guidance experiments \autoref{sec:experiments:cls_guidance} we use train on a single NVIDIA V100 GPU. For text-conditional image generation \autoref{sec:experiments:text2image}, text infilling \autoref{app:infilling} and offline \autoref{app:offline_rl}, we use a single NVIDIA A100 large GPU. The total estimated compute time for all our experiments is 3000 hours.

\clearpage

\newpage
\section*{NeurIPS Paper Checklist}

\begin{enumerate}

\item {\bf Claims}
    \item[] Question: Do the main claims made in the abstract and introduction accurately reflect the paper's contributions and scope?
    \item[] Answer: \answerYes{}%
    \item[] Justification: We discuss our theoretical claims about the RTB ojective in \autoref{sec:methods}, and report experimental results in \autoref{sec:experiments}.%
    \item[] Guidelines:
    \begin{itemize}
        \item The answer NA means that the abstract and introduction do not include the claims made in the paper.
        \item The abstract and/or introduction should clearly state the claims made, including the contributions made in the paper and important assumptions and limitations. A No or NA answer to this question will not be perceived well by the reviewers. 
        \item The claims made should match theoretical and experimental results, and reflect how much the results can be expected to generalize to other settings. 
        \item It is fine to include aspirational goals as motivation as long as it is clear that these goals are not attained by the paper. 
    \end{itemize}

\item {\bf Limitations}
    \item[] Question: Does the paper discuss the limitations of the work performed by the authors?
    \item[] Answer: \answerYes{} %
    \item[] Justification: We discuss the limitations of our proposed method in \autoref{sec:conclusion}. 
    \item[] Guidelines:

\item {\bf Theory Assumptions and Proofs}
    \item[] Question: For each theoretical result, does the paper provide the full set of assumptions and a complete (and correct) proof?
    \item[] Answer: \answerYes{} %
    \item[] Justification: See \autoref{sec:proofs}.
    \item[] Guidelines:
    \begin{itemize}
        \item The answer NA means that the paper does not include theoretical results. 
        \item All the theorems, formulas, and proofs in the paper should be numbered and cross-referenced.
        \item All assumptions should be clearly stated or referenced in the statement of any theorems.
        \item The proofs can either appear in the main paper or the supplemental material, but if they appear in the supplemental material, the authors are encouraged to provide a short proof sketch to provide intuition. 
        \item Inversely, any informal proof provided in the core of the paper should be complemented by formal proofs provided in appendix or supplemental material.
        \item Theorems and Lemmas that the proof relies upon should be properly referenced. 
    \end{itemize}

    \item {\bf Experimental Result Reproducibility}
    \item[] Question: Does the paper fully disclose all the information needed to reproduce the main experimental results of the paper to the extent that it affects the main claims and/or conclusions of the paper (regardless of whether the code and data are provided or not)?
    \item[] Answer: \answerYes{} %
    \item[] Justification: We have provided code to reproduce our experiments in \autoref{app:code}, and described training details in \autoref{app:2d_gmm}, \autoref{app:cls_rtb}, \autoref{app:infilling}, \autoref{app:offline_rl} and \autoref{app:text2image}.
    \item[] Guidelines:
    \begin{itemize}
        \item The answer NA means that the paper does not include experiments.
        \item If the paper includes experiments, a No answer to this question will not be perceived well by the reviewers: Making the paper reproducible is important, regardless of whether the code and data are provided or not.
        \item If the contribution is a dataset and/or model, the authors should describe the steps taken to make their results reproducible or verifiable. 
        \item Depending on the contribution, reproducibility can be accomplished in various ways. For example, if the contribution is a novel architecture, describing the architecture fully might suffice, or if the contribution is a specific model and empirical evaluation, it may be necessary to either make it possible for others to replicate the model with the same dataset, or provide access to the model. In general. releasing code and data is often one good way to accomplish this, but reproducibility can also be provided via detailed instructions for how to replicate the results, access to a hosted model (e.g., in the case of a large language model), releasing of a model checkpoint, or other means that are appropriate to the research performed.
        \item While NeurIPS does not require releasing code, the conference does require all submissions to provide some reasonable avenue for reproducibility, which may depend on the nature of the contribution. For example
        \begin{enumerate}
            \item If the contribution is primarily a new algorithm, the paper should make it clear how to reproduce that algorithm.
            \item If the contribution is primarily a new model architecture, the paper should describe the architecture clearly and fully.
            \item If the contribution is a new model (e.g., a large language model), then there should either be a way to access this model for reproducing the results or a way to reproduce the model (e.g., with an open-source dataset or instructions for how to construct the dataset).
            \item We recognize that reproducibility may be tricky in some cases, in which case authors are welcome to describe the particular way they provide for reproducibility. In the case of closed-source models, it may be that access to the model is limited in some way (e.g., to registered users), but it should be possible for other researchers to have some path to reproducing or verifying the results.
        \end{enumerate}
    \end{itemize}

\item {\bf Open access to data and code}
    \item[] Question: Does the paper provide open access to the data and code, with sufficient instructions to faithfully reproduce the main experimental results, as described in supplemental material?
    \item[] Answer: \answerYes{}%
    \item[] Justification: Code is provided in \autoref{app:code} and can be used to reproduce our main results. We will be releasing our code publicly..%
    \item[] Guidelines:
    \begin{itemize}
        \item The answer NA means that paper does not include experiments requiring code.
        \item Please see the NeurIPS code and data submission guidelines (\url{https://nips.cc/public/guides/CodeSubmissionPolicy}) for more details.
        \item While we encourage the release of code and data, we understand that this might not be possible, so “No” is an acceptable answer. Papers cannot be rejected simply for not including code, unless this is central to the contribution (e.g., for a new open-source benchmark).
        \item The instructions should contain the exact command and environment needed to run to reproduce the results. See the NeurIPS code and data submission guidelines (\url{https://nips.cc/public/guides/CodeSubmissionPolicy}) for more details.
        \item The authors should provide instructions on data access and preparation, including how to access the raw data, preprocessed data, intermediate data, and generated data, etc.
        \item The authors should provide scripts to reproduce all experimental results for the new proposed method and baselines. If only a subset of experiments are reproducible, they should state which ones are omitted from the script and why.
        \item At submission time, to preserve anonymity, the authors should release anonymized versions (if applicable).
        \item Providing as much information as possible in supplemental material (appended to the paper) is recommended, but including URLs to data and code is permitted.
    \end{itemize}

\item {\bf Experimental Setting/Details}
    \item[] Question: Does the paper specify all the training and test details (e.g., data splits, hyperparameters, how they were chosen, type of optimizer, etc.) necessary to understand the results?
    \item[] Answer: \answerYes{}%
    \item[] Justification: Training details are outlined in \autoref{app:2d_gmm}, \autoref{app:cls_rtb}, \autoref{app:infilling}, and \autoref{app:offline_rl}.%
    \item[] Guidelines:
    \begin{itemize}
        \item The answer NA means that the paper does not include experiments.
        \item The experimental setting should be presented in the core of the paper to a level of detail that is necessary to appreciate the results and make sense of them.
        \item The full details can be provided either with the code, in appendix, or as supplemental material.
    \end{itemize}

\item {\bf Experiment Statistical Significance}
    \item[] Question: Does the paper report error bars suitably and correctly defined or other appropriate information about the statistical significance of the experiments?
    \item[] Answer: \answerYes{}%
    \item[] Justification: All experimental results have error bars, except the Stable Diffusion finetuning experiment \autoref{sec:experiments:text2image} which was only trained on one seed per prompt due to compute constraints.%
    \item[] Guidelines:
    \begin{itemize}
        \item The answer NA means that the paper does not include experiments.
        \item The authors should answer "Yes" if the results are accompanied by error bars, confidence intervals, or statistical significance tests, at least for the experiments that support the main claims of the paper.
        \item The factors of variability that the error bars are capturing should be clearly stated (for example, train/test split, initialization, random drawing of some parameter, or overall run with given experimental conditions).
        \item The method for calculating the error bars should be explained (closed form formula, call to a library function, bootstrap, etc.)
        \item The assumptions made should be given (e.g., Normally distributed errors).
        \item It should be clear whether the error bar is the standard deviation or the standard error of the mean.
        \item It is OK to report 1-sigma error bars, but one should state it. The authors should preferably report a 2-sigma error bar than state that they have a 96\% CI, if the hypothesis of Normality of errors is not verified.
        \item For asymmetric distributions, the authors should be careful not to show in tables or figures symmetric error bars that would yield results that are out of range (e.g. negative error rates).
        \item If error bars are reported in tables or plots, The authors should explain in the text how they were calculated and reference the corresponding figures or tables in the text.
    \end{itemize}

\item {\bf Experiments Compute Resources}
    \item[] Question: For each experiment, does the paper provide sufficient information on the computer resources (type of compute workers, memory, time of execution) needed to reproduce the experiments?
    \item[] Answer: \answerYes{}%
    \item[] Justification: Compute resources used in experiments is outlined in \autoref{app:cls_rtb}, \autoref{app:infilling}, \autoref{app:offline_rl}, \autoref{app:text2image} and summarised in \autoref{app:compute}.%
    \item[] Guidelines:
    \begin{itemize}
        \item The answer NA means that the paper does not include experiments.
        \item The paper should indicate the type of compute workers CPU or GPU, internal cluster, or cloud provider, including relevant memory and storage.
        \item The paper should provide the amount of compute required for each of the individual experimental runs as well as estimate the total compute. 
        \item The paper should disclose whether the full research project required more compute than the experiments reported in the paper (e.g., preliminary or failed experiments that didn't make it into the paper). 
    \end{itemize}
    
\item {\bf Code Of Ethics}
    \item[] Question: Does the research conducted in the paper conform, in every respect, with the NeurIPS Code of Ethics \url{https://neurips.cc/public/EthicsGuidelines}?
    \item[] Answer: \answerYes{}%
    \item[] Justification: The paper conforms to the stated ethics guidelines.%
    \item[] Guidelines:
    \begin{itemize}
        \item The answer NA means that the authors have not reviewed the NeurIPS Code of Ethics.
        \item If the authors answer No, they should explain the special circumstances that require a deviation from the Code of Ethics.
        \item The authors should make sure to preserve anonymity (e.g., if there is a special consideration due to laws or regulations in their jurisdiction).
    \end{itemize}

\item {\bf Broader Impacts}
    \item[] Question: Does the paper discuss both potential positive societal impacts and negative societal impacts of the work performed?
    \item[] Answer: \answerYes{}%
    \item[] Justification: We discuss broader impact of the work in \autoref{sec:conclusion}.%
    \item[] Guidelines:
    \begin{itemize}
        \item The answer NA means that there is no societal impact of the work performed.
        \item If the authors answer NA or No, they should explain why their work has no societal impact or why the paper does not address societal impact.
        \item Examples of negative societal impacts include potential malicious or unintended uses (e.g., disinformation, generating fake profiles, surveillance), fairness considerations (e.g., deployment of technologies that could make decisions that unfairly impact specific groups), privacy considerations, and security considerations.
        \item The conference expects that many papers will be foundational research and not tied to particular applications, let alone deployments. However, if there is a direct path to any negative applications, the authors should point it out. For example, it is legitimate to point out that an improvement in the quality of generative models could be used to generate deepfakes for disinformation. On the other hand, it is not needed to point out that a generic algorithm for optimizing neural networks could enable people to train models that generate Deepfakes faster.
        \item The authors should consider possible harms that could arise when the technology is being used as intended and functioning correctly, harms that could arise when the technology is being used as intended but gives incorrect results, and harms following from (intentional or unintentional) misuse of the technology.
        \item If there are negative societal impacts, the authors could also discuss possible mitigation strategies (e.g., gated release of models, providing defenses in addition to attacks, mechanisms for monitoring misuse, mechanisms to monitor how a system learns from feedback over time, improving the efficiency and accessibility of ML).
    \end{itemize}
    
\item {\bf Safeguards}
    \item[] Question: Does the paper describe safeguards that have been put in place for responsible release of data or models that have a high risk for misuse (e.g., pretrained language models, image generators, or scraped datasets)?
    \item[] Answer: \answerNA{}%
    \item[] Justification: We do not release models that have risk of misuse.%
    \item[] Guidelines:
    \begin{itemize}
        \item The answer NA means that the paper poses no such risks.
        \item Released models that have a high risk for misuse or dual-use should be released with necessary safeguards to allow for controlled use of the model, for example by requiring that users adhere to usage guidelines or restrictions to access the model or implementing safety filters. 
        \item Datasets that have been scraped from the Internet could pose safety risks. The authors should describe how they avoided releasing unsafe images.
        \item We recognize that providing effective safeguards is challenging, and many papers do not require this, but we encourage authors to take this into account and make a best faith effort.
    \end{itemize}

\item {\bf Licenses for existing assets}
    \item[] Question: Are the creators or original owners of assets (e.g., code, data, models), used in the paper, properly credited and are the license and terms of use explicitly mentioned and properly respected?
    \item[] Answer: \answerYes{}%
    \item[] Justification: Creaters of code we use are properly credited.%
    \item[] Guidelines:
    \begin{itemize}
        \item The answer NA means that the paper does not use existing assets.
        \item The authors should cite the original paper that produced the code package or dataset.
        \item The authors should state which version of the asset is used and, if possible, include a URL.
        \item The name of the license (e.g., CC-BY 4.0) should be included for each asset.
        \item For scraped data from a particular source (e.g., website), the copyright and terms of service of that source should be provided.
        \item If assets are released, the license, copyright information, and terms of use in the package should be provided. For popular datasets, \url{paperswithcode.com/datasets} has curated licenses for some datasets. Their licensing guide can help determine the license of a dataset.
        \item For existing datasets that are re-packaged, both the original license and the license of the derived asset (if it has changed) should be provided.
        \item If this information is not available online, the authors are encouraged to reach out to the asset's creators.
    \end{itemize}

\item {\bf New Assets}
    \item[] Question: Are new assets introduced in the paper well documented and is the documentation provided alongside the assets?
    \item[] Answer: \answerNA{}%
    \item[] Justification: We do not release new assets.%
    \item[] Guidelines:
    \begin{itemize}
        \item The answer NA means that the paper does not release new assets.
        \item Researchers should communicate the details of the dataset/code/model as part of their submissions via structured templates. This includes details about training, license, limitations, etc. 
        \item The paper should discuss whether and how consent was obtained from people whose asset is used.
        \item At submission time, remember to anonymize your assets (if applicable). You can either create an anonymized URL or include an anonymized zip file.
    \end{itemize}

\item {\bf Crowdsourcing and Research with Human Subjects}
    \item[] Question: For crowdsourcing experiments and research with human subjects, does the paper include the full text of instructions given to participants and screenshots, if applicable, as well as details about compensation (if any)? 
    \item[] Answer: \answerNA{} %
    \item[] Justification: We do not have any crowdsourcing experiments or research with human subjects. %
    \item[] Guidelines:
    \begin{itemize}
        \item The answer NA means that the paper does not involve crowdsourcing nor research with human subjects.
        \item Including this information in the supplemental material is fine, but if the main contribution of the paper involves human subjects, then as much detail as possible should be included in the main paper. 
        \item According to the NeurIPS Code of Ethics, workers involved in data collection, curation, or other labor should be paid at least the minimum wage in the country of the data collector. 
    \end{itemize}

\item {\bf Institutional Review Board (IRB) Approvals or Equivalent for Research with Human Subjects}
    \item[] Question: Does the paper describe potential risks incurred by study participants, whether such risks were disclosed to the subjects, and whether Institutional Review Board (IRB) approvals (or an equivalent approval/review based on the requirements of your country or institution) were obtained?
    \item[] Answer: \answerNA{}{} %
    \item[] Justification: We do not have any experiments with human subjects %
    \item[] Guidelines:
    \begin{itemize}
        \item The answer NA means that the paper does not involve crowdsourcing nor research with human subjects.
        \item Depending on the country in which research is conducted, IRB approval (or equivalent) may be required for any human subjects research. If you obtained IRB approval, you should clearly state this in the paper. 
        \item We recognize that the procedures for this may vary significantly between institutions and locations, and we expect authors to adhere to the NeurIPS Code of Ethics and the guidelines for their institution. 
        \item For initial submissions, do not include any information that would break anonymity (if applicable), such as the institution conducting the review.
    \end{itemize}

\end{enumerate}

\end{document}